%% file: main.tex
\documentclass{article} 
\usepackage{iclr2027_conference,times}

\input{preamble}

\input{macros}

\title{\method: Strategy Logic-Guided Multi-Agent \\ Reinforcement Learning}

\author{Arshia Rafieioskouei, Matthew Lucas, Tzu-Han Hsu, Borzoo Bonakdarpour \\
Department of Computer Science and Engineering\\
Michigan State University\\
East Lansing, MI, USA
}

\iclrfinalcopy 
\begin{document}

\maketitle
\lhead{}

\begin{abstract}
{\em Specification-guided} reinforcement learning (RL) translates formal specifications of desired behavior of agents into shaped rewards.
This paper introduces \method, a significant paradigm shift that advances the state of the art to specification-guided RL in adversarial settings under partial observability. 
We employ {\em Strategy Logic} (\SL), which allows explicit quantification over agents' strategies and uses the linear temporal logic (\LTL) to specify the agents' objectives and constraints.
This enables reasoning not only about how agents should act to satisfy a specification, but also about how they should respond to the strategies and behaviors of other agents. 
We evaluate \method on three popular benchmarks: Predator-Prey, Google Research Football, and our home-grown extension of SMAC-HARD, demonstrating its effectiveness, scalability, and ability to adapt to diverse adversarial strategies.

\end{abstract}

\input{01-Introduction}

\input{02-related}
\input{03-prelim}

\input{04-problem}

\input{05-algorithm}

\input{06-experiments}
\input{07-conclusion}

\newpage
\subsection*{AI use statement}

In this work, we used generative AI tools to implement methods and assist in the writing of proofs.
We have not used generative AI tools to help develop theoretical models or conceptual frameworks, formulate mathematical claims, provide critical ingredients for proving mathematical claims, propose or refine hypotheses, design or provide feedback on research  methodology or experiments, interpret results, generate synthetic data sets, assist with translation, or lean and reformat dataset, support qualitative and thematic data analysis are not applicable to this work.
Additionally, we used generative AI tools for brainstorming and to propose a title or keywords for a research paper. We reviewed all AI-assisted work. We used LLMs to assist with parts of the experimental implementation, and all generated code was tested and verified for correctness. We also used LLMs to review our proofs, provide feedback on their clarity, and edit and polish the manuscript. We take full responsibility for the final content of this work, including all text, claims, code, and artifacts produced with the assistance of generative AI.




\subsection*{Reproducibility statement}

We provide the algorithmic details in~\Cref{sec:algorithm}, complemented by the experimental setup in~\Cref{sec:eval}. Complete implementation details and hyperparameters are provided in~\Cref{app:exp}, while benchmark configurations are described in~\Cref{app:bench}.

\bibliography{reference}
\bibliographystyle{iclr2027_conference}

\input{0X-appendix}

\end{document}

%% file: preamble.tex
\usepackage[dvipsnames]{xcolor}
\usepackage[T1]{fontenc}
\usepackage[utf8]{inputenc} 
\usepackage[T1]{fontenc}    
\usepackage{hyperref}       
\usepackage{url}            
\usepackage{booktabs}       
\usepackage{amsfonts}       
\usepackage{nicefrac}       
\usepackage{microtype}      
\usepackage{mathtools}
\usepackage{bm}
\usepackage{amssymb}
\usepackage{tikz}
\usepackage{scalerel}
\usepackage{dsfont}
\usepackage{pgfplots}
\usepgfplotslibrary{fillbetween}
\usetikzlibrary{patterns}
\usetikzlibrary{arrows.meta, positioning}

\usepackage[normalem]{ulem}
\usepackage{graphicx} 
\usepackage{hyperref} 

\usepackage{xspace}
\usepackage{microtype}

\usepackage{makecell}
\usepackage{orcidlink}

\usepackage{graphicx}
\usepackage{amsmath, amssymb}

\usepackage{lua-visual-debug}
\usepackage{wrapfig}
\usepackage{mdframed}
\usepackage{hyperref}
\hypersetup{%
	colorlinks=true,           
	allcolors=blue!50!black,,   
	pdfstartview=Fit,          
	breaklinks=true,
	pdfauthor={},
	pdftitle={{}}
}

\usepackage[capitalize,noabbrev]{cleveref}
\crefname{figure}{Fig.}{Figs.}
\Crefname{figure}{Fig.}{Figs.}

\crefname{table}{Table}{Tabs.}
\Crefname{table}{Table}{Tabs.}

\crefname{section}{Sec.}{Secs.}
\Crefname{section}{Sec.}{Secs.}

\crefname{equation}{Eq.}{Eqs.}
\Crefname{equation}{Eq.}{Eqs.}
\crefname{appendix}{App.}{Apps.}
\Crefname{appendix}{App.}{Apps.}

\usepackage{stmaryrd}
\usepackage{amsfonts}
\usepackage{marvosym}
\usepackage{multirow}

\usepackage{bussproofs}
\usepackage{booktabs}
\usepackage{colortbl}

\usepackage{listings}

\usepackage{subcaption}
\usepackage[textsize=tiny]{todonotes}
\usepackage{graphicx}

\usepackage{stmaryrd}

\usepackage{ltl}
\newsavebox{\rpsbox}

\definecolor{gray}{rgb}{0.5,0.5,0.5}
\definecolor{darkgreen}{rgb}{0,0.4,0}
\definecolor{darkblue}{rgb}{0.2, 0.2, 0.6}
\definecolor{niceblue}{rgb}{0.16, 0.32, 0.75}
\definecolor{niceblack}{rgb}{0.0, 0.18, 0.39}
\definecolor{prettyred}{rgb}{1.0, 0.13, 0.32}
\definecolor{prettypink}{rgb}{0.98, 0.38, 0.5}
\definecolor{prettyblue}{rgb}{0.16, 0.32, 0.75}
\definecolor{prettygreen}{rgb}{0.0, 0.5, 0.0}
\definecolor{prettyorange}{rgb}{0.99, 0.53, 0.05}
\definecolor{prettypurple}{rgb}{0.6, 0.4, 0.8}
\definecolor{prettyyellow}{rgb}{0.99, 0.93, 0.0}
\definecolor{airforceblue}{rgb}{0.36, 0.54, 0.66}

\usepackage[many]{tcolorbox}    	
\usepackage{multicol}               

\colorlet{main}{black}

\colorlet{sub}{main!30}

\tcbset{
	sharp corners,
	colback = white,
	before skip = 0.2cm,    
	after skip = 0.5cm      
}                           

\newtcolorbox{boxA}{
	fontupper = \bf,
	boxrule = 1.5pt,
	colframe = black 
}

\newtcolorbox{boxB}{
	fontupper = \bf\color{main}, 
	boxrule = 1.5pt,
	colframe = main,
	rounded corners,
	arc = 5pt   
}

\newtcolorbox{boxC}{
	colback = sub, 
	boxrule = 0pt  
}

\newtcolorbox{boxD}{
	colback = sub, 
	colframe = main, 
	boxrule = 0pt, 
	toprule = 3pt, 
	bottomrule = 3pt 
}

\newtcolorbox{boxE}{
	enhanced, 
	boxrule = 0pt, 
	borderline = {0.75pt}{0pt}{main}, 
	borderline = {0.75pt}{2pt}{sub} 
}

\newtcolorbox{boxF}{
	colback = sub,
	enhanced,
	boxrule = 1.5pt, 
	colframe = white, 
	borderline = {1.5pt}{0pt}{main, dashed} 
}

\newtcolorbox{boxG}{
	enhanced,
	boxrule = 0pt,
	colback = sub,
	borderline west = {1pt}{0pt}{main}, 
	borderline west = {0.75pt}{2pt}{main}, 
	borderline east = {1pt}{0pt}{main}, 
	borderline east = {0.75pt}{2pt}{main}
}

\newtcolorbox{boxH}{
	colback = sub, 
	colframe = main, 
	boxrule = 0pt, 
	leftrule = 6pt 
}

\newtcolorbox{boxI}{
	colback = sub, 
	colframe = main, 
	boxrule = 0pt, 
	toprule = 6pt 
}

\newtcolorbox{boxJ}{
	sharpish corners, 
	colback = sub, 
	colframe = main, 
	boxrule = 0pt, 
	toprule = 4.5pt, 
	enhanced,
	fuzzy shadow = {0pt}{-2pt}{-0.5pt}{0.5pt}{black!35} 
}

\newtcolorbox{boxK}{
	sharpish corners, 
	colframe =prettyblue!50,
	boxrule = 0pt,
	toprule = 4.5pt, 
	enhanced,
	fuzzy shadow = {0pt}{-2pt}{-0.5pt}{0.5pt}{prettyblue!50} 
}

\newtcolorbox{boxL}{
	fontupper = \color{main},
	rounded corners,
	arc = 6pt,
	colback = sub, 
	colframe = main!50, 
	boxrule = 0pt, 
	bottomrule = 4.5pt 
}

\newtcolorbox{boxM}{
	fontupper = \color{white},
	rounded corners,
	arc = 6pt,
    colback=blue!5,
	colback = main!80, 
	colframe = main, 
	boxrule = 0pt, 
	bottomrule = 4.5pt,
	enhanced,
	fuzzy shadow = {0pt}{-3pt}{-0.5pt}{0.5pt}{black!35}
}

\definecolor{myPurple}{HTML}{A080E1}

\newtcolorbox{problemstatement}[1][]{
    enhanced,
    title={Problem Statement},
    colback=myPurple!3,
    colframe=myPurple!80!black!70,
    coltitle=white,
    colbacktitle=myPurple!80!black!70,
    fonttitle=\bfseries,
    boxrule=1pt,
    arc=8mm,
    left=1.5pt,
    right=1.5pt,
    top=3pt,
    bottom=2pt
}

\newtcolorbox{theoremboxstyle}{
  enhanced,
  colback=green!8,
  colframe=green!60!black,
  boxrule=1pt,
  arc=8mm,
  left=2pt, right=2pt,
  top=5pt, bottom=2pt
}

\newtcolorbox{bluebox}[1][]{
  enhanced,
  colframe=blue!70!black,  
  boxrule=1pt,
  colback=blue!2,
  arc=4mm,                 
  left=1mm, right=1mm, top=0mm, bottom=0.5mm,
  boxsep=0mm,
  #1
}
\newtcolorbox{bluebox2}[1][]{
  enhanced,
  colframe=myPurple,  
  boxrule=1pt,
  arc=5mm,                 
  left=1mm, right=1mm, top=0.5mm, bottom=0.5mm,
  boxsep=1mm,
  #1
}

\crefname{theorem}{Theorem}{Theorems}
\Crefname{theorem}{Theorem}{Theorems}

\definecolor{boxcolor}{HTML}{E9ECEE}
\definecolor{myyelllow}{HTML}{FDBE00}
\definecolor{myorrange}{HTML}{FF7063}
\definecolor{mygreenup}{HTML}{0A8842}

\newtcolorbox{theorembox}{
    colback=boxcolor,
    colframe=boxcolor,
    boxrule=0pt,
    arc=2mm,
    left=2pt,
    right=2pt,
    top=2pt,
    bottom=2pt
}

\newtcolorbox{contbox}{
    colback=myyelllow!12,
    colframe=myyelllow,
    boxrule=0pt,
    arc=2mm,
    left=4pt,
    right=2pt,
    top=2pt,
    bottom=2pt
}

\newtcolorbox{takeawaybox}{
    colback=myorrange!12,
    colframe=myorrange,
    boxrule=0pt,
    arc=50mm,
    outer arc=50mm,
    left=2pt,
    right=2pt,
    top=2pt,
    bottom=2pt
}

\newtheorem{theorem}{Theorem}[section]

\newtheorem{lemma}[theorem]{Lemma}


%% file: macros.tex
{}

{}

{}

\definecolor{c1}{HTML}{602273}
\definecolor{c2}{HTML}{F39F5A}
\definecolor{c3}{HTML}{AE445A}
\definecolor{c4}{HTML}{6e4909}

\colorlet{statementColor}{c3}

\newtheorem{proof}{Proof}

\usepackage{scalefnt}

\colorlet{command-color}{black!70}
\definecolor{dkcyan}{rgb}{0.1, 0.3, 0.3}
\definecolor{dkgreen}{rgb}{0,0.3,0}
\definecolor{dkblue}{rgb}{0,0.3,1.0}
\colorlet{comment-color}{black!50}

\lstdefinelanguage{code-lang}{
	keywords={def, repeat, return, if, then, else,or},
	keywordstyle=[1]\color{command-color},
	morekeywords=[2]{faultLocalization,instrument, symbolicExecution,SyGuS},
	keywordstyle=[2]\color{dkgreen},
	morekeywords=[3]{iterativeRepair},
	keywordstyle=[3]\color{dkcyan},
	comment=[l][\color{comment-color}]{//},
	literate=%
	{=}{{{\color{command-color}=}}}1
	{|}{{{\color{command-color}|}}}1
	{:}{{{\color{command-color}:}}}1
	{:=}{{{\color{command-color}:=}}}1
	{@}{ }1
}

\lstdefinestyle{code-style}{
	escapeinside={(*}{*)},
	basicstyle=\ttfamily\fontsize{9}{11}\selectfont,
	columns=fullflexible,
	commentstyle=\sffamily\color{black!50!white},
	framexleftmargin=1em,
	framexrightmargin=1ex,
	keepspaces=true,
	mathescape,
	numbers=left,
	numberblanklines=false,
	numbersep=0.5em,
	numberstyle=\relscale{0.65}\color{gray}\ttfamily,
	showstringspaces=true,
	stepnumber=1,
	xleftmargin=1.2em,
}

\lstdefinelanguage{example-lang}{
	keywords={while,do,if, then,else, observe,or},
	keywordstyle=[1]{\color{c3}},
	morekeywords=[2]{string,int,bool},
	keywordstyle=[2]\color{black!60},
	morekeywords=[3]{phase,title,session,decision,print,password,attack,request,username,date,userPassword,info,LOG,credentials,reviewA,reviewB, reviewerAid,reviewerBid,order,notification,withdraw,balance,amount,ErrorLog,TransactionLog},
	keywordstyle=[3]\color{c1},
	morekeywords=[4]{366,1,0,true},
	keywordstyle=[4]\color{c4},
	morestring=[b]",
	stringstyle=\color{c4},
	comment=[l][\color{comment-color}]{//},
	literate=%
	{=}{{{\color{c3}=}}}1
	{!=}{{{\color{c3}!=}}}1
	{>}{{{\color{c3}>}}}1
	{<}{{{\color{c3}<}}}1
	{+}{{{\color{c3}+}}}1
	{-}{{{\color{c3}-}}}1
	{*}{{{\color{c3}*}}}1
	{@}{ }1
}

\lstdefinestyle{example-style}{
	escapeinside={(*}{*)},
	basicstyle=\ttfamily\fontsize{8}{10}\selectfont,
	columns=fullflexible,
	commentstyle=\sffamily\color{black!50!white},
	framexleftmargin=0em,
	framexrightmargin=0ex,
	keepspaces=true,
	mathescape,
	numbers=left,
	numberblanklines=false,
	numbersep=1.0em,
	numberstyle=\relscale{0.65}\color{gray}\ttfamily,
	showstringspaces=true,
	stepnumber=1,
	xleftmargin=1.2em
}

\lstdefinestyle{example-style-large}{
	escapeinside={(*}{*)},
	basicstyle=\ttfamily\fontsize{10}{12}\selectfont,
	columns=fullflexible,
	commentstyle=\sffamily\color{black!50!white},
	framexleftmargin=0em,
	framexrightmargin=0ex,
	keepspaces=true,
	mathescape,
	numbers=left,
	numberblanklines=false,
	numbersep=0.5em,
	numberstyle=\relscale{0.65}\color{gray}\ttfamily,
	showstringspaces=true,
	stepnumber=1
}

\lstnewenvironment{code}[1][]
{\small
	\lstset{
		style=code-style, 
		language=code-lang,
		#1
	}
}
{}

\lstnewenvironment{exampleCode}[1][]
{\small
	\lstset{
		style=example-style, 
		language=example-lang,
		#1
	}
}
{}

\newcommand{\AP}{\mathsf{AP}}

\newcommand{\finite}{*}

\newcommand{\definedas}{\triangleq}

\newcommand{\posg}{G}
\newcommand{\states}{S}
\newcommand{\state}{\mathit{s}}
\newcommand{\labels}{\mathit{L}}
\newcommand{\labelsof}[1]{\mathit{L}(#1)}

\newcommand{\actions}{A}
\newcommand{\action}{\mathit{a}}

\newcommand{\trans}{\mathbf{P}}

\newcommand{\init}{\states_0}

\newcommand{\ObsSys}{\mathit{O}}
\newcommand{\ObsFun}{\Omega}
\newcommand{\actobshist}{\mathcal{H}}
\newcommand{\obs}{o}
\newcommand{\Ag}{{Ag}}
\newcommand{\agent}{{\alpha}}
\newcommand{\horizon}{T}

\newcommand{\edgeof}[1]{\xrightarrow{#1}}
\newcommand{\prob}{\mathbb{P}}

\newcommand{\optimal}{\star}

\newcommand{\traj}{\zeta}
\newcommand{\trajs}{\mathcal{Z}}

\newcommand{\actionsj}{\mathcal{A}}

\newcommand{\actionj}{\mathbf{a}}

\newcommand{\SL}{{SL}\xspace}

\newcommand{\LTL}{{LTL}\xspace}

\newcommand{\trajfromto}[3]{{#1}_{\bracketof{{#2}:{#3}}}}

\newcommand{\predicatefunc}{\mathit{f}}
\newcommand{\predicatefuncof}[1]{\predicatefunc\big({#1}\big)}

\newcommand{\trace}{\mathit{t}}
\newcommand{\traceof}[1]{\mathsf{Tr}(#1)}
\newcommand{\traceoff}[1]{\mathsf{Tr}\big(#1\big)}
\newcommand{\tracesof}[1]{\mathsf{Traces}(#1)}

\newcommand{\OR}{\vee}

\newcommand{\always}{\G}
\newcommand{\until}{\mathbin{\mathcal{U}}}

\newcommand{\tru}{\mathsf{true}}
\newcommand{\fals}{\mathsf{false}}
\newcommand{\G}{\LTLsquare}
\newcommand{\F}{\LTLdiamond}
\newcommand{\X}{\LTLcircle}
\newcommand{\then}{\mathcal{T}}
\newcommand{\U}{\mathbin{\mathcal{U}}}

\newcommand{\startVars}{\mathit{Vars(\varphi)}}
\newcommand{\strategyVars}{\mathit{Vars(\varphi)}}

\newcommand{\var}{p}

\newcommand{\quant}{\mathds{Q}}

\newcommand{\rbvalue}{{\rho}}
\newcommand{\rbvalueof}[1]{\rbvalue\big({#1}\big)}

\newcommand{\maximum}{\textsf{max}}
\newcommand{\minimum}{\textsf{min}}
\newcommand{\maxamong}[1]{\underset{#1}{\textsf{max}}}
\newcommand{\minamong}[1]{\underset{#1}{\textsf{min}}}
\newcommand{\distrubition}{\Delta}
\newcommand{\distributionof}[1]{\distrubition({#1})}

\newcommand{\precc}{\mathbf{Pre}}

\newcommand{\setof}[1]{ \{#1\} }

\newcommand{\brfunc}{\mathbf{BR}}
\newcommand{\bfunc}{\mathbf{RS}}
\newcommand{\bfunca}{\mathbf{RS}^a}
\newcommand{\brfunca}{\mathbf{BR}^a}

\newcommand{\allforall}{\mathds{\quant}}
\newcommand{\allexists}{\mathds{\quant}}
\newcommand{\forallsof}[1]{{\allforall^\forall_{#1}}}
\newcommand{\existsof}[1]{{\allexists^\exists_{#1}}}

\newcommand{\sizeof}[1]{{|#1|}}

\newcommand{\assignment}{\mathcal{X}}
\newcommand{\assignments}{\mathbb{X}}

\newcommand{\strategies}{\Pi}

\newcommand{\expected}{\mathbb{E}}
\newcommand{\discount}{\gamma}

\newcommand{\history}{\mathit{h}}

\newcommand{\tupleof}[1]{\langle #1 \rangle}

\newcommand{\pos}{\texttt{pos}}

\newcommand{\locmark}{\it} 

\newcommand{\ship}{\textsf{\small ship}\xspace} 
\newcommand{\pirate}{\textsf{\small pirate}\xspace}

\newcommand{\orderedunion}{\oplus}

\newcommand{\bracketof}[1]{ [ #1 ] }

\newcommand{\method}{\textsc{SLICR}\xspace}
\newcommand{\cavcomments}[1]{}

\newcommand{\converge}{\overset{\optimal}{\rightarrow}}

\newcommand{\orangeline}{(\textcolor{orange}{orange line})\xspace}

\newcommand{\redline}{(\textcolor{red}{red line})\xspace}

\newcommand{\blueline}{(\textcolor{blue}{blue line})\xspace}

\newcommand{\greenline}{(\textcolor{Green}{green line})\xspace}

\newcommand{\stratspace}{\Lambda}
\newcommand{\latent}{\lambda}

\newcommand{\enc}{E}
\newcommand{\deca}{\mathit{DA}}
\newcommand{\deco}{\mathit{DO}}
\newcommand{\future}{F}

\newcommand\shieldbase{%
  \tikz[baseline=0.15ex, x=1ex, y=1ex]{
    \draw[
      line width=1.2pt,
      line cap=round,
      line join=round
    ]
    (1,2.25)
      .. controls (0.72,2.00) and (0.38,1.78) .. (0.08,1.78)
      -- (0.08,0.85)
      .. controls (0.08,0.15) and (0.45,-0.38) .. (1,-0.55)
      .. controls (1.55,-0.38) and (1.92,0.15) .. (1.92,0.85)
      -- (1.92,1.78)
      .. controls (1.62,1.78) and (1.28,2.00) .. (1,2.25)
      -- cycle;
  }%
}

\DeclareRobustCommand{\shield}{%
  \mathord{\scalerel*{\shieldbase}{\bigcirc}}%
}

\newcommand{\accept}{%
  \mathbin{\vcenter{\hbox{\includegraphics[height=2ex]{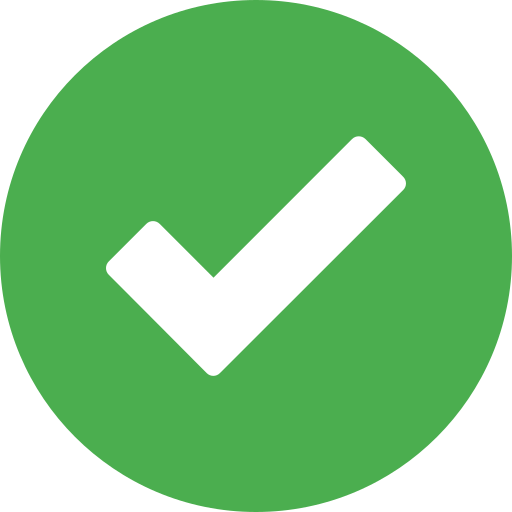}}}}%
}

\newcommand{\warn}{%
  \mathbin{\vcenter{\hbox{\includegraphics[height=2.5ex]{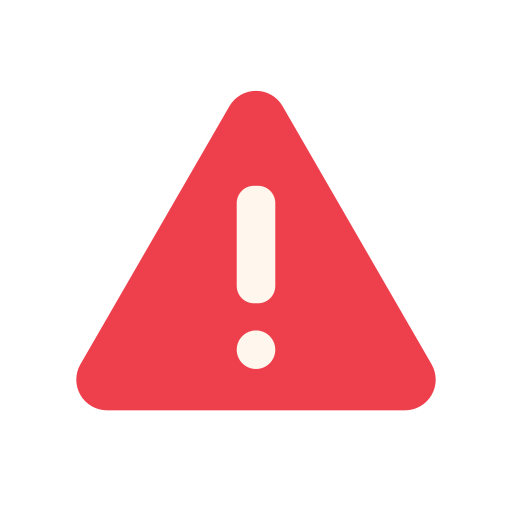}}}}%
}

\newcommand{\context}{\Xi}
\newcommand{\pp}{\text{pp}}
\newcommand{\formulapp}{\varphi_\pp}
\newcommand{\formulars}{\varphi_{\text{R2S}}}
\newcommand{\formularsh}{\varphi_{\text{R2S-DEF}}}
\newcommand{\formulasmac}{\varphi_{\text{SMAC}}}
\newcommand{\gain}[1]{{\color{mygreenup} \uparrow \hspace{-1mm}#1}}

\newcommand{\LEMMABS}{Under the order-preservation assumption, any tuple of best-response functions $\tupleof{\brfunc_{x_i}}_{i\in\existsof{}}$ that maximizes~\Cref{eq:opt} also solves the optimization problem stated in~\Cref{sec:problem}.}

\newcommand{\LEMMARB}{If the objective in~\Cref{eq:learning} strictly preserves the ordering induced by~\Cref{eq:opt} over well-typed response-function tuples, then every maximizer of~\Cref{eq:learning} also maximizes~\Cref{eq:opt}.}

\newcommand{\THEOREMFINAL}{Given a POSG $\posg$ and an \SL sentence $\varphi$, suppose that $\latent_\context$ is an information-preserving latent embedding of each $\context$ and that the action-level response functions can represent every well-typed response function.
Then any $\tupleof{\brfunca_{x_i}}_{i\in\existsof{}}$ that maximizes the expected cumulative discounted robustness induces a  $\tupleof{\brfunc_{x_i}}_{i\in\existsof{}}$ that maximizes the objective in~\Cref{eq:learning}.
By Lemma~\ref{lem:optimality-robustness}, this tuple also optimizes~\Cref{eq:opt}, 
and hence, by Lemma~\ref{lem:optimality-best-response}, optimizes the probability of satisfying $\varphi$ in $\posg$ as defined in~\Cref{sec:problem}. }

\newcommand{\revise}[1]{\textcolor{black}{#1}}

%% file: 01-Introduction.tex
\vspace{-3mm}
\section{Introduction}

{\em Specification-guided} reinforcement learning has recently gained significant attention by using desired agent behaviors expressed in formal languages to guide the learning process \citep{10.1145/3744706}.
Formal specification provides rigorous syntax and semantics for reasoning about sequences of events and states over time.
Recently, approaches such as~\citep{hyprl2025,rafieioskouei2026hypole} have demonstrated that formal specifications not only provide precise objectives for multiple agents, but also improve learning performance.
However, existing specification-guided multi-agent RL (MARL) approaches~\citep{elsayedaly2022logicbasedrewardshapingmultiagent,liu2024guidingmultiagentmultitaskreinforcement,leon2020extendedmarkovgameslearn} typically assume that other agents follow stationary strategies.
More specifically, these approaches have two main limitations: (1) they learn a single strategy (i.e., policy) for a fixed task or game, without explicitly adapting to the strategies of other agents; and (2) adversarial formulations often focus on rational equilibrium behavior, such as Nash equilibria~\citep{jothimurugan2022specification}, without considering arbitrary or irrational opponent strategies.
These limitations severely limit the ability of agents to respond to a changing collection of adversarial strategies.

\begin{wrapfigure}{r}{0.28\textwidth}
\vspace{-7mm}
    \centering
    \includegraphics[width=0.28\textwidth]{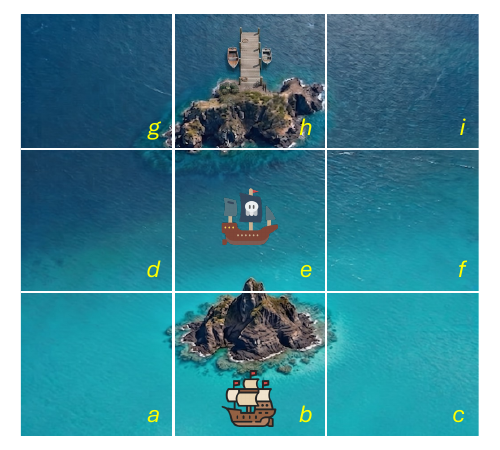}
    \vspace{-7mm}
    \caption{Voyage Scenario.}
    \vspace{-3mm}
    \label{fig:posg}
\end{wrapfigure}
In this paper, we introduce \method (\underline{\bf S}trategy \underline{\bf L}ogic-\underline{\bf I}nformed \underline{\bf C}ontrol through \underline{\bf R}einforcement Learning), advancing the state of the art in specification-guided model-free MARL to adversarial settings under partial observability and non-stationary adversarial behavior.
We employ the highly expressive {\em strategy logic} (\SL)~\citep{10.1145/2631917}, which enables explicit quantification over multiple agents' strategies, along with the linear temporal logic (\LTL) to specify the agents' objectives and constraints.
This enables reasoning not only about how agents should act to satisfy a specification, but also about how they should respond to the strategies of other agents without making any assumptions about them. 
Consider the partially observable \emph{Voyage} scenario shown in \cref{fig:posg}. The agents, \ship (\raisebox{-0.15em}{\includegraphics[height=1.2em]{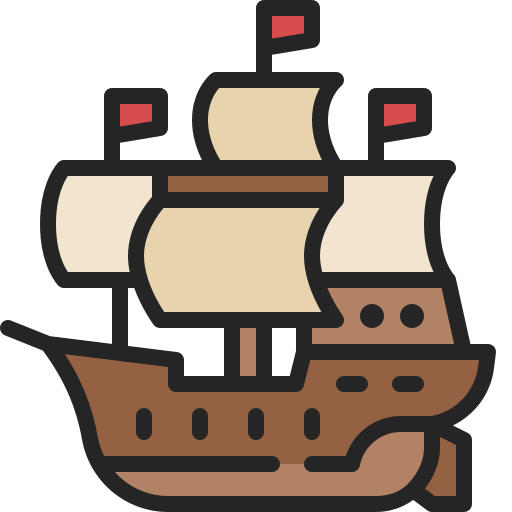}}) and \pirate (\raisebox{-0.15em}{\includegraphics[height=1.2em]{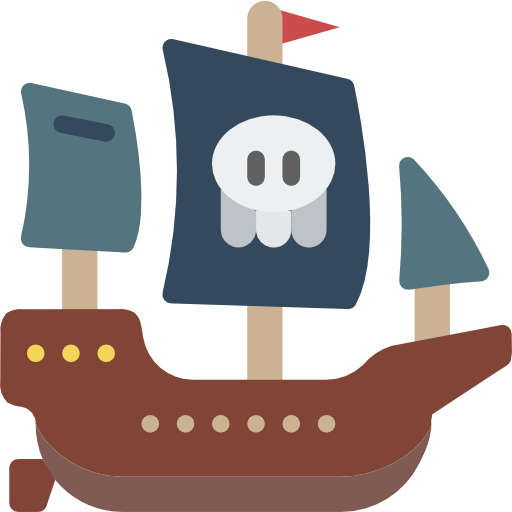}}) observe only their own locations and can move between adjacent cells, except between $b$ and $e$ or $h$ and $e$. 
The \SL formula $\exists y.\forall x(\ship,y)(\pirate,x).\big(\G\neg\textsf{Inter}(\pirate)\ \land \F \textsf{h}(\ship)\big)$ states that there exists a strategy for \ship such that, for every strategy of \pirate, \ship \revise{\emph{eventually}} reaches $h$ and \revise{\emph{always} avoids interception}.
This formula cannot be satisfied because \pirate can wait at $d$ or $f$, depending on the route chosen by \ship. By contrast, $\forall x.\exists y(\pirate,x)(\ship,y).\psi$ stipulates that for every strategy of \pirate, there exists a strategy for \ship to satisfy the objective.
Thus, \SL can express adversarial interactions by capturing dependencies between agents' strategies and specifying how agents should respond to one another.

\begin{figure}
\vspace{-9mm}
    \centering
    \includegraphics[width=0.94\linewidth]{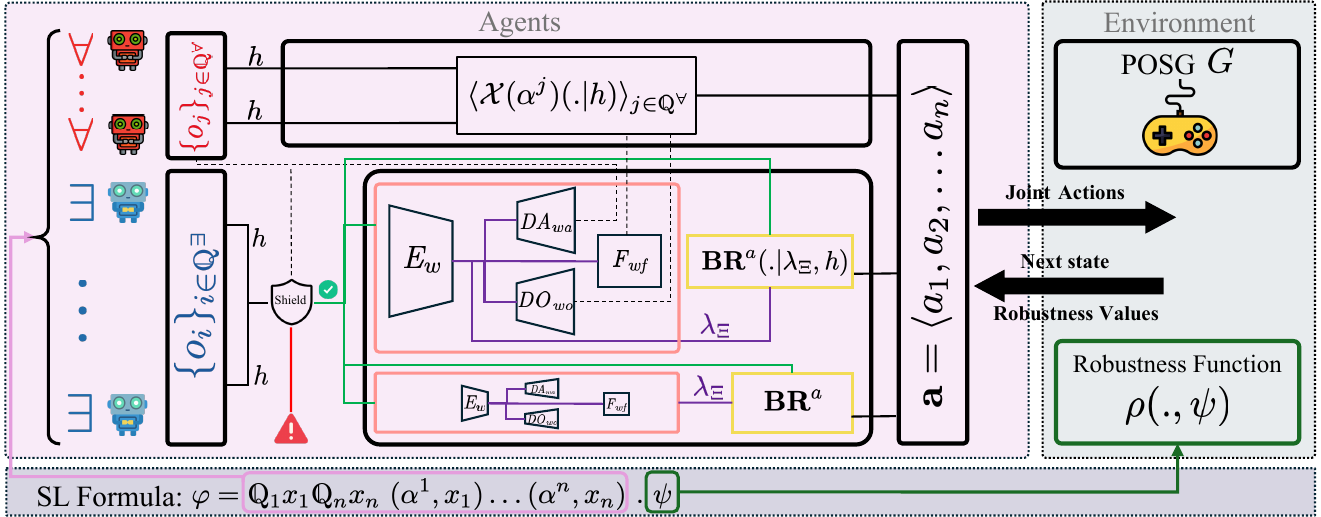}
    \vspace{-2mm}
    \captionof{figure}{\small \textbf{Overview of \method}. Solid lines apply during training and evaluation; dashed lines apply only during training.}
    \label{fig:SLICR-method}
    \vspace{-6mm}
\end{figure}

\begin{contbox}
    \paragraph{Contributions.} We \textbf{(1)} formulate the strategy synthesis problem for a multi-agent system as a learning problem, where the goal is to maximize the probability of satisfying an \SL specification under partial observability (\Cref{sec:problem}); \textbf{(2)} solve this problem (see~\Cref{fig:SLICR-method}) by {\em (i)} transforming the temporal satisfaction problem into a robustness optimization problem, {\em (ii)} identifying irrational strategies assigned to universally quantified strategy variables and preventing them from misleading learning through a shielding mechanism, {\em (iii)} new opponent modeling based on a novel encoder-decoder technique, {\em (iv)} following the centralized training with decentralized execution (CTDE) paradigm to learn decentralized response functions under partial observation (\Cref{sec:algorithm}), and \textbf{(3)} evaluate \method on three MARL benchmarks: Predator-Prey~\citep{NIPS2017_68a97503}, Google Research Football (GRF)~\citep{DBLP:conf/aaai/KurachRSZBERVMB20}, and a home-grown extension of SMAC-HARD~\citep{deng2024smachardenablingmixedopponent}, using MARL algorithms to construct diverse opponent strategies.
    Our evaluation demonstrates the effectiveness of \method, with improvements of {\bf up to $\mathbf{90\%}$} over opponent-modeling baseline, and scalability to complex environments such as GRF and SMAC-HARD~(\cref{sec:eval}).
\end{contbox}
\vspace{-3mm}

We provide proofs, implementation and benchmark details, and additional results in App.~\ref{app:outline}.

%% file: 02-related.tex
\vspace{-4mm}
\section{Related Work}
\vspace{-3mm}

\paragraph{Specification-Guided MARL.}
%
%
Specification-guided MARL has been extensively studied~~\citep{elsayedaly2022logicbasedrewardshapingmultiagent,liu2024guidingmultiagentmultitaskreinforcement,hyprl2025,leon2020extendedmarkovgameslearn,jothimurugan2022specification}.
\citet{hammond2021multiagentreinforcementlearningtemporal} learn a joint policy that maximizes a weighted sum of the satisfaction probabilities of multiple \LTL specifications. ~\citet{10504951} instead consider a turn-based stochastic game and learn a controller policy that maximizes the probability of satisfying an \LTL specification against a worst-case adversary. 
Unlike these settings, our framework uses \SL under partial observability to express richer strategic relationships.
In~\citep{rafieioskouei2026hypole}, the authors use encoder-decoder models to reason about the dependencies represented by Skolem functions. Their encoder-decoder models reconstruct the states of universally quantified trace variables to guide cooperative MARL. In contrast, we use encoder-decoder models to reason over strategies of other agents for \SL-guided adversarial MARL.
We believe the work most closely related to \method is~\citep{doi:10.3233/FAIA240676}, which studies RL in fully observable settings for strategic specifications expressed in probabilistic ATL (PATL), an extension of ATL~\citep{10.1145/585265.585270}.
Unlike \method, their approach supports only a single outermost strategic quantifier without alternation.

\vspace{-3mm}

\paragraph{Opponent Modeling.} 
In \method, we reason about the strategies (behavior) of other agents, connecting our approach to opponent modeling.
\citet{pmlr-v80-rabinowitz18a} uses character and mental-state embeddings to represent agents' long-term and transient behavior, relational forward models to predict future actions or returns from global states~\citep{tacchetti2018relational}, and hierarchical sequential variational autoencoders to infer beliefs over other agents' strategies~\citep{DBLP:conf/atal/ZintgrafDCWH21}.
Policy representations can also be learned from observation-action trajectories, although representing unseen agents requires access to several of their previous trajectories~\citep{pmlr-v80-grover18a}.
In~\citep{NEURIPS2024_6fb9ea51}, opponents' subgoals are inferred as embeddings of future states from historical trajectories under full observability and used to condition the learning agent's policy.
Extending opponent modeling to partially observable settings, \citep{NEURIPS2021_a03caec5} reconstructs other agents' actions and observations from the controlled agent's local history, while \citep{odrowazsypniewski2026generalizedintentionmodelingmultiagent} additionally models future states and returns and aggregates the resulting embeddings.
In contrast, our learning is further guided by a specification and protected by a shield that excludes experiences associated with irrational strategies.

%% file: 03-prelim.tex
\vspace{-3mm}
\section{Preliminaries}
\label{sec:prelim}
\vspace{-2mm}

Let $\distributionof{X} \triangleq \left\{ p : X \rightarrow [0,1] \;\middle|\; \sum_{x \in X} p(x) = 1 \right\}$ denote the probability simplex over a set $X$,
and let $\mu(X)$ denote the uniform distribution,
We abbreviate a tuple $\tupleof{x_1, \ldots, x_n}$ as $\tupleof{x_i}_{i \in \setof{1, \ldots, n}}$.

\vspace{-2mm}

\subsection{Partially Observable Stochastic Games}
\vspace{-1mm}
We model the {\em environment} as an {\em $n$-agent partially observable stochastic game} (POSG)~\citep{1597148.1597262}.
A POSG is defined as a tuple $ \posg \definedas \tupleof{\states, \init, \Ag, \actions, \trans, \ObsSys, \tupleof{\ObsFun_{\agent^i}}_{i \in \setof{1, \ldots, n}}, \AP, \labels}$, where $\states$ is the set of environment {\em states}, and $\states_0$ is the {\em initial-state} distribution. 
$\Ag \definedas \{\agent^1,\ldots,\agent^n\}$ is a finite set of $n$ {\em agents}.
We assume each agent $\agent^i$ has the same {\em action space} $\actions$ and we define
$\actionsj \definedas \{\actionj = \tupleof{\action_1, \ldots, \action_n} \mid a_i \in \actions\}$ as the joint action space.
%
The transition function $\trans:\states\times\actionsj\rightarrow\distributionof{\states}$, where $\trans(\state'\mid\state,\actionj)$ denotes the probability of transitioning from $\state\in\states$ to $\state'\in\states$ under the joint action $\actionj\in\actionsj$.
%
%
In a POSG, agents do not have direct access to the full environment state. 
Instead, they receive a partial observation determined probabilistically by the state and the joint action. 
For simplicity, we assume each agent $\agent^i$ has the same {\em observation space} $\ObsSys$.
%
For each agent $\agent^i$, the observation function $\ObsFun_{\agent^i}:\states\times\actionsj\rightarrow\distributionof{\ObsSys}$, where $\ObsFun_{\agent^i}(\obs_i\mid\state,\actionj)$ denotes the probability that $\agent^i$ receives $\obs_i\in\ObsSys$ in state $\state\in\states$ under the joint action $\actionj\in\actionsj$.
Lastly, 
$\AP$ denotes a finite set of {\em atomic propositions}, 
and $\labels: \states \rightarrow 2^\AP$ is a {\em labeling} function.
\vspace{-2mm}

\paragraph{Example.}
The POSG for the Voyage scenario in~\cref{fig:posg} is defined as follows. The environment is a $3\times3$ grid world with cells labeled ${\locmark a},{\locmark b},\ldots,{\locmark i}$.
$\Ag = \{\ship (\raisebox{-0.15em}{\includegraphics[height=1.2em]{img/ship.png}}), \pirate (\raisebox{-0.15em}{\includegraphics[height=1.2em]{img/pirate-ship.png}})
\}$.
Each state is of the form $(\pos_{\ship}, \pos_\pirate)$, for the position of \ship and \pirate, respectively, where $\pos_\ship, \pos_\pirate \in \{{\locmark a},{\locmark b},\ldots,{\locmark i}\}$. 
Let $(\locmark b,\locmark e)$ be the initial state.
Both agents can move to adjacent cells using actions $\actions = \{U,D,R,L,S\}$, except  between ${\locmark b}$ and ${\locmark e}$ or ${\locmark h}$ and ${\locmark e}$.
%
Each agent can only observe its own current position and not the position of the other agent (i.e., $\ObsSys = \setof{ {\locmark a},{\locmark b},\ldots,{\locmark i}}$).
For example, in state $s = (b,e)$, the observations of \ship and \pirate are $b$ and $e$, respectively. 


A \emph{history} for an agent $\agent^i \in \Ag$ is an observation-action sequence 
$\history_{\agent^i} \triangleq \obs_{i_{0}} \edgeof{\action_{i_{0}}} \obs_{i_{1}} \edgeof{\action_{i_{1}}} \cdots $, 
where for all $j \geq 0$, $\obs_{i_{j}} \in \ObsSys$ and $a_{i_{j}} \in \actions$. 
A {\em sub-history} $\history_{{\agent^i}_{\bracketof{\ell:k}}}$ is a finite segment $\obs_{i_{\ell}} \edgeof{\action_{i_{\ell}}} \cdots \edgeof{\action_{i_{{k-1}}}} \obs_{i_{k}}$, for $0 \leq \ell < k < \sizeof{h_{\agent^i}}$.
%
%
We denote the set of all finite histories by $\actobshist^\finite$.
A \emph{strategy}
$\pi \colon \actobshist^\finite \rightarrow \distributionof{\actions}$, where $\pi(\action \mid h)$ denotes the probability of choosing an action 
$\action \in \actions$ after a finite history $\history \in \actobshist^\finite$.
We denote the set of all strategies in $\posg$ by $\strategies$.
%
%
A {\em path} in a POSG is a sequence of environment states $\traj \triangleq \state_0 \edgeof{\actionj_0} \state_1 \edgeof{\actionj_1} \cdots$, where $\state_j \in \states$ and $\actionj_j \in \actionsj$ for all $j \geq 0$, and 
a {\em sub-path} $\traj_{\bracketof{\ell:k}}$ is a finite segment $\state_\ell \xrightarrow{\actionj_\ell} \cdots \xrightarrow{\actionj_{k-1}} \state_k$, for $0 \leq \ell < k < \sizeof{\traj}$. 
The {\em trace} $\trace$ of a path $\traj$ is a sequence of labels 
$\traceof{\traj} \triangleq \trace(0)\trace(1)\cdots$, such that $\trace(j) = 
\labels(\state_j)$ for all $j \geq 0$
(the trace of a sub-path, $\traceof{\traj_{\bracketof{\ell:k}}}$, is defined similarly).
Notice that, we only consider \textit{valid} paths and sub-paths, where $\trans(s' \mid s, \actionj) > 0$ for all transitions.
%
%
We write $\trajs^\finite$ to denote the set of all finite paths, and
$\tracesof{\trajs^\finite}$ to denote the set of all finite traces.
Furthermore, our setting is episodic, and each episode (a finite sequence of interactions from an initial state to termination) terminates after $\horizon$ steps, where $\horizon \in \mathbb{Z}_{>0}$.
\vspace{-2mm}

\subsection{Strategic Logic (SL)}

\paragraph{Syntax.} \SL~\citep{10.1145/2631917} is a temporal logic that allows explicit quantification over the strategies of agents.
\SL syntax is as follows:
\begin{align*}
\varphi ::= \exists x.\,\varphi \mid \forall x.\,\varphi \mid (\agent,x).\varphi \mid \psi \quad ~~ \psi  ::= \var \mid \lnot \psi \mid \psi \OR \psi \mid \X \psi \mid \psi \until \psi \mid \psi \then \psi
\end{align*}
where $\var \in \AP$ is an atomic proposition, $\agent \in \Ag$ is an agent, and $x$ is a {\em strategy variable}.
We read $\exists x$ as ``there exists a strategy assigned to $x$'', and $\forall x$ as ``for all strategies assigned to $x$''. 
An {\em agent binding} $(\alpha,x)$ binds an agent $\agent$ to the strategy associated with variable $x$. 
{We write $\strategyVars$ for the set of all strategy variables appearing in $\varphi$.} 
The grammar of the inner sub-formula $\psi$ yields a standard \LTL formula with the Boolean connectives $\lnot$ and $\OR$, as well as the temporal operators $\X$ (``next''), $\until$ (``until''), and $\then$ (``then'').
Other Boolean and temporal operators are 
defined as 
syntactic sugar: $\tru \triangleq \var \vee \neg \var$, $\fals \triangleq \neg \tru$, $\psi_1 \rightarrow \psi_2 \triangleq \neg \psi_1 \vee \psi_2$, $\F \psi \triangleq \tru \until \psi$, $\always \psi \triangleq \neg \F \neg \psi$, $\psi_1\until_W\psi_2\definedas(\psi_1\until\psi_2)\lor\always\psi_1$, and $\psi_1\then_{W}\psi_2\triangleq(\psi_1\then\psi_2)\lor(\F\psi_1)$ where $\F$, $\always$, $\until_W$, and $\then_W$ denote `eventually', `always', `weak until', and `weak then', respectively.
We say a formula $\varphi$ is \emph{agent-closed} if every agent $\alpha \in \Ag$ occurring in the formula is bound to some variable $x \in \strategyVars$, and is 
\emph{variable-closed} if every variable $x \in \strategyVars$ is quantified.
A formula that is both variable-closed and agent-closed is called a \emph{sentence}.
\paragraph{Semantics.} Let $\varphi \definedas \quant_1 x_1 \ldots \quant_n x_n ~(\alpha^1, x_1) \ldots (\alpha^n, x_n).\psi$ be an \SL sentence, where each $\quant_i \in \{\forall, \exists\}$ and $\psi$ is the inner \LTL formula.
The original semantics of \SL in ~\citep{10.1145/2631917} is defined over deterministic concurrent games. Since we model RL environments by POSGs, we present the semantics of \SL in two steps. The first step is concerned with the semantics of strategy quantifiers. The second step is related to the semantics of the inner LTL formula $\psi$ for (sampled) paths from a POSG:


\vspace{-1mm}
\noindent --- An {\em assignment} is a partial mapping $\assignment \colon \Ag \, \cup \, \startVars \rightharpoonup \strategies$ that maps each agent and strategy variable in its domain to a strategy.
%
%
Given an assignment $\assignment$, a strategy variable $x_i$, an agent $\alpha^i$, and a strategy $\pi\in\strategies$, we write $\assignment[x_i\mapsto\pi]$ for the assignment that agrees with $\assignment$ except that $x_i$ is mapped to strategy $\pi$. 
Similarly, $\assignment[\alpha^i\mapsto\pi]$ denotes the assignment that agrees with $\assignment$ except that $\alpha^i$ is mapped to $\pi$.
We also write $\assignment[x_i,\alpha^i\mapsto\pi]
\definedas
\assignment[x_i\mapsto\pi][\alpha^i\mapsto\pi]$.
An assignment $\assignment$ is \emph{complete} if it assigns a strategy to every agent and every variable. 
%
%
%
We use $\assignments_c$ to denote the set of all complete assignments and $\assignment_\emptyset$ for the empty assignment.
The satisfaction relation is defined over 
a {\em model configuration} $(\posg,\strategies, \assignment, i)$ and $\varphi$, where $i \in \mathbb{Z}_{\geq 0}$ indicates the evaluation position. 
\vspace{-2mm}
\[
\begingroup
\renewcommand{\arraystretch}{1.2}
{
\setlength{\arraycolsep}{3pt}
\begin{array}{@{}ll@{\hspace{.5em}}l@{\hspace{1em}}l@{}}
(\posg,\strategies,\assignment,0) & \models \exists x.\varphi & \text{iff} &
\text{there exists } \pi \text{ in } \strategies \text{ st. } (\posg,\strategies,\assignment[x \mapsto \pi],0) \models \varphi,\\
(\posg,\strategies,\assignment,0) & \models \forall x.\varphi & \text{iff} &
\text{for all } \pi \text { in } \strategies \text{ st. } (\posg,\strategies,\assignment[x \mapsto \pi],0) \models \varphi,\\
(\posg,\strategies,\assignment,0) & \models (\agent,x).\varphi & \text{iff} &
(\posg,\strategies,\assignment[\agent \mapsto \assignment(x)],0) \models \varphi.
%
%
\end{array}
}
\endgroup
\]
\vspace{-2mm}

\vspace{-3mm}

\noindent --- Now, for each complete assignment $\assignment\in\assignments_c$, let $\posg_{\assignment}$ denote the {\em distribution} over finite paths induced by the strategies assigned to the agents by $\assignment$.
Observe that a complete assignment induces a \emph{set} of paths in $\posg$, while the semantics of \SL should be in terms of an individual induced path for each $\assignment$.
Let $\traj\sim\posg_{\assignment}$ denote a {\em sampled} path $\traj$ from $\posg_{\assignment}$.
Next, we define the semantics of the inner-\LTL formula for $(\traceof{\traj\sim\posg_{\assignment}},0)\models\psi$:
\vspace{-4mm}
\[
\begingroup
\renewcommand{\arraystretch}{1.2}
{
\setlength{\arraycolsep}{3pt}
\begin{array}{@{}ll@{\hspace{.5em}}l@{\hspace{1em}}l@{}}
(\traceof{\traj},i) & \models \var & \text{iff} &
\var \in \traceof{\traj}(i),\\
(\traceof{\traj},i) & \models \lnot\psi & \text{iff} &
(\traceof{\traj},i)\not\models\psi,\\
(\traceof{\traj},i) & \models \psi_1 \OR \psi_2 & \text{iff} &
(\traceof{\traj},i)\models\psi_1
\text{ or }(\traceof{\traj},i)\models\psi_2,\\
(\traceof{\traj},i) & \models \X\psi & \text{iff} &
i+1 < \left|\traceof{\traj}\right|
\text{ and }(\traceof{\traj},i+1)\models\psi,\\
(\traceof{\traj},i) & \models \psi_1 \until \psi_2 & \text{iff} &
\exists i' \geq i \text{ with }i'<\left|\traceof{\traj}\right|
\text{ such that }(\traceof{\traj},i')\models\psi_2\\
&&&
\text{ and }\forall i''\in[i,i'),\,
(\traceof{\traj},i'')\models\psi_1,\\
(\traceof{\traj},i) & \models \psi_1 \then \psi_2 & \text{iff} &
\exists i' \geq i \text{ with }i'<\left|\traceof{\traj}\right|
\text{ such that }(\traceof{\traj},i')\models\psi_2\\
&&&
\text{ and }\exists i''\in[i,i'),\,
(\traceof{\traj},i'')\models\psi_1.
\end{array}
}
\endgroup
\]
\vspace{-4mm}


%
Roughly speaking, given the above semantics, our goal in this paper is to learn functions that map each existentially quantified strategy variable to a strategy in response to the choice of universally quantified strategy variables such that the probability of satisfaction of an \SL specification in POSG is maximized.
We will formally state this optimization problem in the next section.
\vspace{-3mm}

%
%


%% file: 04-problem.tex
\section{Problem Statement}
\label{sec:problem}
\vspace{-2mm}
Let $\varphi = \quant_1 x_1 \ldots \quant_n x_n~(\alpha^1,x_1)\ldots(\alpha^n,x_n).~\psi$ be an \SL sentence, where each $\quant_i \in \setof{\exists,\forall}$ for $1 \leq i \leq n$.
Intuitively, our objective is to identify a tuple of \emph{best-response functions} that select a {\em strategy witness} for each $\quant_i = \exists$, such that the probability of induced paths satisfying $\psi$ is maximized.
\revise{We focus on SL fragments with at most one quantifier alternation (i.e., $\forall^*\exists^*$ and $\exists^*\forall^*$), which capture two mostly common objectives in adversarial RL: learning a response strategy for each opponent strategy and learning a single strategy for the controlled agents that works against all opponent strategies, respectively. These objectives are relevant to games involving two players or two opposing teams, such as Markov soccer \cite{3091574.3091594}, Kuhn poker \cite{kuhn2016simplified}, Predator--Prey, GRF, and SMAC. In addition, the existence of a Nash equilibrium can be expressed in SL using only one quantifier alternation \cite{3635637.3662866}.}

We define $\existsof{} \definedas \setof{i \mid \quant_i=\exists}$ and
$\forallsof{} \definedas \setof{j \mid \quant_j=\forall}$
as the sets of indices of existentially and universally quantified strategy variables, respectively.
Let $\precc \definedas \setof{ j \in \forallsof{} \mid j < \min(\existsof{})}$ denote the set of indices of universal quantifiers preceding the first existentially quantified variable. 
Given a complete strategy assignment $\assignment$, we define a {\em strategy context} 
$\context \definedas \tupleof{\assignment(x_j)}_{j \in \precc}$ 
as a tuple of strategies assigned to all the preceding universally quantified variables.

We now introduce our key objective. 
For each $i\in\existsof{}$, we define a {\em response function} $\bfunc_{x_i}:\strategies^{|\precc|}\rightarrow\strategies$, which maps the strategy context $\context \in \strategies^{|\precc|}$ to a concrete strategy $\pi \in \strategies$.
%
%
Whenever $\varphi$ is of the form $\exists^*\forall^*$, then function $\bfunc_{x_i}$ becomes a constant function for all $i \in \existsof{}$, because $\precc = \{\}$ and $\context=\tupleof{}$.
Finally, among all possible response functions, we denote the {\em best} response function by $\brfunc$; i.e., the one that gives {\em optimal} probability of satisfying $\varphi$. 

\begin{problemstatement}
Given a POSG $\posg$ and \SL sentence $\varphi=\quant_1x_1\ldots\quant_nx_n~(\alpha^1,x_1)\ldots(\alpha^n,x_n).\psi$ with at most one quantifier alternation
our goal is to identify the tuple of best-response functions $\tupleof{\brfunc_{x_i}}_{i\in\existsof{}}$ for all existentially quantified strategy variables $\tupleof{x_i}_{i\in\existsof{}}$ such that:
\vspace{-2mm}
\begin{align}
\label{statement:main}
\tupleof{\brfunc_{x_i}}_{i\in\existsof{}}
\in
\bigg[
\underset{\tupleof{\bfunc_{x_i}}}{\arg\max}
\;
\underset{\assignment\sim \mu(\assignments_c)}{\prob}
\Big[
({\posg, \Pi, {\assignment[x_i,\agent^i\mapsto\bfunc_{x_i}(\context)]}, 0) \models \psi }
\Big]
\bigg]_{i\in\existsof{}}
\end{align}
%
where $\underset{\assignment\sim \mu(\assignments_c)}{\prob}$ denotes the probability of satisfying $\psi$ when $\assignment$ is sampled uniformly from $\assignments_c$.
\end{problemstatement}
\vspace{-2mm}


\vspace{-2mm}
\paragraph{Example.} Consider the POSG in~\cref{fig:posg} and the \SL formula $\varphi_{\text{exp}}$ defined below.
\[
\varphi_{\text{exp}} = \forall x. \exists y. ~(\pirate, x)(\ship,y) ~.~ 
\G (\texttt{pos}_\ship\neq \texttt{pos}_\pirate) ~ \land ~ 
\F(\texttt{pos}_\ship = h )
\]
Suppose that the strategy space is as follows (where $*$ denotes an arbitrary history prefix):
{\small
\begin{align*}
\strategies =
\left\{
\begin{aligned}
\pi_1 &= \{(e,R),(*f,D),(*c,L),(*b,L)\}, ~~
\pi_3 = \{(b,L),(*a,U),(*d,U),(*g,R),(e,S)\},\\ 
\pi_2 &= \{(e,L),(*d,S),(*a,R),(*b,R)\},~~
\pi_4= \{(b,R),(*c,U),(*f,U),(*i,L),(e,S)\}
\end{aligned}
\right\}
\end{align*}}
\noindent 
Consider a response function $\bfunc_y^1 = \{(\tupleof{\pi_1}, \pi_3),  (\tupleof{\pi_2}, \pi_3),  (\tupleof{\pi_3},$ $ \pi_4), (\tupleof{\pi_4}, \pi_3)\}$.
Essentially, $(\tupleof{\pi_1}, \pi_3)$ means that when the \pirate (i.e., $x$) is mapped to the $\pi_1$, then $\bfunc_y^1$ maps \ship $\exists$-quantified variable $y$ to $\pi_3$ (i.e., as the response strategy of \ship against to \pirate).
The probability of satisfying the inner-\LTL formula $\psi_{\text{exp}}$ using $\bfunc_y^1$ is computed by considering the following four cases, each corresponding to a complete assignment:
\textbf{Case 1.} For $(\tupleof{\pi_1}, \pi_3)$,   
a path sampled from the distribution associated with this assignment is $(e,b) \edgeof{R,L} (f,a) \edgeof{D,U} (c,d) \edgeof{L,U} (b,g) \edgeof{L,R} (a,h)$, which satisfies the inner-\LTL formula (i.e., $\models \psi_{\text{exp}}$).
\textbf{Case 2.} For $(\tupleof{\pi_2}, \pi_3)$,    
A path sampled from the distribution associated with this assignment is $(e,b) \edgeof{L,L} (d,a) \edgeof{S,U} (d,d) \edgeof{S,U} (d,g) \edgeof{S,R} (d,h)$, which does not satisfy the inner-\LTL formula (i.e., $\not\models\psi_{\text{exp}}$, because \ship is intercepted \raisebox{-0.15em}{\includegraphics[height=1.2em]{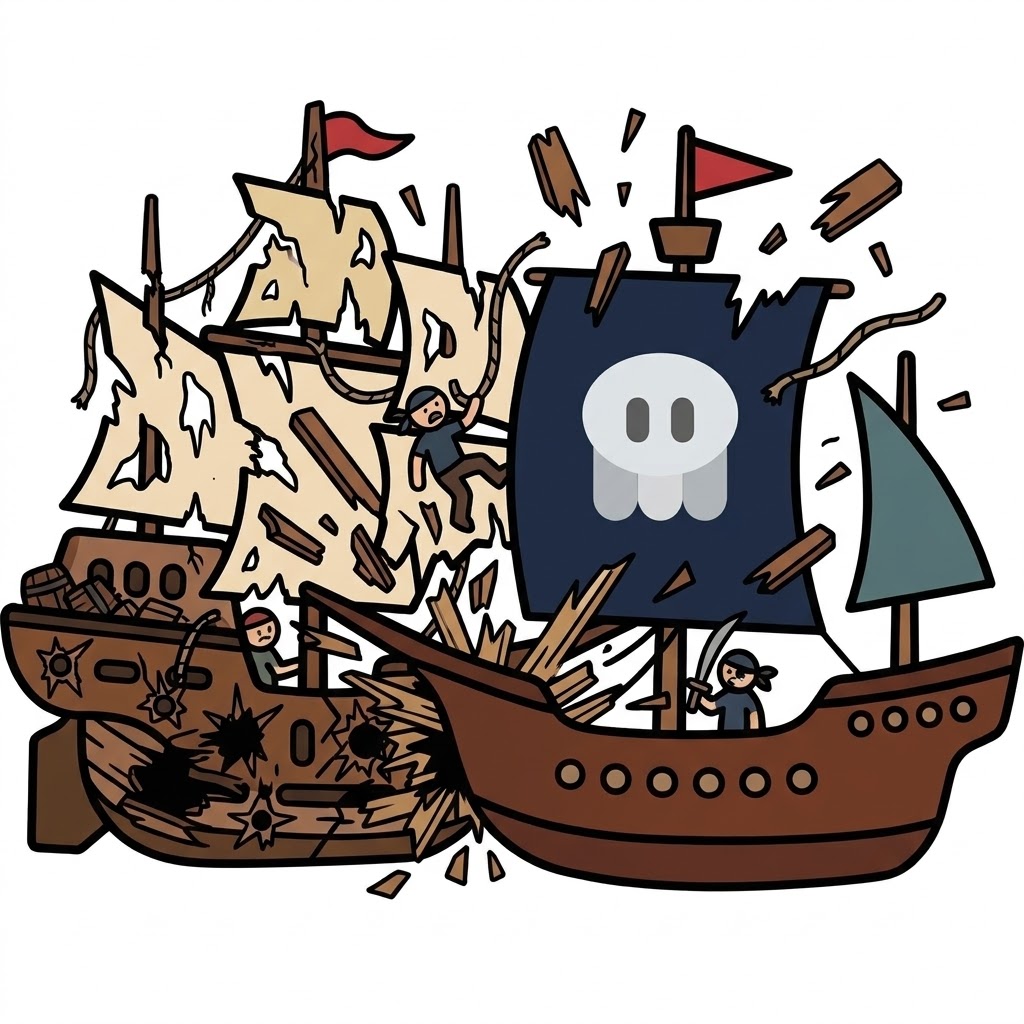}} at cell $d$).
\textbf{Case 3.} Similarly, for $(\tupleof{\pi_3}, \pi_4)$, the induced path satisfies the inner-\LTL formula.   
\textbf{Case 4.} Similarly, for $(\tupleof{\pi_4}, \pi_3)$, the induced path also satisfies the inner-\LTL formula. 
Hence, the probability of satisfying the \SL formula $\varphi_{\text{exp}}$ using $\bfunc_y^1$ is $0.75$ 
.

However, suppose that in the next iteration we identify another response function $\bfunc_y^2 = \{(\tupleof{\pi_1}, \pi_3), (\tupleof{\pi_2}, \pi_4), (\tupleof{\pi_3}, \pi_4), (\tupleof{\pi_4}, \pi_3)\}$. 
When the strategy context is $\tupleof{\pi_2}$, $\bfunc_y^2$ returns $\pi_4$ for \ship, and the induced path {\em satisfies} the inner-\LTL formula.
All remaining I/O in $\bfunc_y^2$ satisfy $\varphi$ with the same analysis as $\bfunc_y^1$. 
Therefore, the probability of satisfying $\varphi_{\text{exp}}$ using $\bfunc_y^2$ increases to $1$ (as it remedies Case 2 which $\bfunc_y^1$ fails).  
Hence, $\bfunc_y^2$ is returned as the best-response function.
\vspace{-2mm}

%% file: 05-algorithm.tex
\section{Algorithmic Details of \method}
\label{sec:algorithm}

In this section, we describe the algorithmic details of \method (see~\Cref{fig:SLICR-method}). We first identify the challenges involved in solving the problem formulated in~\Cref{sec:problem}, and then present the components of \method designed to address each challenge.
%

\vspace{-2mm}
\paragraph{Challenges and Overview of SLICR.} First, recall that our objective in~\Cref{sec:problem} is to find best-response functions that satisfy an \SL specification under its binary semantics. 
We transform this temporal satisfaction problem into a robustness optimization problem using the quantitative semantics (\Cref{sec:quant}).
Second, universally quantified strategy variables may be assigned irrational strategies, resulting in noisy strategy contexts that can mislead the learning agents and degrade their performance. 
Therefore, SLICR identifies these strategies and prevents the corresponding experiences from affecting the learning process through a shielding mechanism (\Cref{sec:shield}).
Third, agents bound by existentially quantified strategy variables must respond to a wide range of strategies assigned to agents bound by preceding universally quantified variables.
%
%
Thus, each response function takes as input a strategy context, which is a tuple of strategies.
Since each strategy is itself a function, this tuple cannot be directly provided as input to a learnable response function.
\method infers an embedding of the strategy context from the limited observations available to the agent (\Cref{sec:learning}).

\vspace{-2mm}


\subsection{Quantitative Semantics of Inner-LTL Formulas}
\label{sec:quant}
\vspace{-2mm}
To obtain an optimization objective for \SL satisfaction, we equip the inner-\LTL formula with quantitative semantics~\citep{li2017reinforcement}. 
Given a complete assignment $\assignment \in \assignments_c$ and those selected by $\bfunc$ for agents bound to existentially quantified variables, we sample a path with length $T$ from the POSG $\posg$ and quantitatively evaluate the inner-\LTL formula $\psi$ over this path.
Let $\mathbb{R}$ denote the set of real numbers, $\Psi$ the set of all \LTL formulas, and $\predicatefunc:2^{\AP}\rightarrow\mathbb{R}$ a \emph{valuation function} that assigns a real value to each set of atomic propositions.
For a state $\state \in \states$ in $\posg$, quantitative semantics are defined by predicates of the form $\predicatefuncof{\labelsof{\state}}$ (see App.~\ref{app:fullquant} for full semantics).
The robustness function $\rbvalue:\tracesof{\trajs^\finite}\times\Psi\rightarrow\mathbb{R}$ assigns a real-valued \emph{robustness value} to each finite trace and \LTL formula, bounded below by $\rbvalue_{\mathit{min}}$ and above by $\rbvalue_{\mathit{max}}$.
Formally, given an inner-\LTL formula $\psi$ and a POSG $\posg$, we reformulate the problem statement introduced in~\cref{sec:problem} as follows:
\vspace{-2mm}
\begin{align}
\tupleof{\brfunc_{x_i}}_{i\in\existsof{}} \in
    \bigg[
    \underset{
        \tupleof{\bfunc_{x_i}}
    }
    {{\arg\max}}
~~ \underset{\assignment\sim \mu(\assignments_c)}{\prob}
    \Big[ 
    \rbvalueof{\traceoff{(\traj_{[0:\horizon]} \sim \posg_{\assignment[{x_i},\agent^i\mapsto\bfunc_{x_i}(\context)]})}, \psi} \converge \rbvalue_{\mathit{max}}
    \Big]
    \bigg]_{i \in \existsof{}}
    \label{eq:opt}
\end{align}
\vspace{-3mm}

where $\converge$ denotes convergence. 
That is, the tuple of best-response functions $\tupleof{\brfunc_{x_i}}_{i\in\existsof{}}$ maximizes the probability that the robustness of a path sampled from $G$ under the complete assignment $\assignment$ converges to $\rbvalue_{\mathit{max}}$, thereby optimizing the satisfiability of the inner-\LTL formula $\psi$.
%
\begin{theorembox}
\begin{lemma}
\label{lem:optimality-best-response}
\LEMMABS
\end{lemma}
\end{theorembox}


\subsection{Shielding from Noisy Strategies}
\label{sec:shield}
\vspace{-2mm}
Since universally quantified variables may be assigned any strategy from $\strategies$, the resulting strategy contexts may contain \emph{irrational strategies} that hinder the learning of $\tupleof{\bfunc_{x_i}}_{i \in \existsof{}}$.
We define an irrational strategy as an observation-independent strategy that selects actions uniformly at random. 
To mitigate this issue, before the main learning phase, we train a shield $\shield:\actobshist^\finite\rightarrow\setof{\warn,\accept}$ that classifies the strategy context from the history of an agent bound to an existentially quantified variable as either rational ($\accept$) or irrational ($\warn$).
Though not a function, we use the inverse of $\shield$ to denote the set of observation histories associated with rational strategies; i.e., $\shield^{-1}(\accept)$.
Formally, $\shield$ is trained on $m$-step histories $h_{\agent^i_{[0:m]}}\in\actobshist^\finite$, for each $i\in\existsof{}$, collected under rational and irrational strategies assigned to universally quantified variables.
Experiences labeled $\warn$ are excluded from training, while those labeled $\accept$ are included.
During the main learning phase, the shield is used only for inference. 
Further details on shield are provided in Appendix~\ref{app:shield}.
\vspace{-2mm}

\subsection{Learning Best-Response Functions}
\label{sec:learning}
\vspace{-2mm}
We now address the optimization problem in~\cref{eq:opt}. 
%
Our learning framework follows the \emph{Centralized Training with Decentralized Execution (CTDE)} paradigm~\citep{oliehoek2016concise}. 
Conceptually, during centralized training, agents bound to existentially quantified variables have access to observations and actions of agents bound to preceding universally quantified variables. 
During execution, each agent has access only to its local information.
\vspace{-2mm}


%

\paragraph{Robustness Values as Reward Signals.} 
To solve~\Cref{eq:opt}, we derive reward signals that guide the learning of response functions.
Given a $\assignment \in \assignments_c$, the reward for each $\agent^i$, where $i \in \existsof{}$, at step $k$ is computed as $\rbvalueof{\traceoff{(\traj \sim \posg_{\assignment})_{[0:k+1]}}, \psi}$.
Given a discount factor $\discount\in[0,1)$, we learn best-response functions that maximize the expected cumulative discounted robustness:
\vspace{-1mm}
\begin{equation}
\label{eq:learning}
\tupleof{\brfunc_{x_i}}_{i\in\existsof{}}
\in
\bigg[
\underset{\tupleof{\bfunc_{x_i}}}{\arg\max}
\;
\underset{
\substack{
\assignment\sim \mu(\assignments_c)\\
\traj\sim
\posg_{\assignment[
x_i,\agent^i\mapsto\bfunc_{x_i}(\context)
]}
}
}{\expected}
\bigg[
\sum_{k=0}^{T-1}
\discount^k
\rbvalueof{
\traceoff{\traj_{[0:k+1]}},
\psi
}
\bigg]
\bigg]_{i\in\existsof{}}
\end{equation}
\vspace{-3mm}
\begin{theorembox}
\begin{lemma}
\label{lem:optimality-robustness}
\LEMMARB
\end{lemma}
\end{theorembox}

Note that, in this setting, reward signals are derived from the inner-\LTL formula, so all existentially quantified agents receive the same reward. 
However, more generally, a POSG may assign distinct rewards to individual agents.
We can capture this setting by decomposing $\psi$ into agent-specific subformulas, $\psi = \bigwedge_{i\in\existsof{}} \psi_i$, where each $\psi_i$ represents an objective or constraint associated with an agent $\agent^i$. 
The reward for agent $\agent^i$ at step $k$ is then individually computed as $\rbvalueof{\traceoff{(\traj \sim \posg_{\assignment})_{[0:k+1]}}, \psi_i}$.
For simplicity, we present the remainder of our algorithm using a single full inner-\LTL formula $\psi$.

\vspace{-3mm}
\paragraph{Representing Strategy Context.}
As shown in~\cref{eq:learning}, for each $i \in \existsof{}$, the response function $\bfunc_{x_i}$ takes $\context$ as an input.
This raises two difficulties.
First, $\context \in \strategies^{|\precc|}$ is a $|\precc|$-tuple of strategies, each of which is itself a function. Therefore, providing this tuple as input to a learnable response function requires a suitable finite-dimensional representation.
Second, due to the partial observability of the POSG, each agent $\agent^i$, where $i \in \existsof{}$, must infer this representation from its local history $h_{\agent^i} \in \actobshist^\finite$.
We associate the strategy-context space $\strategies^{|\precc|}$ with a latent context space $\stratspace\subseteq\mathbb{R}^{d}$, where $d\in\mathbb{Z}_{>0}$ denotes the dimension of the latent context embedding.
Let $\latent\in\stratspace$ denote a latent embedding. We call $\latent$ an \emph{information-preserving latent context embedding} of $\context$ (denoted $\latent_\context$), if it represents $\context$ without information loss, meaning that $\context$ can be uniquely recovered from $\latent$.

We are inspired by the agent-modeling approach proposed in~\citep{NEURIPS2021_a03caec5}, which uses an encoder-decoder architecture to learn embeddings of other agents from the controlled agent's local information under partial observability.
We define {\em encoder} $\enc_w:\actobshist^\finite\rightarrow\stratspace$, parameterized by $w$, that maps the history $h_{\agent^i}\in\actobshist^\finite$ of any agent $\agent^i$, where $i\in\existsof{}$, to a latent embedding $\latent\in\stratspace$.
Next, we define two {\em decoders}: an {\em action-reconstruction} decoder $\deca_{wa}:\stratspace\rightarrow \Delta(\actions)^{|\precc|}$, and an {\em observation-reconstruction} $\deco_{wo}:\stratspace\rightarrow\ObsSys^{|\precc|}$, parameterized by $wa$ and $wo$, respectively.
These decoders map a latent embedding $\latent\in\stratspace$ to the reconstructed actions and observations of agents bound to the preceding universally quantified variables, respectively.
In addition,
we use the latent embedding to predict their actions over the next $L$-steps. 
This prediction encourages the latent embedding to capture information about their future behavior. 
Formally, we define a {\em future-action predictor} $\future_\mathit{wf}:\stratspace\rightarrow \Delta(\actions)^{|\precc|\times L}$,  parameterized by $\mathit{wf}$, where $L\in\mathbb{Z}_{>0}$ denotes the prediction window.
Further details on computing the latent embedding are provided in~\cref{app:detencdec}.

\vspace{-3mm}

\paragraph{Learning Response Functions.}

Recall that, as defined in~\cref{sec:prelim,sec:problem}, each complete assignment $\assignment\in\assignments_c$ induces a distribution $\posg_{\assignment}$ over finite paths, from which a path $\traj\sim\posg_{\assignment}$ is sampled.
For each $i\in\existsof{}$, we seek a $\bfunc_{x_i}$ that maximizes the objective in~\cref{eq:learning}.
Given a $\context$, the $\bfunc_{x_i}(\context)$ returns a strategy for variable $x_i$ and agent $\agent^i$, where each strategy is a mapping $\pi:\actobshist^\finite\rightarrow\Delta(\actions)$.
Hence, assigning a strategy $\pi$ alone does not determine its action distribution during sampling, as $\pi$ depends on the history observed by $\agent^i$.
%
As described earlier, we use the latent embedding $\latent_\context$ to represent the strategy context $\context$.
%
%
Hence, to explicitly represent how the assigned strategies generate actions during learning, we define its equivalent action-level response function $\bfunca_{x_i}:\stratspace\times\actobshist^\finite\rightarrow\Delta(\actions)$, which maps a latent embedding and a history to a distribution over the actions.

Now, let $\tupleof{\action_i}_{i\in I}$ and $\tupleof{\action_j}_{j\in J}$ be two action tuples, where $I \cup J = \setof{1,\ldots,n}$ and $I \cap J = \emptyset$. 
To combine these tuples while preserving the order according to their indices, we use the operator $\orderedunion$, which returns a single tuple ordered by index.
Given a complete assignment $\assignment\in \assignments_c$ and a tuple of action-level response functions $\tupleof{\bfunca_{x_i}}_{i \in \existsof{}}$, a joint action $\actionj$ at step $k$ is defined as: 
\begin{equation*}
    \actionj_k = \tupleof{\action_{i_k} \sim \bfunca_{x_i}(.|\enc_w(h_{\agent^i_{[0:k]}})_\context \mid_{\shield^{-1}(\accept)}, h_{\agent^i_{[0:k]}})\mid_{\shield^{-1}(\accept)}}_{i \in \existsof{}} \orderedunion \tupleof{\action_{j_k} \sim \assignment(x_j)(.|h_{\agent^j_{[0:k]}})}_{j \in \forallsof{}}
\end{equation*}
%
Notice that we are restricting the inputs of $\enc_{w}$ and $\bfunca_{x_i}$ by observation histories that yield a rational strategy by the $\shield$ during training (and not execution).
For simplicity, we write it as $\actionj_k \sim \tupleof{\bfunca_{x_i}(.|\latent_\context, h_{\agent^i_{[0:k]}})}_{i \in \existsof{}} \orderedunion \tupleof{\assignment(x_j)(.|h_{\agent^j_{[0:k]}})}_{j \in \forallsof{}}$, where $\lambda_\context = \enc_w(h_{\agent^i_{[0:k]}})_\context \mid_{\shield^{-1}(\accept)}$.
Accordingly, the next state is computed as $\state_{k+1} \sim \trans(.|\state_k, \actionj_k)$, where $s_0 \sim \init$.
We now update the learning objective introduced in~\cref{eq:learning} to identify the action-level best-response functions, formally:
\begin{equation}
\label{eq:learning-update}
\tupleof{\brfunca_{x_i}}_{i\in\existsof{}}
\in
\bigg[
~~
\underset{\tupleof{\bfunca_{x_i}}}{\arg\max}
\;
\hspace{-15mm}
\underset{
\substack{
\assignment\sim \mu(\assignments_c)\\
s_0 \sim \init \\ 
\actionj_k\sim
\tupleof{\bfunca_{x_i}(.|\latent_\context, h_{\agent^i_{[0:k]}} \hspace{-2mm})} \orderedunion \tupleof{\assignment(x_j)(.|h_{\agent^j_{[0:k]}} \hspace{-2mm})}_{j \in \forallsof{}}\\
s_{k+1} \sim \trans(.|s_k, \actionj_k)
}
}{\expected}
\hspace{-15mm}
\bigg[
\sum_{k=0}^{T-1}
\discount^k
\rbvalueof{
\traceoff{\state_0 \edgeof{\actionj_0} \cdots \edgeof{\actionj_{k}} \state_{k+1} },
\psi
}
\bigg]
\bigg]_{i\in\existsof{}}
\end{equation}

\begin{theorembox}
\begin{theorem}
\THEOREMFINAL 
\label{theory:final}
\end{theorem}
\end{theorembox}

%% file: 06-experiments.tex
\section{Experiments and Results}
\label{sec:eval}
\vspace{-3mm}

\revise{Our experiments target the maximization problem in \cref{eq:learning-update}. Its connection to the objective in \cref{statement:main} is established theoretically through \cref{theory:final} and Lemmas~\ref{lem:optimality-robustness} and~\ref{lem:optimality-best-response}, with the resulting guarantees conditional on their stated assumptions.}
\paragraph{Implementation Pipeline.} 
We first construct $\strategies$ using off-the-shelf MARL algorithms, including IPPO and MAPPO~\citep{NEURIPS2022_9c1535a0}, VDN~\citep{sunehag2017valuedecompositionnetworkscooperativemultiagent},QMIX~\citep{pmlr-v80-rashid18a}, and QTRAN~\citep{son2019qtranlearningfactorizetransformation}, together with irrational strategies that select actions uniformly at random.
Let $\strategies^{\text{Train}}$ and $\strategies^{\text{Test}}$ denote the strategy sets used during training and evaluation, respectively. We evaluate performance in two settings: (1) \texttt{Seen}, where $\strategies^{\text{Train}}=\strategies^{\text{Test}}$, and (2) \texttt{Unseen}, where $\strategies^{\text{Train}}\cap\strategies^{\text{Test}}=\emptyset$.
During training phase, we use the quantitative semantics (\cref{sec:quant}) to compute robustness-based rewards from the \SL formula.
For each $i\in\existsof{}$, we then train the response function using $h_{\agent^i}$ and $\latent$ with A2C, PPO~\citep{schulman2017proximalpolicyoptimizationalgorithms} or IPPO, while the pretrained $\shield$ excludes experiences associated with irrational strategies.
We report results over ten independent runs with different random seeds.
Further experimental and implementation details are provided in~\Cref{app:exp}.

%
%

\paragraph{Benchmarks.} 
We evaluate \method on three benchmarks.
(1) \textbf{Predator and Prey (PP)}~\citep{NIPS2017_68a97503}. We use the implementation provided by~\citep{NEURIPS2021_a03caec5}. 
PP consists of one existentially quantified prey that must be captured by exactly one of three universally quantified predators. We adopt the strategy sets from~\citep{papoudakis2021benchmarking} and develop novel, diverse IPPO, MAPPO, and irrational predator strategies.
(2) \textbf{Google Research Football (GRF)}~\citep{DBLP:conf/aaai/KurachRSZBERVMB20}. 
We evaluate two scenarios. 
First, in ``Run to Score'' (R2S), a player (bound to an existentially quantified variable) starts at the midfield and must score against a goalkeeper while five opposing players (bound to universally quantified variables) are chasing from behind. 
Second, in ``Run to Score with two Defenders'' (R2S-DEF), inspired by~\citep{odrowazsypniewski2026generalizedintentionmodelingmultiagent}, two of the opposing players are positioned between the existentially quantified player and the goal, while the remaining three are chasing from behind.
\revise{Third, in ``Run to Score with four Defenders'' (R2S-4DEF), where four of the opposing players are positioned between the existentially quantified player and the goal, while the one remaining is chasing from behind.}
To construct a diverse strategy set, we train IPPO, MAPPO, \revise{and built-in AI opponent} teams with different tactical behaviors.
(3) \textbf{SMAC-HARD}~\citep{deng2024smachardenablingmixedopponent}. We evaluated our extension with pretrained MARL opponent populations on three maps: \texttt{3m}, \texttt{2s3z}, and \texttt{3s\_vs\_3z}. 
%
The objective of the existentially quantified agents is to eliminate all universally quantified agents. 
To construct a diverse strategies set, we use VDN, QMIX, QTRAN, and built-in AI opponents.
Further details on the environment settings, \SL formulas, and opponent strategies are provided in App.~\ref{app:bench}.
\vspace{-1mm}

\begin{figure*}[t]
    \centering

\vspace{-2mm}
    
    \begin{subfigure}[t]{\textwidth}
        \centering
        \includegraphics[width=0.8\linewidth]{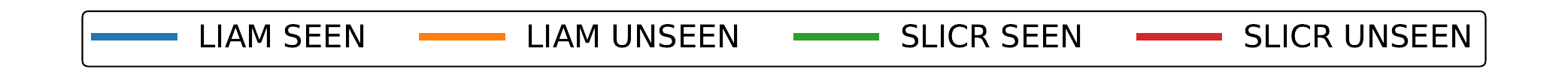}
    \end{subfigure}
    \begin{subfigure}[t]{0.32\textwidth}
        \centering
        \includegraphics[width=\linewidth]{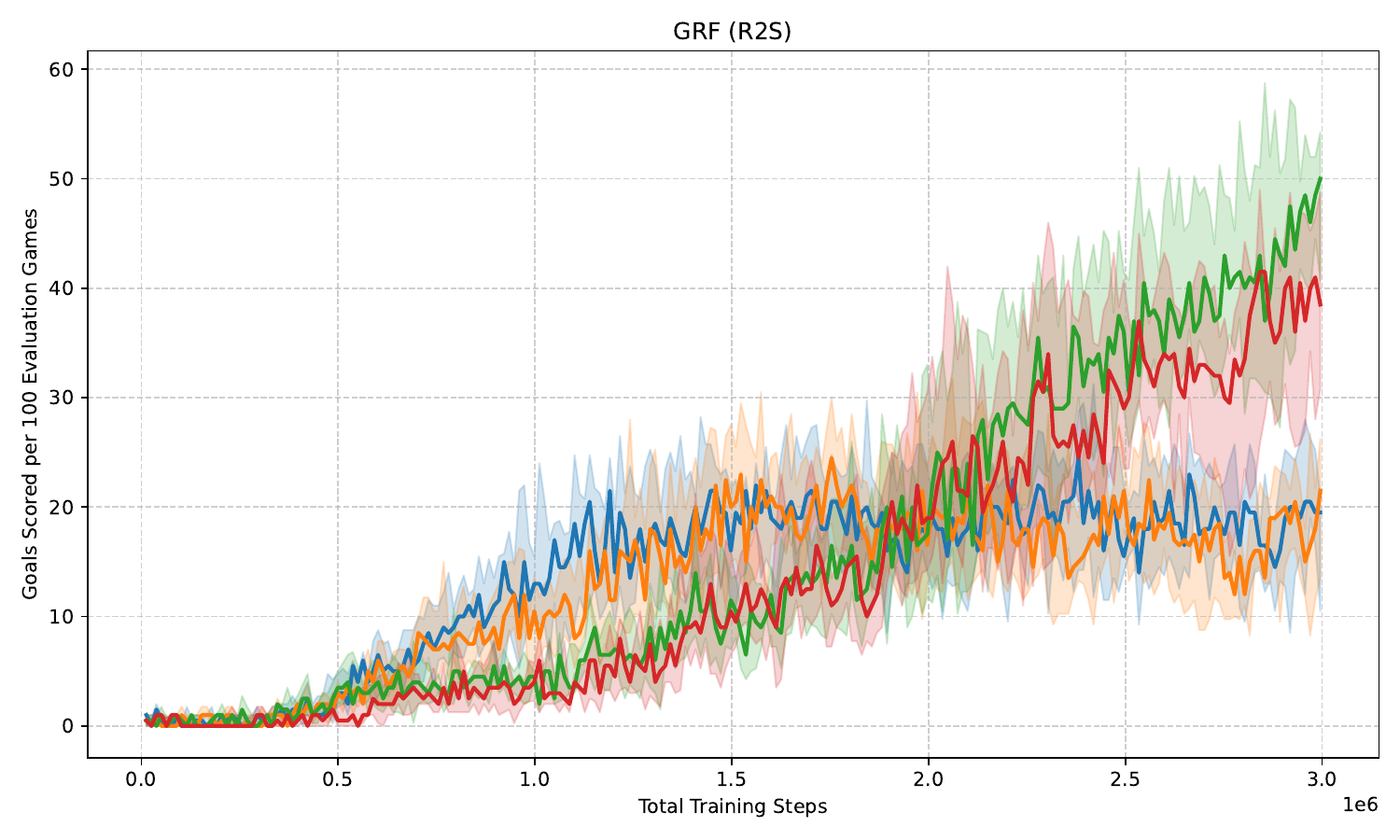}
                \vspace*{-4mm}
        \caption{\small GRF (R2S)}
        \label{fig:GRF(R2S)} 
    \end{subfigure}\hfill
    \begin{subfigure}[t]{0.32\textwidth}
        \centering
        \includegraphics[width=\linewidth]{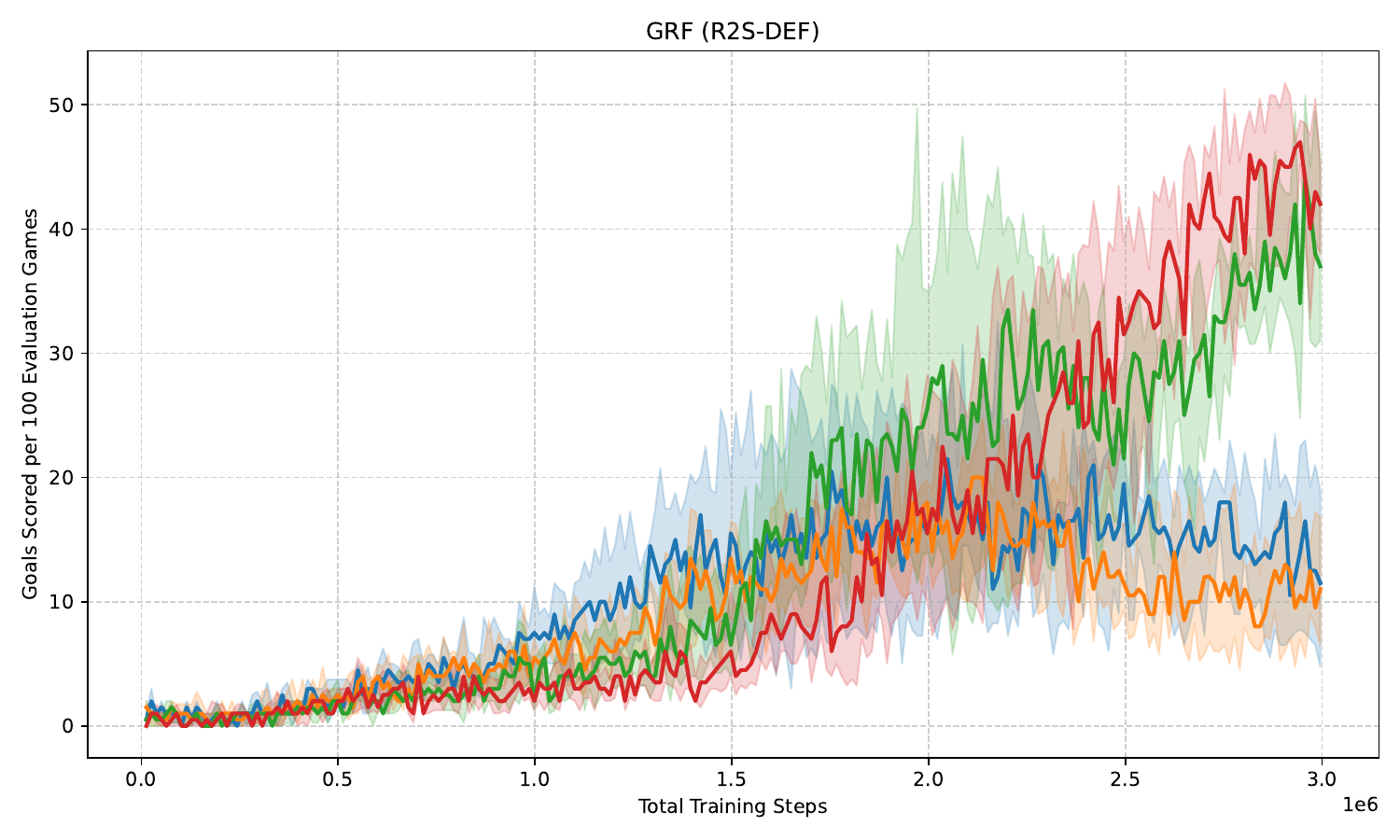}
                \vspace*{-4mm}
        \caption{\small GRF (R2S-DEF)}
        \label{fig:GRF(R2S-HARD)}
    \end{subfigure}\hfill
    \begin{subfigure}[t]{0.32\textwidth}
        \centering
        \includegraphics[width=\linewidth]{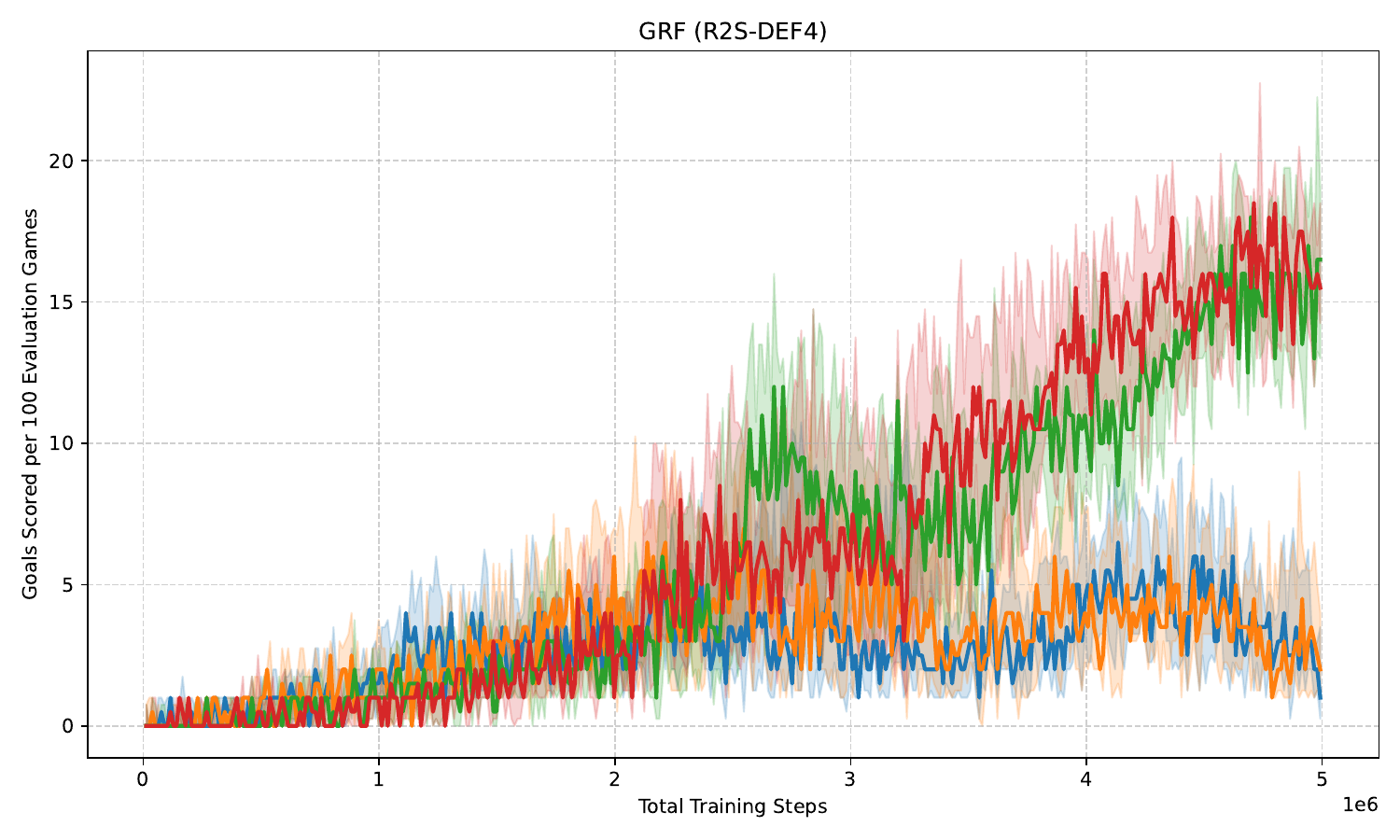}
                \vspace*{-4mm}
        \caption{\small GRF (R2S-4DEF)}
        \label{fig:GRF(R2S-4def)}
    \end{subfigure}
        \vspace{-1mm}
    \begin{subfigure}[t]{0.32\textwidth}
        \centering
        \includegraphics[width=\linewidth]{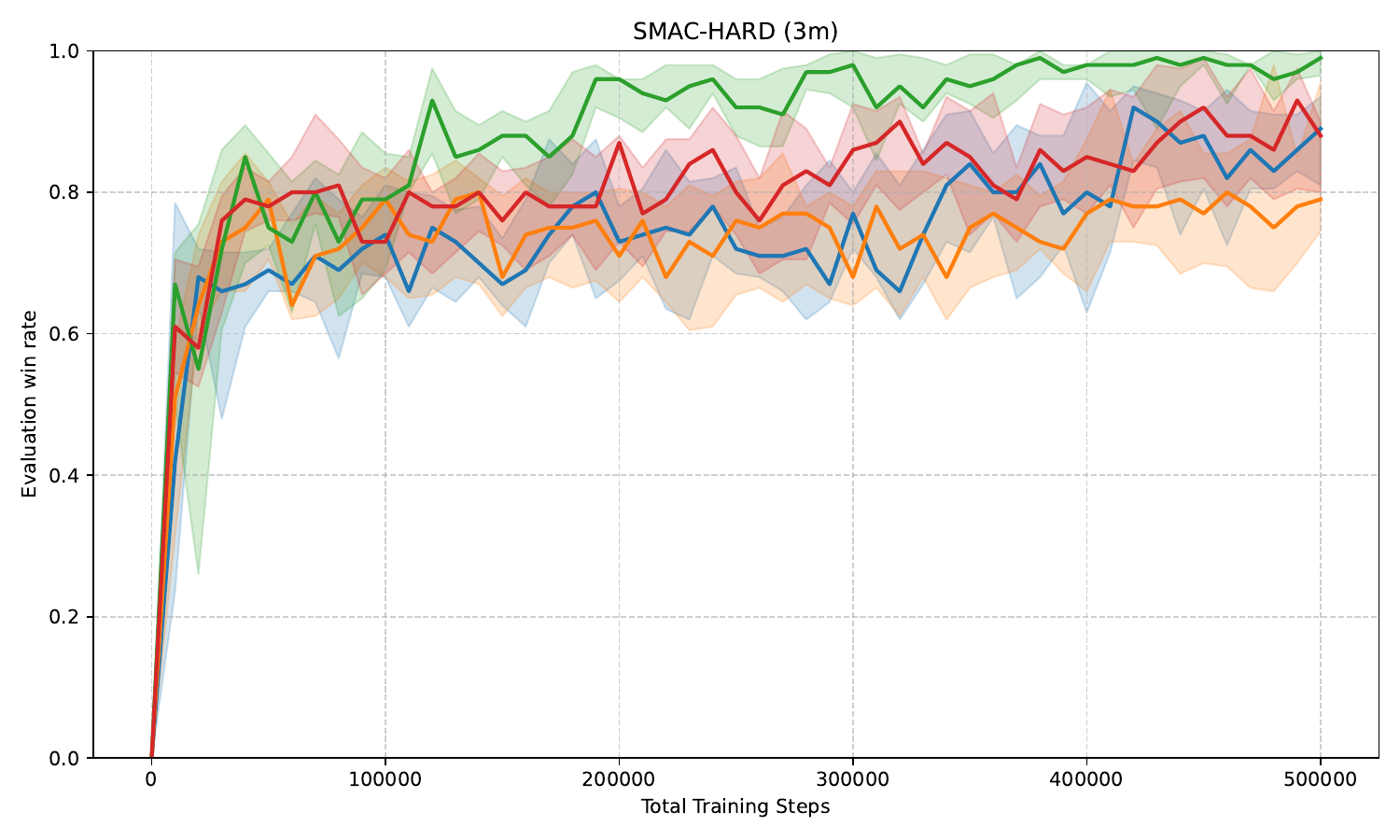}
                \vspace*{-4mm}
        \caption{\small SMAC-HARD (3m--PPO)}
        \label{fig:3m}
    \end{subfigure}
    \hfill
    \begin{subfigure}[t]{0.32\textwidth}
        \centering
        \includegraphics[width=\linewidth]{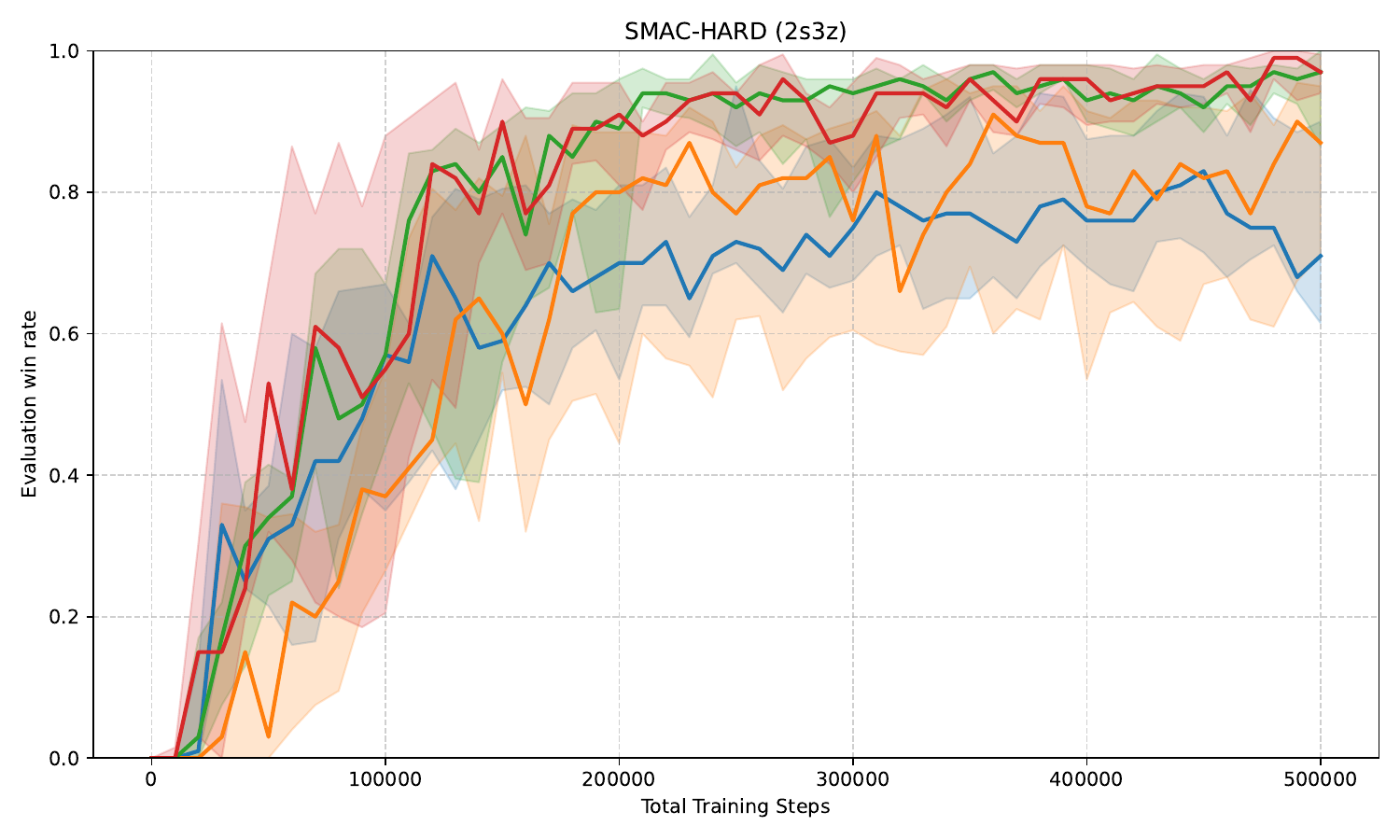}
                \vspace*{-4mm}
        \caption{\small SMAC-HARD (2s3z--PPO)}
        \label{fig:2s3z}
    \end{subfigure}
    \hfill
    \begin{subfigure}[t]{0.32\textwidth}
        \centering
        \includegraphics[width=\linewidth]{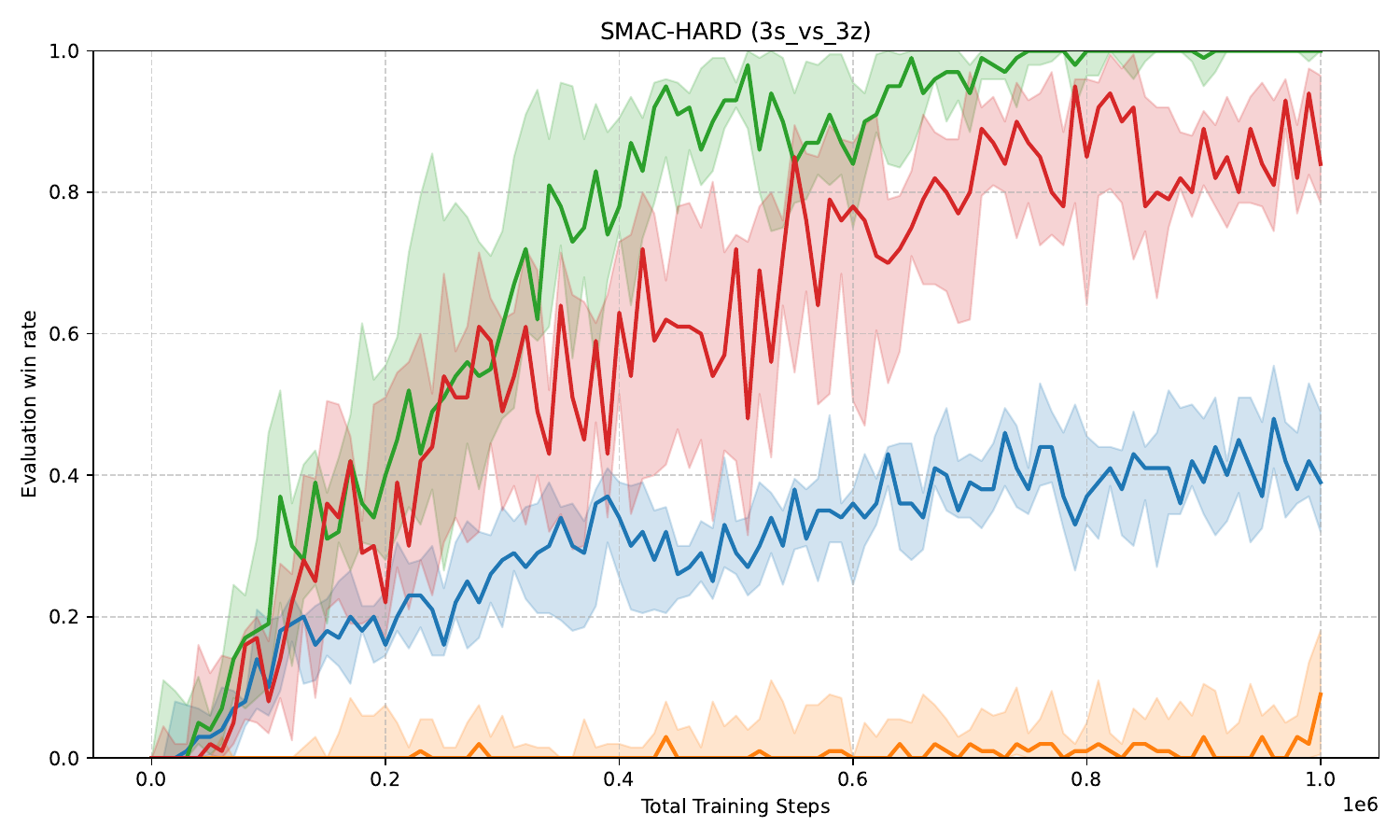}
                \vspace*{-4mm}
        \caption{\small SMAC-HARD (3s\_vs\_3z--PPO)}
        \label{fig:3svs3z}
    \end{subfigure}
    \begin{subfigure}[t]{0.32\textwidth}
        \centering
        \includegraphics[width=\linewidth]{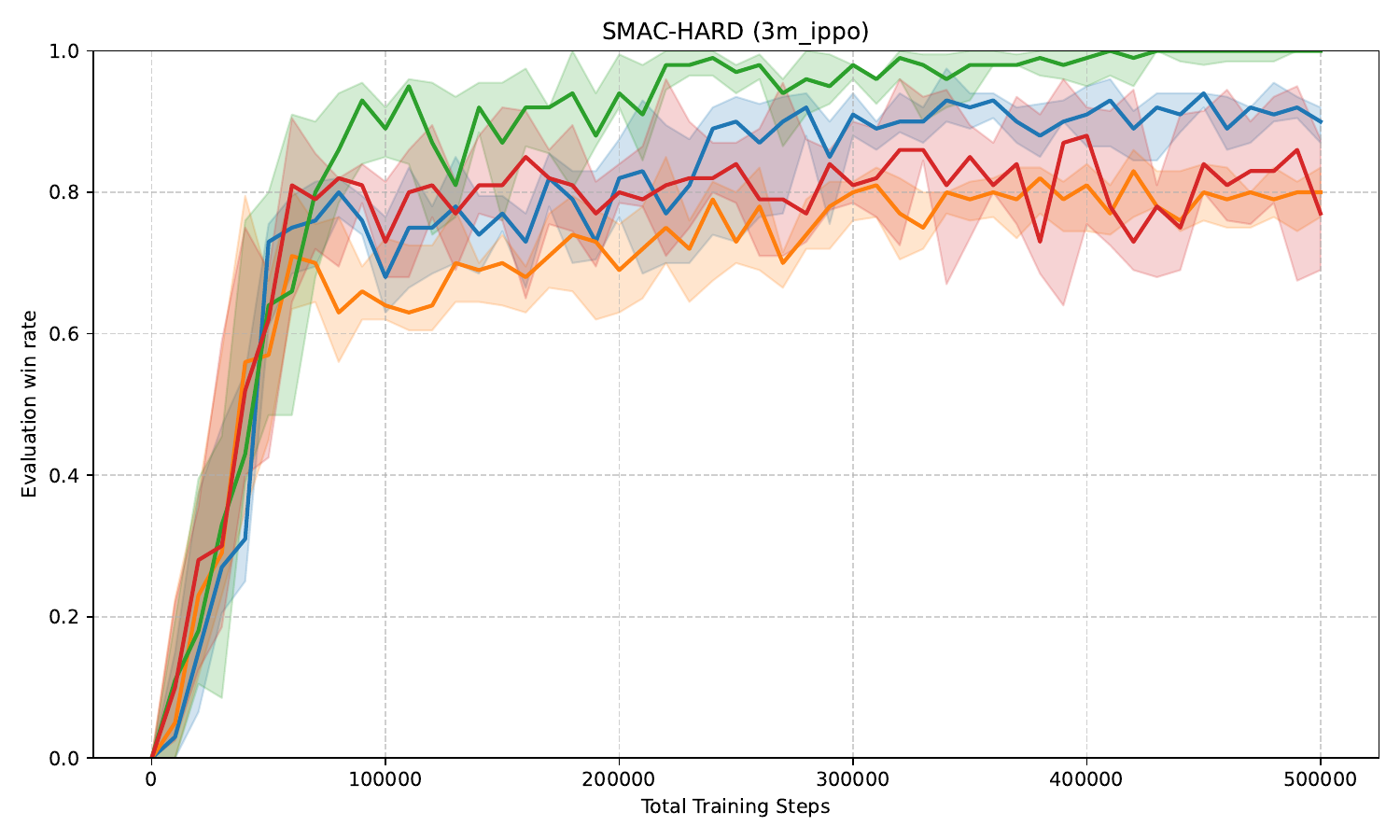}
                \vspace*{-4mm}
        \caption{\small SMAC-HARD (3m--IPPO)}
        \label{fig:3mippo}
    \end{subfigure}
    \hfill
    \begin{subfigure}[t]{0.32\textwidth}
        \centering
        \includegraphics[width=\linewidth]{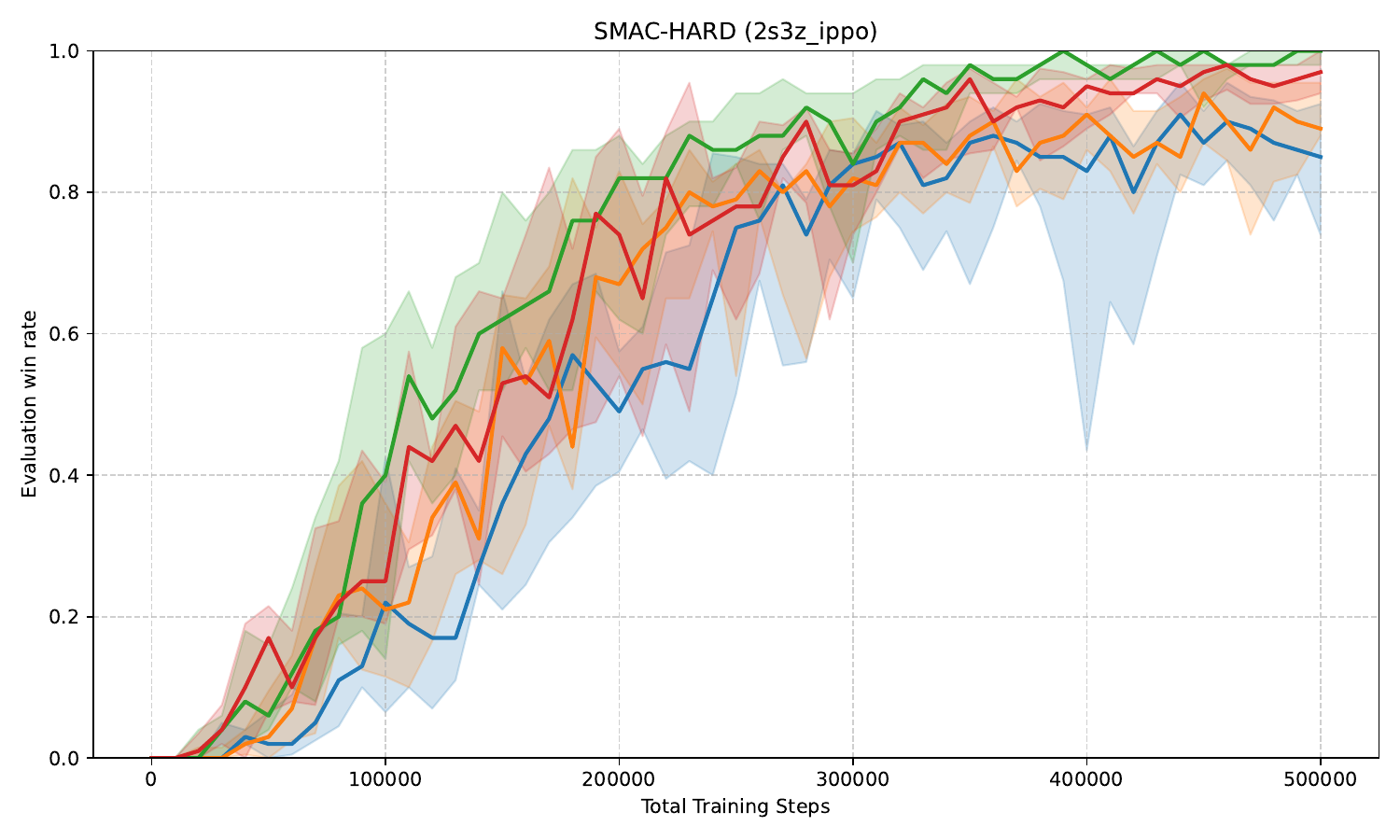}
                \vspace*{-4mm}
        \caption{\small SMAC-HARD (2s3z--IPPO)}
        \label{fig:2s3zippo}
    \end{subfigure}
    \hfill
    \begin{subfigure}[t]{0.32\textwidth}
        \centering
        \includegraphics[width=\linewidth]{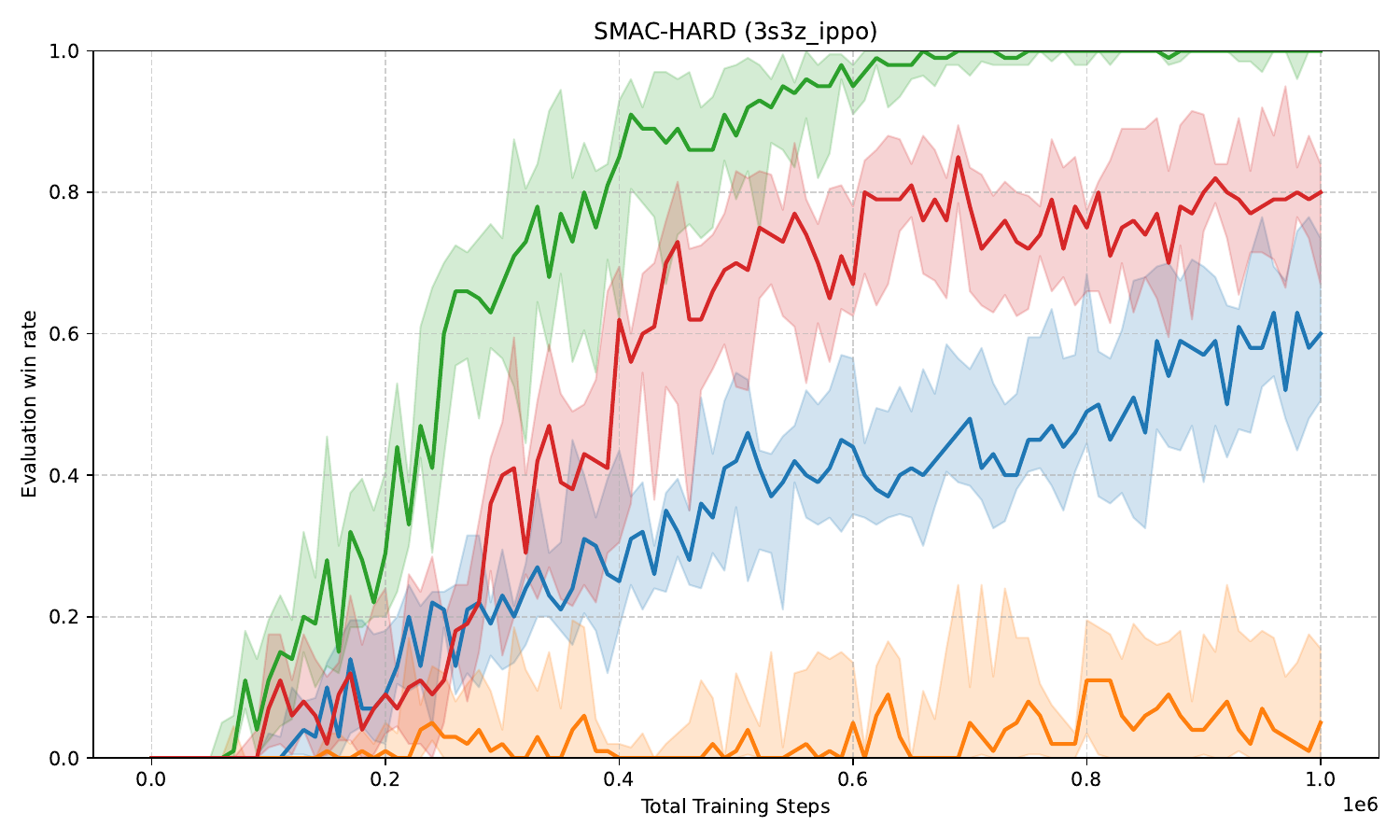}
                \vspace*{-4mm}
        \caption{ \small SMAC-HARD (3s\_vs\_3z--IPPO)}
        \label{fig:3svs3zippo}
    \end{subfigure}
    
    \vspace{-1mm}
    \caption{\small Learning curves for \method and LIAM, showing the median and 25th–75th percentiles.}
        \vspace{-1mm}
    \label{fig:main-base-slicr}
\end{figure*}

\begin{figure}[t!]
\centering
\begin{minipage}[t]{0.33\textwidth}
    \vspace{0pt}
    \centering
    \includegraphics[width=\linewidth]{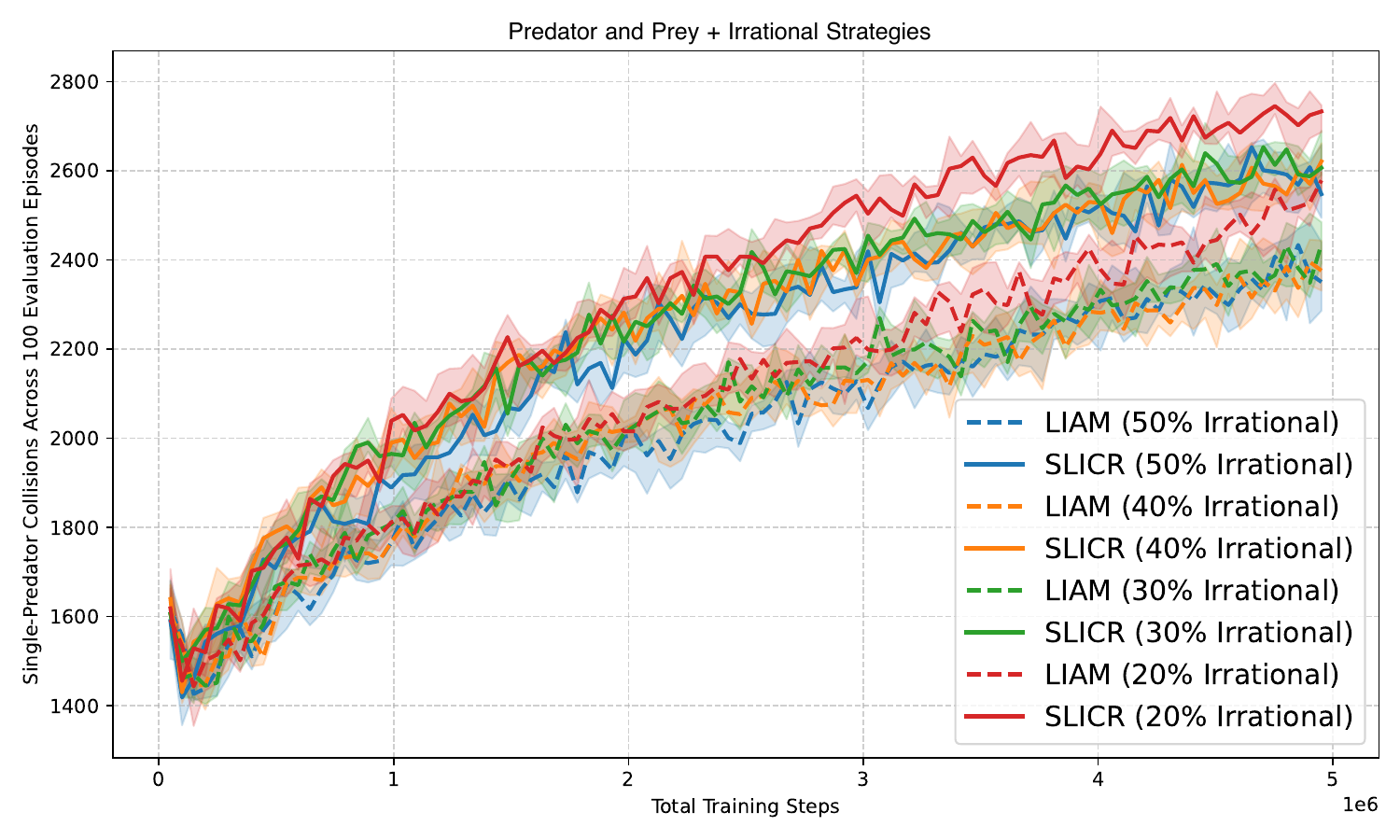}
        \vspace{-5.8mm}
    \captionof{figure}{\small Irrational strategies(PP).}
    \label{fig:noise}
\end{minipage}
\begin{minipage}[t]{0.66\textwidth}
    \vspace{0pt}
    \centering
    \begin{subfigure}[t]{0.50\linewidth}
        \vspace{0pt}
        \includegraphics[width=\linewidth]{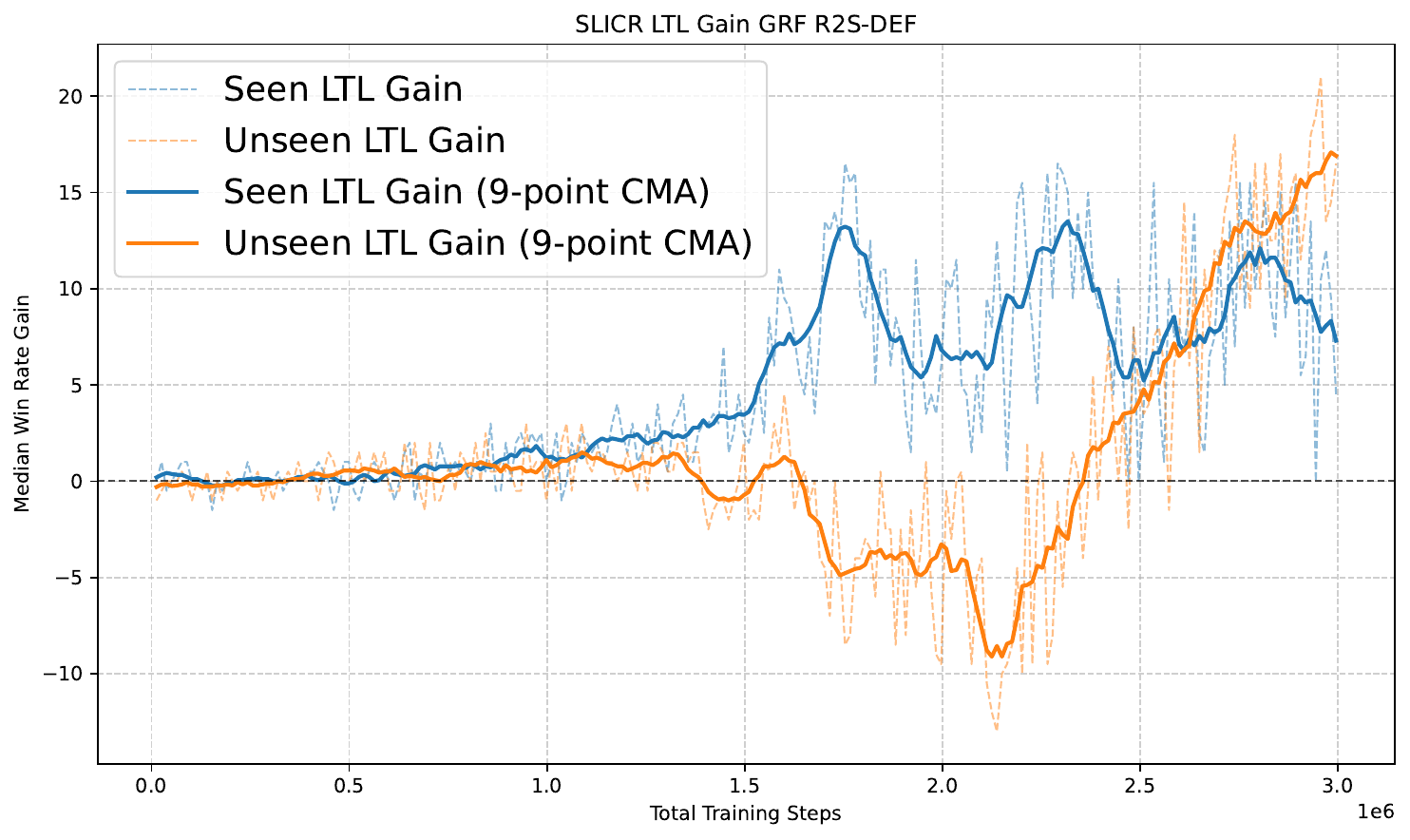}
    \end{subfigure}\hfill
    \begin{subfigure}[t]{0.50\linewidth}
        \vspace{0pt}
        \includegraphics[width=\linewidth]{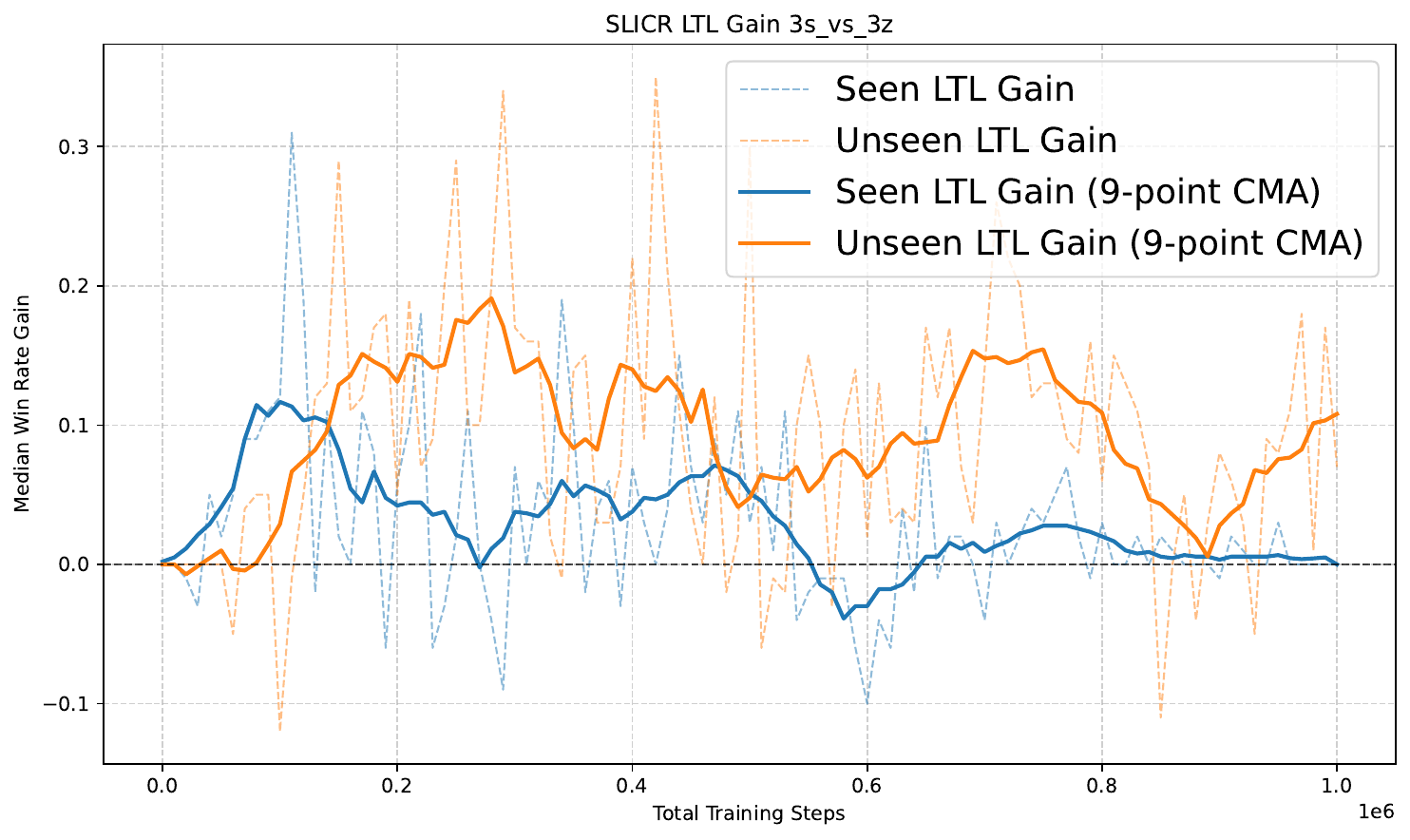}
    \end{subfigure}
    \vspace{-2mm}
    \captionof{figure}{\small Effect of inner-\LTL on R2S-DEF (left), \texttt{3s\_vs\_3z} (right).}
    \label{fig:gain}
\end{minipage}
\vspace{-4mm}
\end{figure}

\paragraph{Experimental Organization.}
We assess \method along four dimensions:
(1) Comparing \method with LIAM~\citep{NEURIPS2021_a03caec5}, an opponent-modeling baseline for partially observable environments.
(2) Comparing the $\forall^*\exists^*$ and $\exists^*\forall^*$ fragments of the target \SL formulas in terms of effectiveness and parameter efficiency.
(3) Comparing \method and LIAM in the presence of irrational strategies.
(4) Investigating how the inner-\LTL formula guides and improves learning.

\paragraph{(1) \method vs. LIAM.}
\begin{wrapfigure}{r}{0.35\textwidth}
\vspace{-7mm}
    \centering
    \includegraphics[width=0.35\textwidth]{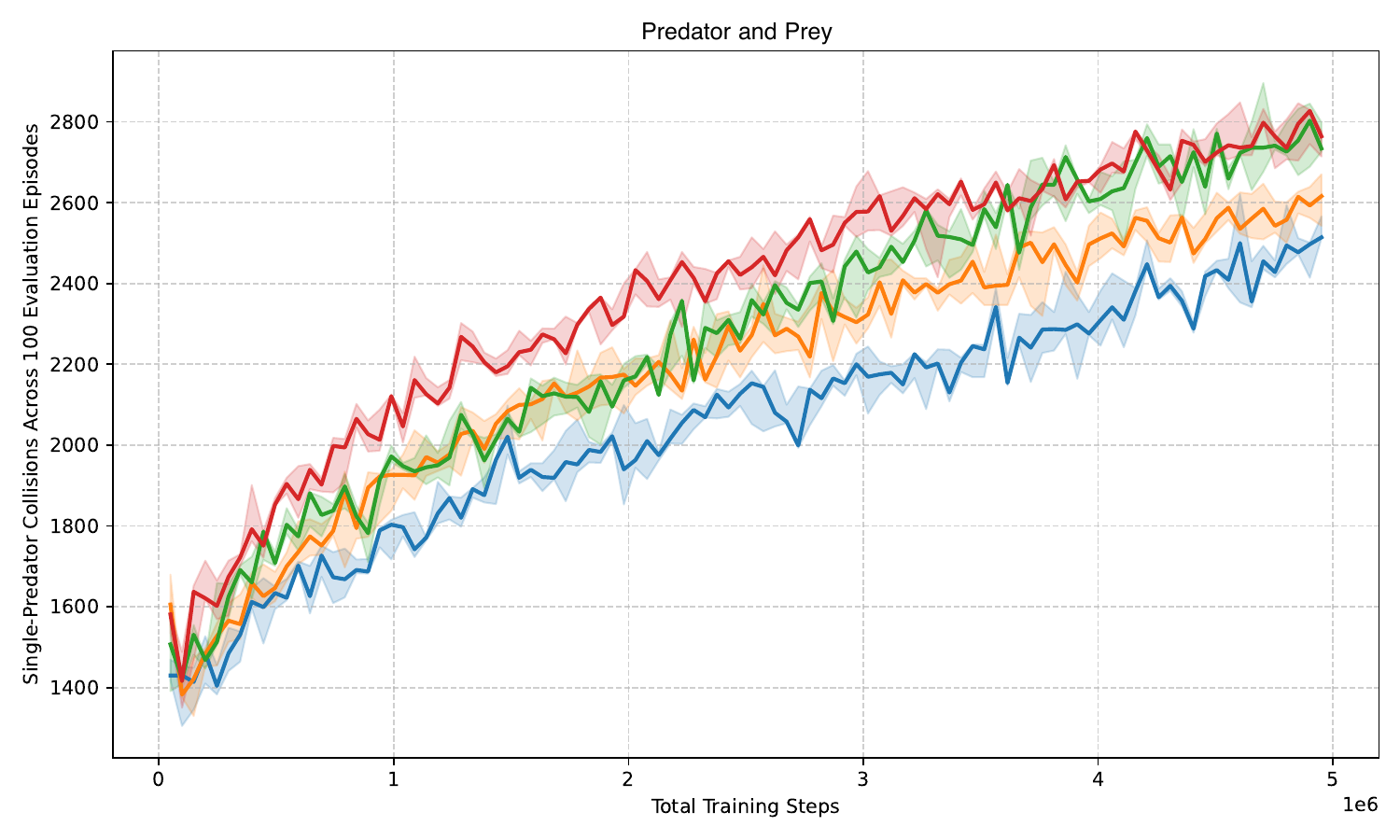}
    \vspace{-8mm}
    \caption{Learning Curve (PP)}
    \vspace{-10mm}
    \label{fig:pp-first}
\end{wrapfigure}
Since \method represents strategy context as a latent embedding, we compare it with LIAM, an opponent-modeling method for partially observable settings.
Both methods use A2C on PP and PPO on GRF and SMAC-HARD.
In PP (see~\cref{fig:pp-first}), \method achieves up to $\gain{400}$ and $\gain{200}$ more single-predator collisions than LIAM over 100 evaluation episodes in the \texttt{Seen} and \texttt{Unseen} settings, respectively.

On the more complex GRF benchmark, \method significantly outperforms LIAM. In R2S (see~\cref{fig:GRF(R2S)}), \method surpasses LIAM after 2M training steps and ultimately scores $\gain{30}$ and $\gain{20}$ more goals over 100 evaluation episodes in the \texttt{Seen} and \texttt{Unseen} settings, respectively.
Similarly, in the R2S-DEF scenario (see~\cref{fig:GRF(R2S-HARD)}), \method outperforms LIAM, scoring up to $\gain{30}$ and $\gain{35}$ more goals in the \texttt{Seen} and \texttt{Unseen} settings, respectively.
\revise{In addition, in the harder R2S-4DEF scenario (see~\cref{fig:GRF(R2S-4def)}), \method outperforms LIAM, scoring up to $\gain{10}$ more goals in the \texttt{Seen} and \texttt{Unseen} settings.}

The superiority of \method extends to the SMAC-HARD as well. On \texttt{3m} (see~\cref{fig:3m}), \method achieves up to $\gain{25\%}$ and $\gain{15\%}$ higher win rates than LIAM in the \texttt{Seen} and \texttt{Unseen} settings, respectively. 
In the \texttt{Seen} setting, \method reaches $\approx$$100\%$ win rate after 400K steps, whereas LIAM does not within the 500K-step training budget.
On \texttt{2s3z} (see~\cref{fig:2s3z}), \method reaches $\approx$$95\%$ win rate after 400K steps, outperforming LIAM by up to $\gain{30\%}$ and $\gain{20\%}$ in the \texttt{Seen} and \texttt{Unseen} settings, respectively.
The performance of \method is particularly substantial on \texttt{3s\_vs\_3z} (see~\cref{fig:3svs3z}). In the \texttt{Seen} setting, \method reaches a $100\%$ win rate after $\approx$700K steps, while LIAM reaches only $\approx$$40\%$ within the 1M-step training, yielding a gain of $\gain{60\%}$. In the \texttt{Unseen} setting, \method achieves $\approx$$90\%$ win rate, whereas LIAM rarely wins, yielding a gain up to $\gain{90\%}$.

\revise{We also investigate decentralized learning of action-level best responses in SMAC-HARD using IPPO, since it is the only benchmark with multiple controlled agents (agents bound to existentially quantified variables). On the \texttt{3m} map (see~\cref{fig:3mippo}), we observe that in the \texttt{Seen} setting, \method converges to a $100\%$ win rate after $\approx$$400$K training steps, while LIAM achieves $\approx$$80\%$ after 400K training steps. On the \texttt{2s3z} map (see~\cref{fig:2s3zippo}), \method performs slightly better than LIAM in both the \texttt{Seen} and \texttt{Unseen} settings. On \texttt{3s\_vs\_3z} (see~\cref{fig:3svs3zippo}), similar to the PPO setting, \method performs substantially better in both the \texttt{Seen} and \texttt{Unseen} settings. In the \texttt{Seen} setting, \method achieves a $100\%$ win rate after $\approx$$700$K steps, while LIAM reaches only $\approx$$60\%$ within the 1M-step training budget, yielding a gain of $\gain{40}\%$ points. In the \texttt{Unseen} setting, \method achieves a win rate of $\approx$$80\%$, whereas LIAM achieves $\approx$$10\%$, yielding a gain of up to $\gain{70}\%$ points.}
Further results are in App.~\ref{app:liam}.

\paragraph{(2) \method $\boldsymbol{\forall^*\exists^*}$ vs. $\boldsymbol{\exists^*\forall^*}$.}
These two \SL fragments induce different learning pipelines. Under $\forall^*\exists^*$, each response function is conditioned on the agent's history and latent embedding (i.e. strategy context) of the preceding universal agents' strategies, whereas under $\exists^*\forall^*$, it uses only the local information.
Consequently, the former is more parameter-intensive, as quantified across benchmarks in~\cref{tab:param}. This is expected because solving the $\forall^*\exists^*$ problem is generally harder; however, it yields adaptive strategies.
However, \method($\forall^*\exists^*$) outperforms \method($\exists^*\forall^*$) by up to $\gain{526}$ and $\gain{380}$ in the \texttt{Seen} and \texttt{Unseen} settings, respectively. This improvement is particularly substantial in the GRF scenarios. In R2S, \method($\forall^*\exists^*$) achieves approximately {\color{mygreenup}$\times 4$} and {\color{mygreenup}$\times 3$} scores in the \texttt{Seen} and \texttt{Unseen} settings, respectively. Similarly, in R2S-DEF, it achieves approximately {\color{mygreenup}$\times1.5$} and {\color{mygreenup}$\times 4$} scores, respectively.
In contrast, in the SMAC-HARD scenarios, ($\forall^*\exists^*$) and ($\exists^*\forall^*$) perform similarly, although ($\forall^*\exists^*$) performs slightly better in 4/6 cases. This suggests that the heavier ($\forall^*\exists^*$) pipeline is not necessary for all cases, and ($\exists^*\forall^*$) can be sufficient. Additional analysis is provided in App.~\ref{app:aevsea}.
\paragraph{(3) \method vs. LIAM Under Irrational Strategies}
\Cref{fig:noise} shows how the shield $\shield$ protects \method when $\strategies^{\text{Train}}$ contains irrational strategies. Across settings containing $20\%$, $30\%$, $40\%$, or $50\%$ irrational strategies, \method~(solid lines) achieves up to $\gain{300}$ more single-predator collisions than LIAM~(corresponding dashed lines). A shield ablation study is provided in App.~\ref{app:shield-ab}.

\paragraph{(4) Inner-\LTL vs. Shaped Reward.}
\Cref{fig:gain} compares inner-\LTL guidance with shaped rewards. On GRF R2S-DEF (left), guidance from $\formularsh$ enables \method to achieve up to $\gain{15}$ and $\gain{20}$ more goals in the \texttt{Seen}~\blueline and \texttt{Unseen}~\orangeline settings, respectively. On \texttt{3s\_vs\_3z} (right), guidance from $\formulasmac$ improves the win rate by up to $\gain{30\%}$ in \texttt{Seen} and $\gain{35\%}$ in \texttt{Unseen}. 
In \texttt{Seen}, both variants eventually approach a $100\%$ win rate, resulting in similar behavior at the end of training, while \texttt{Unseen} gain persists throughout.
Additional ablations are provided in App.~\ref{app:ltl-ab}.

\begin{table}[t!]
\centering
\caption{\small Median performance of \method across \SL fragments using the same inner-\LTL specification and training budget. \texttt{S}, \texttt{U}, and \texttt{WR} denote \texttt{Seen}, \texttt{Unseen}, and win rate, respectively.}
\label{tab:param}
\resizebox{\textwidth}{!}{%
\begin{tabular}{c|ccc|ccc|ccc|ccc|ccc|ccc}
\toprule
\multirow{2}{*}{\method}
& \multicolumn{3}{c|}{\textbf{PP} (\# Single-Collision)} 
& \multicolumn{3}{c|}{\textbf{R2S} (\# Goal)} 
& \multicolumn{3}{c|}{\textbf{R2S-DEF} (\# Goal)} 
& \multicolumn{3}{c|}{\texttt{3m} (WR\%)} 
& \multicolumn{3}{c|}{\texttt{2s3z} (WR\%)}
& \multicolumn{3}{c}{\texttt{3s\_vs\_3z} (WR\%)} \\
\cmidrule(lr){2-4}
\cmidrule(lr){5-7}
\cmidrule(lr){8-10}
\cmidrule(lr){11-13}
\cmidrule(lr){14-16}
\cmidrule(lr){17-19}
& \#\text{Prms} & \texttt{S} & \texttt{U} 
& \#\text{Prms} & \texttt{S} & \texttt{U} 
& \#\text{Prms} & \texttt{S} & \texttt{U} 
& \#\text{Prms} & \texttt{S} & \texttt{U} 
& \#\text{Prms} & \texttt{S} & \texttt{U} 
& \#\text{Prms} & \texttt{S} & \texttt{U} \\
\midrule
($\forall^*\exists^*$) &250K & 2765& 2749& 630K& 47& 37&630K &40 & 44 & 636K&99 & 89& 748K& 98& 98& 645K &100 & 86\\ \midrule
($\exists^*\forall^*$) &111K & 2239& 2369& 334K& 12& 14& 334K& 24& 12 & 445K&96 & 87& 496K& 98& 92& 451K& 98& 87\\ 
\bottomrule
\end{tabular}%
}
\vspace{-2mm}
\end{table}

%% file: 07-conclusion.tex
\vspace{-2mm}
\section{Conclusion}

\method is a specification-guided MARL framework for partially observable adversarial settings.
By employing Strategic Logic, \method explicitly reasons about agents' strategies and their responses to adversaries. 
Our evaluation on Predator-Prey, Google Research Football, and SMAC-HARD demonstrates that \method is effective, scalable, and adaptable across adversarial strategies, improving opponent-modeling baselines by up to ${90\%}$.
\vspace{-2mm}
\paragraph{Limitations.}
\method is limited to \SL fragments with at most one quantifier alternation, which we believe are sufficient to express many adversarial RL problems. Also, its theoretical guarantees rely on information-preserving latent embedding, which may be hard to achieve under partial observability.
\revise{The information-preserving latent embedding condition directly concerns the opponent-strategy information available to \method in practice. The remaining assumptions are not prerequisites for executing \method, whose practical objective is to learn action-level best-response functions that maximize \cref{eq:learning-update}. These assumptions support the theoretical guarantees connecting this optimization objective to the maximization of satisfaction probability in \cref{statement:main} of \cref{sec:problem}.}

%% file: 0X-appendix.tex
\newpage
\appendix

\section{Appendix Outline}
\label{app:outline}
\begin{itemize}
    \item Appendix~\ref{app:proof}: Proofs
    \begin{itemize}
        \item \ref{appendix:proof-lemma1}: Proof of Lemma~\ref{lem:optimality-best-response}
        \item \ref{app:proof_lemma2}: Proof of Lemma~\ref{lem:optimality-robustness}
        \item \ref{app:proof_theory}: Proof of Theorem~\ref{theory:final}
    \end{itemize}

    \item Appendix~\ref{app:fullquant}: Full Quantitative Semantics of Inner-\LTL Formulas

    \item Appendix~\ref{app:exp}: Experimental Setup and Implementation Details
    \begin{itemize}
        \item \ref{app:shield}: Details on the Shield
        \item \ref{app:detencdec}: Encoder, Decoders, and Future-Action Prediction
        \item \ref{app:policy}: Training the Action-Level Response Functions
    \end{itemize}

    \item Appendix~\ref{app:bench}: Benchmark Details
    \begin{itemize}
        \item \ref{app:pp}: Predator-Prey (PP)
        \item \ref{app:grf}: Google Research Football (GRF)
        \item \ref{app:smac-hard}: SMAC-HARD
    \end{itemize}

    \item Appendix~\ref{app:add-exp}: Additional Experimental Results
    \begin{itemize}
        \item \ref{app:liam}: \method vs.\ LIAM
        \item \ref{app:aevsea}: \method $\forall^*\exists^*$ vs.\ $\exists^*\forall^*$
        \item \ref{app:shield-ab}: Shield Ablation Study
        \item \ref{app:ltl-ab}: Inner-\LTL vs.\ Shaped Reward 
    \end{itemize}
    \item Appendix~\ref{app:compute}: Computational Resources
\end{itemize}

\section{Proofs}
\label{app:proof}

\subsection{Proof of Lemma~\ref{lem:optimality-best-response}}\label{appendix:proof-lemma1}
\begin{theorembox}
\textit
\LEMMABS
\end{theorembox}

\begin{proof}
We assume that the convergence objective is order-preserving with respect to satisfaction, meaning that a tuple of response functions with a higher probability of convergence to the maximum robustness value cannot have a lower probability of satisfying the specification.
We now begin the proof by let,
$\mathfrak{B}
=
\tupleof{\brfunc_{x_i}}_{i\in\existsof{}}$
be a tuple of best-response functions that maximizes~\Cref{eq:opt}, and let
$\traceof{\traj_{\assignment}^{\mathfrak{B}}}$ denote a trace sampled under a complete assignment $\assignment$ updated by a tuple $\mathfrak{B}$.

Since $\rbvalue_{\mathit{max}}$ is the greatest positive robustness value, convergence to $\rbvalue_{\mathit{max}}$ implies positive robustness. By the soundness of the quantitative semantics~\cite{li2017reinforcement}, positive robustness implies satisfaction. Therefore,
\[
\rbvalueof{\traceof{\traj_{\assignment}^{\mathfrak{B}}},\psi}
\xrightarrow{*}
\rbvalue_{\mathit{max}}
\quad\Longrightarrow\quad
\rbvalueof{\traceof{\traj_{\assignment}^{\mathfrak{B}}},\psi}>0
\quad\Longrightarrow\quad
\traceof{\traj_{\assignment}^{\mathfrak{B}}}\models\psi.
\]

Thus, convergence to $\rbvalue_{\mathit{max}}$ guarantees satisfaction of $\psi$. By the order-preservation assumption, a tuple with a higher probability of convergence cannot have a lower probability of satisfaction. Since $\mathfrak{B}$ maximizes the probability of convergence in~\Cref{eq:opt}, it also maximizes the probability of satisfying $\psi$.
 Hence, maximizing this satisfaction probability is exactly the optimization problem stated in~\Cref{sec:problem}. Therefore, $\mathfrak{B}$ also solves the original optimization problem.

\end{proof}

\subsection{Proof of Lemma~\ref{lem:optimality-robustness}}
\label{app:proof_lemma2}

\begin{theorembox}
\textit
\LEMMARB
\end{theorembox}

\begin{proof}
Let $\mathcal{B}$ denote the set of well-typed response-function tuples, and let $J_2,J_3:\mathcal{B}\to\mathbb{R}$ denote the objectives in~\Cref{eq:opt} and ~\Cref{eq:learning}, respectively.
By assumption, $J_3$ strictly preserves the ordering induced by $J_2$; that is, for any $\mathbf{B},\mathbf{B}'\in\mathcal{B}$, we have:
\[
    J_2(\mathbf{B}) > J_2(\mathbf{B}')
    \quad\Longrightarrow\quad
    J_3(\mathbf{B}) > J_3(\mathbf{B}').
\]

Let $\mathbf{B}^{*}\in \arg\max\limits_{\mathbf{B}\in\mathcal{B}} J_3(\mathbf{B})$.
Suppose, for contradiction, that $\mathbf{B}^{*}$ does not maximize $J_2$.
Then there exists some $\widehat{\mathbf{B}}\in\mathcal{B}$ such that:
\[
J_2(\widehat{\mathbf{B}})>J_2(\mathbf{B}^{*}).
\]
By strict order preservation, this implies:
\[
    J_3(\widehat{\mathbf{B}})>J_3(\mathbf{B}^{*}),
\]
which contradicts the assumption that $\mathbf{B}^{*}$ maximizes $J_3$.
Therefore,
\[
    \arg\max_{\mathbf{B}\in\mathcal{B}} J_3(\mathbf{B})
    \subseteq
    \arg\max_{\mathbf{B}\in\mathcal{B}} J_2(\mathbf{B});
\]
hence, every maximizer of~\Cref{eq:learning} also maximizes~\Cref{eq:opt}.

\end{proof}

\subsection{Proof of~\cref{theory:final}}
\label{app:proof_theory}

\begin{theorembox}
\textit
\THEOREMFINAL
\end{theorembox}

\begin{proof}
Let us write:
\[
\mathfrak{B}^{a}
=
\langle \mathbf{BR}_{x_i}^{a}\rangle_{i\in\mathbb{Q}^{\exists}}
\]
as a tuple of action-level best-response functions maximizing the
expected cumulative discounted robustness value.
Since $\lambda_{\Xi}$ is information-preserving for every strategy context $\Xi$, the latent representation retains all information required by a well-typed best-response function. 
Hence, for every $\mathbf{BR}_{x_i}^{a}$ in the optimal tuple $\mathfrak{B}^{a}$, there exists an induced best-response function $\mathbf{BR}_{x_i}$ on the original strategy context, such that:
\[
\mathbf{BR}_{x_i}(\Xi)(\history)
=
\mathbf{BR}_{x_i}^{a}(\lambda_{\Xi}, \history),
\]
and the two functions induce the same action choices given the same history $\history$, and therefore the same distribution over trajectories.
Consequently, for every complete assignment $\assignment \in \assignments_c$, the trajectory distribution induced by $\mathfrak{B}^{a}$ in the latent representation coincides with that induced by the corresponding best-response function tuple:
\[
\mathfrak{B}
=
\langle \mathbf{BR}_{x_i} \rangle_{i\in\mathbb{Q}^{\exists}}
\]
in the original representation.
In particular, their expected cumulative discounted robustness values are equal:
\[
J_a(\mathfrak{B}^{a}) = J_3(\mathfrak{B}),
\]
where $J_a$ denotes the action-level objective defined immediately before \cref{theory:final} and $J_3$ denotes
the objective in~\Cref{eq:learning}.

Because the embedding is information-preserving and the action-level response functions can represent every well-typed response function, every well-typed response tuple has an equivalent action-level tuple that induces the same trajectory distribution and objective value.
Therefore, if $\mathfrak{B}^{a}$ maximizes $J_a$, the induced tuple $\mathfrak{B}$ maximizes $J_3$; otherwise, a well-typed tuple with a strictly larger value of $J_3$ would induce a latent-level response tuple with a strictly larger value of $J_a$, contradicting the optimality of $\mathfrak{B}^{a}$.

Thus,
\[
\mathfrak{B}\in\arg\max_{\mathfrak{B} \in \mathcal{B}} J_3.
\]
By Lemma~\ref{lem:optimality-robustness}, every maximizer of~\cref{eq:learning} also maximizes the objective in~\Cref{eq:opt}. 
Hence,
\[
\mathfrak{B}\in\arg\max_{\mathfrak{B} \in \mathcal{B}} J_2.
\]

Finally, by Lemma~\ref{lem:optimality-best-response}, every maximizer of~\Cref{eq:opt} solves the optimization problem of~\Cref{sec:problem} (i.e., it maximizes the probability of satisfying the SL sentence $\psi$ in $\posg$).
Therefore, any tuple of optimal latent-level best-response functions $\mathfrak{B}^{a}$ induces a tuple $\mathfrak{B}$ that solves the original satisfaction-probability optimization problem formalized in~\Cref{statement:main}.

\end{proof}


\section{Full Quantitative Semantics of Inner-\LTL formulas}
\label{app:fullquant}
Below, we present the full quantitative semantics of the inner-\LTL formulas of \SL, adapted from~\citet{li2017reinforcement}.

\input{img/robustnessLTL}

\section{Experimental Setup and Implementation Details}
\label{app:exp}

In this section, we describe the implementation details, architectures, hyperparameters, and loss functions of the models used in our experiments.

\subsection{Details on Shield}
\label{app:shield}

We train a recurrent binary classifier to distinguish rational predator (agent bound to universally quantified variable) compositions from irrational tasks. 
At each time step, the classifier receives the prey's (agent bound to existentially quantified variable) current observation concatenated with its five previous actions.
The previous action vector is zero at the beginning of each interaction sequence. 
The architecture consists of a single-layer LSTM with 128 hidden units, followed by a fully connected layer with 64 units, a ReLU activation, and a scalar output layer. 
The recurrent hidden and cell states are reset independently for each interaction sequence.

Data are collected using a uniform-random prey strategy across $m=20$ training tasks, comprising of ten rational and ten irrational tasks. 
For each task, we collect 1,000 training interaction sequences and 200 independent calibration interaction sequences, each lasting 20 steps.
Ten interaction sequences from the same task form a training bundle. 
Their final-step logits are averaged before applying a sigmoid to obtain the bundle’s classification score. 
The classifier is trained using standard class-weighted binary cross-entropy; the balanced dataset gives equal weights to both classes.
Interaction sequence are randomly regrouped within each task at every epoch, and the resulting bundles are shuffled. 
Training runs for 100 epochs using Adam with a learning rate of 0.001, minibatches of 64 bundles, and gradient-norm clipping at 1.0.



\paragraph{Shield in Deployment.}
During deployment (main training phase), each collected trajectory receives the stored classification of its opponent task. 
Experience classified as random is excluded from the actor, critic, and opponent-model losses, while the remaining losses are averaged over retained experience; if none remains in a batch, we skip the update.

\subsection{Details on Encoder, Decoders and Future-Action Prediction}
\label{app:detencdec}
\paragraph{Encoder ($\enc$).}
\method uses a recurrent encoder to summarize the action-observation history of each agent bound to an existentially quantified variable into a latent context representation. 
At each time step, the encoder receives the agent's current local observation concatenated with its previous one-hot-encoded action.
The encoder consists of a single-layer LSTM with 128 hidden units, followed by a fully connected layer with 128 units and a ReLU activation. A final linear layer produces a 32-dimensional ($d =32$) embedding without an output activation.

Training uses five-step sequences with truncated backpropagation through time, storing the exact recurrent states at each sequence’s start for replay. 
The encoder is optimized using Adam with a fixed learning rate of 0.0007 and gradient-norm clipping at 0.5.

\paragraph{Action-Reconstruction Decoder ($\deca$).}
The action-reconstruction decoder predicts the current action of each agent bound to a preceding universally quantified variable from the encoder's latent context embedding ($\latent \in \stratspace \subseteq \mathbb{R}^{32}$).
It uses a shared feed-forward network with two fully connected hidden layers of 128 units, each followed by a ReLU activation. 
Separate linear outputs produce action logits for each agent bound to a preceding universally quantified variable, and a softmax converts these logits into categorical action probabilities. 
This decoder is trained jointly with the encoder using Adam with a learning rate of 0.0007 and gradient-norm clipping at 0.5.

\paragraph{Observation-Reconstruction Decoder ($\deco$).}
The observation-reconstruction decoder predicts the agents bound to a preceding universally quantified variable observations from the encoder's latent context embedding ($\latent \in \stratspace \subseteq \mathbb{R}^{32}$). 
It consists of two fully connected hidden layers with 128 units each, followed by ReLU activations, and a final linear layer that outputs the reconstructed observation vector without an output activation. 
This branch has separate parameters from the action-reconstruction branch. 
It is trained jointly with the encoder using Adam with a learning rate of 0.0007 and gradient-norm clipping at 0.5.

 \paragraph{Future-Action Predictor ($\future$).}

 The future-action predictor estimates each agent bound to a preceding universally quantified variable action distribution over the next 5 steps from the encoder’s  latent embedding. 
 It consists of two fully connected hidden layers with 128 units each and ReLU activations, followed by a linear output layer. 
 The output is organized by prediction horizon and an agent bound to a preceding universally quantified variable, with a softmax producing a categorical action for each combination. 
The predictor is trained jointly with the encoder using Adam with a learning rate of 0.0007 and gradient-norm clipping at 0.5.

\paragraph{Training.}
We jointly train the encoder ($\enc$), observation-reconstruction decoder ($\deco$), action-reconstruction decoder ($\deca$), and future-action predictor ($\future$) using samples collected. 
Samples are divided into five-step sequences, and the encoder is replayed from its stored hidden and cell states. 
Its latent embedding at each step is shared by all three prediction branches.
Observation reconstruction ($\deco_{wo}$) uses a standard squared-error loss. 
We sum the squared differences between predicted and recorded observation features, multiply by 0.5, and average over retained steps and environments.
Action reconstruction ($\deca_{wa}$) uses standard categorical cross-entropy against the opponents’ recorded actions. The losses are summed across opponents and averaged over retained steps and environments.
Future-action predictor ($\future_{wf}$) uses standard categorical cross-entropy against recorded future actions. The loss is averaged over steps, environments, agents bound to a preceding universally quantified variables , and the five prediction horizons. Targets beyond the available rollout or across episode boundaries are excluded.

The combined loss for learning the latent context embedding is calculated as follows:
\[
\mathcal{L}_{\latent}
=
\mathcal{L}_{\deco}
+
\mathcal{L}_{\deca}
+
0.1\,\mathcal{L}_{\future}.
\]

Gradients from all three terms update the shared encoder, while each prediction branch is updated through its own term. We use a shared Adam optimizer with a learning rate of 0.0007, clipping gradient norms to 0.5 separately for the encoder, decoders, and future-action predictor. 
Backpropagation is truncated at five-step boundaries, while recurrent states carry information across sequences.

\paragraph{Information-preserving latent context embedding} Let $w^*$ denote the encoder parameters obtained by minimizing the combined loss $\mathcal{L}_{\latent}$. The resulting embedding is $\widehat{\latent}=\enc_{w^\optimal}(h_{\agent^i})$. By jointly reconstructing current actions and observations and predicting future actions, the encoder is encouraged to preserve the information needed to approximate $\latent$.
However, because the encoder has access only to the agent's local action-observation history, different strategy contexts may be observationally indistinguishable.
Therefore, the learned embedding may only approximate the information-preserving latent context embedding.

\subsection{Details on Training the Action-level Response Functions}
\label{app:policy}

\paragraph{Predator-Prey.} \method uses A2C algorithm with a single-layer LSTM containing 128 hidden units. 
The LSTM processes the current observation and previous one-hot action. 
Its output is concatenated with the current observation and corrected 32-dimensional latent embedding, producing a 174-dimensional input to the actor and critic. 
Each network contains two independent fully connected hidden layers with 128 units and ReLU activations. The actor produces a categorical distribution over five actions, while the critic produces a scalar value estimate.

\paragraph{GRF.} \method uses a recurrent PPO algorithm with a 128-unit LSTM. 
The current observation, 32-dimensional latent embedding, and history output form a 275-dimensional input to the actor and critic. 
Each network contains three independent fully connected hidden layers with 256 units and ReLU activations. The actor produces a categorical distribution over 19 actions, and the critic produces a scalar value estimate. 
PPO performs five optimization epochs per rollout update, using a clipping threshold of 0.2 for both the probability ratio and value prediction.

\paragraph{SMAC-HARD.}
\revise{\method uses a 128-unit policy LSTM and separate actor and critic
networks, each with three 256-unit ReLU hidden layers.
Each controlled unit receives its own local observation, previous
action, and locally computed latent embedding.
Under PPO, response-function parameters are shared across controlled
units. Under IPPO, each unit has independent policy LSTM, actor,
critic, and future-action-adapter parameters, with its own Adam
optimizer; the auxiliary encoder and prediction heads remain shared.
In both cases, recurrent hidden and cell states are maintained
separately for each unit.
Observation and action dimensions depend on the map. Unavailable
actions are masked before constructing the categorical action
distribution.
Both PPO and IPPO use five optimization epochs and clipping thresholds
of 0.2 for the probability ratio and value updates.}

\paragraph{Training.} 
\revise{Across all three benchmarks, the actor and critic share a policy LSTM,
which remains separate from $\enc$.
Training uses five-step recurrent sequences for A2C and PPO, and
20-step sequences for IPPO. Hidden and cell states are carried across
sequences and reset at episode boundaries.
The response function's learning rate remains constant for the first
30\% of training and then decreases linearly to one-tenth of its
initial value.
For both PPO and IPPO, behavior log probabilities, behavior value
estimates, and advantages are fixed before optimization begins;
recurrent chunks are shuffled between epochs while timestep order
within each chunk is preserved.
For IPPO, advantages are normalized separately for each controlled
unit, and each unit's policy parameters are updated using an independent
Adam optimizer with gradient-norm clipping applied separately per unit.
Details on the hyperparameters for training the response functions
appear in~\cref{tab:hyperparameters}.}

\begin{table}[t]
\centering
\caption{Hyperparameters of the response function of \method.
Learning rates refer to the actor--critic optimizer.}
\label{tab:hyperparameters}
\begin{tabular}{lcccc}
\hline
Hyperparameter & PP & GRF & SMAC-HARD & SMAC-HARD (IPPO) \\
\hline
Algorithm & A2C & PPO & PPO & IPPO \\
LSTM layers & 1 & 1 & 1 & 1 \\
LSTM hidden units & 128 & 128 & 128 & 128 \\
Hidden layers per actor/critic & 2 & 3 & 3 & 3 \\
Units per hidden layer & 128 & 256 & 256 & 256 \\
Hidden activation & ReLU & ReLU & ReLU & ReLU \\
Opponent embedding dimension & 32 & 32 & 32 & 32 \\
Optimizer & Adam & Adam & Adam & Adam \\
Initial learning rate
    & $3\times10^{-4}$ & $3\times10^{-4}$
    & $7\times10^{-4}$ & $3\times10^{-4}$ \\
Final learning rate
    & $3\times10^{-5}$ & $3\times10^{-5}$
    & $7\times10^{-5}$ & $3\times10^{-5}$ \\
Learning-rate decay start & 30\% & 30\% & 30\% & 30\% \\
Learning-rate decay schedule & Linear & Linear & Linear & Linear \\
Maximum gradient norm & 0.5 & 0.5 & 10.0 & 10.0 \\
Recurrent sequence length & 5 & 5 & 5 & 20 \\
Discount factor $\gamma$ & 0.99 & 0.99 & 0.99 & 0.99 \\
GAE parameter $\lambda$ & 0.95 & 0.95 & 0.95 & 0.95 \\
Entropy coefficient & 0.01 & 0.01 & 0.01 & 0.01 \\
PPO optimization epochs & -- & 5 & 5 & 5 \\
PPO probability-ratio clipping & -- & 0.2 & 0.2 & 0.2 \\
PPO value clipping & -- & 0.2 & 0.2 & 0.2 \\
\hline
\end{tabular}
\end{table}

\section{Benchmark Details}
\label{app:bench}

\subsection{Predator and Prey}
\label{app:pp}

\paragraph{Environment.}
We used \citep{NEURIPS2021_a03caec5} implementation of predator and prey. The predator–prey environment is a two-dimensional arena containing one prey, three predators, and two stationary obstacles. All agents choose among five discrete actions: $\setof{Stay,left,right,Up,Down}$. Episodes terminate after 50 steps.
The prey has partial observation. It observes its own position and velocity, together with the relative positions of obstacles and predators within visible range.

\paragraph{\SL Formula.}
We used the reward used in \citep{NEURIPS2021_a03caec5} for the guiding the LIAM for the baseline. For \method we use $\formulapp$ as the \SL formula in our experiments:
\begin{align*}
\formulapp
\definedas {}&
\forall x_1\,\forall x_2\,\forall x_3\,\exists x_4\,
(\text{pred\_1},x_1)
(\text{pred\_2},x_2)
(\text{pred\_3},x_3)
(\text{prey},x_4).\\
&\G\F \textsf{Exact\_One}(\texttt{pos\_pred\_1}, \texttt{pos\_pred\_2},\texttt{pos\_pred\_3},  \texttt{pos\_prey})
\land \\
&\neg \textsf{Two\&More\_Eq}(\texttt{pos\_pred\_1}, \texttt{pos\_pred\_2},\texttt{pos\_pred\_3},  \texttt{pos\_prey}) \land \\
&\neg \textsf{Out\_of\_Boundary}(\texttt{pos\_prey})
\end{align*}
Although $\G\F$ conventionally expresses a Büchi recurrence condition over infinite traces, under our finite-prefix evaluation it checks whether $\textsf{Exact\_One}$ holds at the final step of each prefix. 
Because we evaluate the robustness of every prefix throughout an episode, this condition rewards repeated occurrences of contact with exactly one predator. 
\paragraph{Learning set of Strategies.}
We have two sets of strategies. 
The first set adopted from~\citep{NEURIPS2021_a03caec5} contains four heuristic strategies: pursuing the prey, following a randomly selected predator, approaching the nearest other agent, and moving toward the largest absolute relative-coordinate displacement among other agents. 
It also includes three pretrained IA2C \citep{papoudakis2021benchmarking} strategies, three pretrained MADDPG~\citep{NIPS2017_68a97503}, and observation-independent strategies (irrational strategies) that sample uniformly from the five available actions. 
Opponent tasks are formed by assigning a strategy to each of the three predator slots (Simulating the job of complete assignment $\assignment \in \assignments_c$).

Second, we construct a learned set of strategies with four training styles: direct, intercept, flank, and spread. Direct strategies use the environment reward without additional shaping. 
Intercept strategies are encouraged to approach the prey’s projected position using its current velocity, flank strategies are encouraged to approach from different directions, and spread strategies are encouraged to maintain greater separation between predators. 
These behaviors are encouraged through reward shaping with a coefficient of 0.05 rather than prescribed action rules. 
The set of strategies comprises IPPO training strategies with direct, intercept, and flank styles, plus held-out strategies covering new instances, the unseen spread style, and MAPPO~\citep{NEURIPS2022_9c1535a0} opponents. Check more details on the opponent strategy sets in \cref{tab:predator_prey_strategy_sets}.

\begin{table}[h!]
\centering
\caption{Training and evaluation strategy sets in Predator--Prey.
Each team contains three predators.
D: direct pursuit; I: interception; F: flanking; S: spreading.
Tactic counts denote distinct individual strategies.}
\label{tab:predator_prey_strategy_sets}
\small
\begin{tabular}{c|p{0.29\textwidth}|p{0.29\textwidth}}
\hline
Setting & \texttt{Seen} & \texttt{Unseen} \\
\hline
Novel Strategy
& $\strategies^{\text{Train}}$:Sampled from 30 IPPO
12 D + 9 I + 9 F
$\strategies^{\text{Test}}$: 30 fixed teams from IPPO
&
$\strategies^{\text{Train}}$:Sampled from 30 IPPO
12 D + 9 I + 9 F
$\strategies^{\text{Test}}$: 21 IPPO + 12 MAPPO  \\
\hline
Inspired by LIAM
& $\strategies^{\text{Train}}$: 4 heuristic + 3 IA2C + 3 MADDPG 
$\strategies^{\text{Test}}$:4 heuristic + 3 IA2C + 3 MADDPG 
&$\strategies^{\text{Train}}$: 4 heuristic + 3 IA2C + 3 MADDPG 
$\strategies^{\text{Test}}$: New Composition from the Training set \\
\hline
\end{tabular}
\end{table}

\subsection{Google Research Football (GRF)}
\label{app:grf}
\paragraph{Environment.}
We use two scenarios in Google Research Football \citep{DBLP:conf/aaai/KurachRSZBERVMB20}: Academy Run to Score (R2S) with Keeper and a modified version of this scenario (R2S-DEF) \revise{and (R2S-4DEF)}. 
In both, the agent controls an attacker who starts near midfield with the ball and attempts to score against five defenders and a goalkeeper. 
In (R2S), all five defenders initially approach from behind the attacker. 
In (R2S-DEF), two defenders start between the attacker and the goal, while the remaining three approach from behind. 
\revise{and In (R2S-4DEF), four defenders start between the attacker and the goal, while the one remaining approach from behind.}
We cap episodes at 200 steps in both scenarios, with earlier termination following a goal, the ball leaving play, or a change in possession.

Both scenarios use 19 discrete actions, including movement, passing, shooting, sprinting, dribbling, and tackling. 
Observations use the 115-dimensional embedding, containing player positions and movement directions, ball information, possession, the active player, and the game mode. 
In the partially observed setting, opponent positions and movement directions are available within a Euclidean distance of 0.6 in normalized coordinates; otherwise, these values are replaced with -5.

\paragraph{\SL Formula.}
We use the default reward used in \citep{DBLP:conf/aaai/KurachRSZBERVMB20} for the guiding the LIAM for the baseline. For \method we use $\formulars$  and $\formularsh$ \revise{(for R2S-DEF and R2S-4DEF)} as the \SL formula in our experiments:

\begin{align*}
\formulars
\definedas {}&
\forall x_1 \ldots\forall x_6\,\exists x_7\,
(\text{def\_1},x_1)
(\text{def\_2},x_2)
\ldots
(\text{def\_5},x_5)
(\text{keeper},x_6)
(\text{atk},x_7).\\
&\big( \textsf{Reach\_check\_1}(\texttt{pos\_atk}) ~ \then_W ~\textsf{Reach\_check\_2}(\texttt{pos\_atk}) \then_W \ldots  \then_W \\ & \textsf{Reach\_check\_10}(\texttt{pos\_atk})  \then_W   ~~\textsf{Shot\_on\_Target}(\texttt{atk})~~ \then_W~~ \textsf{Score}(\texttt{atk}) \big) \\  &\land \G \neg \textsf{Terminate}(\texttt{atk}) \land \G (\bigwedge_{i=1}^5 \neg \textsf{Score}(\texttt{def\_i}))
\end{align*}

\begin{align*}
\formularsh
& \definedas {}
\forall x_1 \ldots\forall x_6\,\exists x_7\,
(\text{def\_1},x_1)
(\text{def\_2},x_2)
\ldots
(\text{def\_5},x_5)
(\text{keeper},x_6)
(\text{atk},x_7).\\
&\big( \textsf{Reach\_check\_1}(\texttt{pos\_atk}) ~ \then_W ~\textsf{Reach\_check\_2}(\texttt{pos\_atk}) \then_W \ldots  \then_W \\ & \textsf{Reach\_check\_10}(\texttt{pos\_atk})  \then_W   ~~\textsf{Shot\_on\_Target}(\texttt{atk})~~ \then_W~~ \textsf{Score}(\texttt{atk}) \big) \\  &\land \G (\bigwedge_{i=1}^5 \neg \textsf{Ball\_Pos}(\texttt{def\_i}))  \land \G (\bigwedge_{i=1}^5 \neg \textsf{Score}(\texttt{def\_i})) 
\end{align*}

\paragraph{Learning set of Strategies.}
We construct separate strategy sets for the R2S, R2S-DEF, R2S-4DEF GRF scenarios by pretraining six-player teams against the built-in football AI. Each team contains five outfield players and a goalkeeper, with separate strategies for each player. Every strategy uses a feed-forward network with two 64-unit ReLU hidden layers, mapping a 115-dimensional observation to a categorical distribution over 19 actions. Teams train for one million environment steps with independent random seeds. IPPO uses a separate critic for each player, whereas MAPPO~\citep{NEURIPS2022_9c1535a0} uses a shared centralized critic during training.

All six players receive the same terminal reward: +1 for gaining possession, -1 for conceding a goal, and 0 for other episode endings. We add a small behavioral reward at each step, weighted by 0.001, to encourage different defensive styles. Low-block rewards favor positions near the penalty area and designated defensive locations. Medium-block rewards favor a defensive band farther from goal and discourage retreating into the penalty area. Pressing rewards encourage the three closest outfield players to approach the ball. Zonal rewards encourage players to follow assigned defensive regions that shift with the ball, together with a smaller incentive to apply pressure.
For each scenario, the strategy set consists of IPPO teams exhibiting low-block, medium-block, and pressing tactics. Independently trained MAPPO teams~\citep{NEURIPS2022_9c1535a0}, exhibiting low-block, medium-block, pressing, and zonal tactics, are reserved for evaluation.
The exact opponent-strategy settings used in our experiments are shown in~\cref{tab:grf_strategy_sets}.

\begin{table}[h!]
\centering
\caption{Strategy sets in GRF. LB: low block; MB: medium block;
PR: pressing; ZN: zonal. $\mathrm{AI}_{d}$ denotes a built-in AI
task at difficulty $d$. Seen evaluation uses the training teams.
Repeated AI tasks at the same difficulty use separate random streams;
difficulty 0.7 is familiar in unseen R2S-4DEF evaluation.}
\label{tab:grf_strategy_sets}
\begin{tabular}{c|p{0.29\textwidth}|p{0.29\textwidth}}
\hline
Maps & \texttt{Seen} & \texttt{Unseen} \\
\hline
\texttt{R2S}
& $\strategies^{\text{Train}}$:
4 LB + 3 MB + 3 PR (IPPO).
\par
$\strategies^{\text{Test}}$:
Same as training.
& $\strategies^{\text{Train}}$:
4 LB + 3 MB + 3 PR (IPPO).
\par
$\strategies^{\text{Test}}$:
3 LB + 2 MB + 2 PR + 3 ZN (MAPPO). \\
\hline
\texttt{R2S-DEF}
& $\strategies^{\text{Train}}$:
4 LB + 3 MB + 3 PR (IPPO).
\par
$\strategies^{\text{Test}}$:
Same as training.
& $\strategies^{\text{Train}}$:
4 LB + 3 MB + 3 PR (IPPO).
\par
$\strategies^{\text{Test}}$:
3 LB + 2 MB + 2 PR + 3 ZN (MAPPO). \\
\hline

\color{red}
\texttt{R2S-4DEF}
& \color{red} $\strategies^{\text{Train}}$:
2 LB + 1 MB + 2 PR (IPPO),
2 AI-0.6 + 3 AI-0.7.
\par
$\strategies^{\text{Test}}$:
Same as training.
& \color{red} $\strategies^{\text{Train}}$:
2 LB + 1 MB + 2 PR (IPPO),
2 AI-0.6 + 3 AI-0.7.
\par
$\strategies^{\text{Test}}$:
2 LB + 1 MB + 1 PR + 1 ZN (MAPPO),
4 AI-0.8 + 1 AI-0.10. \\
\hline
\end{tabular}
\end{table}

\subsection{SMAC-HARD}
\label{app:smac-hard}

\paragraph{Environment.}
We use SMAC-HARD \citep{deng2024smachardenablingmixedopponent}, an extension of the StarCraft Multi-Agent Challenge (SMAC) \citep{samvelyan19smac}. 
Whereas standard SMAC uses the built-in StarCraft II AI for the opposing team, SMAC-HARD provides multiplayer maps and an interface for externally controlled enemy strategies. 
It also introduces extensions for fog of war, unit abilities, terrain, and visibility. 
These features are inherited from SMAC-HARD. 
We extend SMAC-HARD by integrating PyMARL to train enemy-team strategies and deploy their frozen checkpoints as opponent populations.
Our implementation supports learned, scripted, and built-in-AI opponents, with explicit task IDs defining controlled training and seen/unseen evaluation splits.
We expose opponent action labels and physical features for auxiliary supervision while preserving the allied agents' original observations.
We synchronize all participating clients so that each environment transition advances exactly eight frames.
Each episode starts with fresh, living teams, and simulation step and reset failures raise errors rather than being recorded as battle losses.
Games are restarted between episodes when needed to avoid the simulation time limit, while preserving training progress.
Evaluation preserves the training random-number states and performs no parameter updates.
In our experiments, we use the \texttt{3s\_vs\_3z} map, where the first team contains three Stalkers opposing three Zealots, the \texttt{3m} map, where each team contains three Marines, and the \texttt{2s3z}, where each team contains two Stalkers and three Zealots.
Agents receive local SMAC observations and select available movement, stop, and targeted attack actions.
The first-team agents are bound to existentially quantified variables, while the second-team agents are bound to universally quantified variables.

\paragraph{\SL Formula.}
We use the shaped reward introduced in \citep{deng2024smachardenablingmixedopponent,samvelyan19smac} for the guiding the LIAM for the baseline. For \method we use $\formulasmac$ as the \SL formula in our experiments:

\begin{align*}
    \formulasmac \definedas& \forall x_1 \ldots x_l \exists y_{1}\ldots y_m (\text{ene\_1},x_1) \ldots (\text{ene\_l},x_l) (\text{ally\_1},y_{1}) \ldots (\text{ally\_m},y_{m}) \\ & ~\bigg[ \textsf{Diff}\bigg(\textsf{Diff}\Big(\textsf{Sum}(\texttt{Health\_ene\_1}, \ldots, \texttt{Health\_ene\_l}), \\ & \textsf{Sum}(\texttt{Health\_prev\_ene\_1}, \ldots, \texttt{Health\_prev\_ene\_l}) \Big), \\ &\textsf{Diff}\Big(\textsf{Sum}(\texttt{health\_ally\_1}, \ldots, \texttt{health\_ally\_m}), \\ &\textsf{Sum}(\texttt{health\_prev\_ally\_1}, \ldots, \texttt{health\_prev\_ally\_m})\Big) \bigg] ~~ \U_W 
   \\& \textsf{Eliminate}(\texttt{ene\_1}, \ldots, \texttt{ene\_l}) ~\U_W ~ \textsf{Victory}(\texttt{ally\_1}, \ldots, \texttt{ally\_m})
\end{align*}

\paragraph{Learning set of Strategies.}
We extend SMAC-HARD with an explicit opponent-task interface. Each task selects a complete enemy team at the beginning of an episode, and that team remains active throughout the battle. The interface supports scripted strategies, pretrained PyMARL~\citep{samvelyan19smac} teams, and native StarCraft II AI at different difficulty levels (i.e. 8, 9, 10). We additionally integrate PyMARL with the enemy side, enabling teams trained using VDN, QMIX, and QTRAN to be deployed as fixed opponents. The exact opponent-strategy settings used in our experiments are shown in~\cref{tab:smac-setting}.
\begin{table}[h!]
\centering
\caption{Strategy set setting in SMAC-HARD.}
\label{tab:smac-setting}
\begin{tabular}{c|p{0.29\textwidth}|p{0.29\textwidth}}
\hline
Maps & \texttt{Seen} & \texttt{Unseen} \\
\hline
\texttt{3m}
& $\strategies^{\text{Train}}$: $2\times$ VDN + $3\times$ QMIX, $\strategies^{\text{Test}}$: $2\times$ VDN + $3\times$ QMIX
& $\strategies^{\text{Train}}$: $2\times$ VDN + $3\times$ QMIX, $\strategies^{\text{Test}}$:$5\times$ QTRAN \\ \hline
\texttt{2s3z}
& $\strategies^{\text{Train}}$: $2\times$ VDN + $3\times$ AI-8 $\strategies^{\text{Test}}$: $2\times$ VDN + $3\times$ AI-8
& $\strategies^{\text{Train}}$: $2\times$ VDN + $3\times$ AI-8 $\strategies^{\text{Test}}$: $2\times$ QMIX + $3\times$ AI-10 \\ \hline
\texttt{3s\_vs\_3z}
& $\strategies^{\text{Train}}$: $3\times$ VDN + $2\times$ QMIX $\strategies^{\text{Test}}$: $3\times$ VDN + $2\times$ QMIX
& $\strategies^{\text{Train}}$: $3\times$ VDN + $2\times$ QMIX $\strategies^{\text{Test}}$:$5\times$ QTRAN \\
\hline
\end{tabular}
\end{table}

\section{Additional Experimental Results}
\label{app:add-exp}

\begin{figure}
    \centering
    \includegraphics[width=0.7\linewidth]{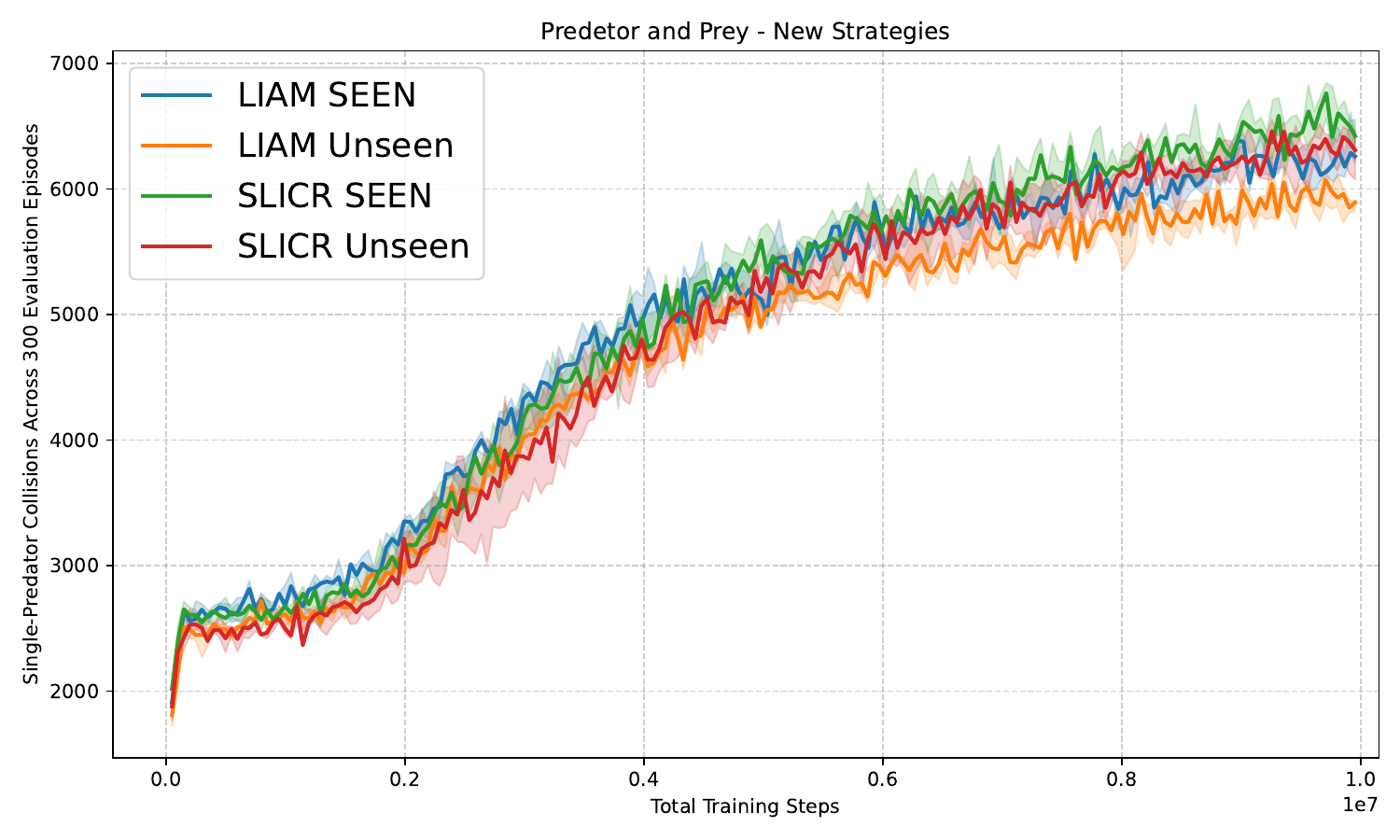}
    \captionof{figure}{Learning curves for \method and LIAM on PP with updated population set of strategies for opponents, showing the median number of single-collision of prey and predator and the 25th–75th percentiles.}
    \label{fig:pp-new}
\end{figure}

\subsection{\method vs. LIAM}
\label{app:liam}
\Cref{fig:pp-new} compares \method and LIAM on PP using a new population of opponent strategies constructed with IPPO and MAPPO. In the \texttt{Seen} setting, \method~\greenline performs similarly to LIAM~\blueline during most of the training process. However, during the final training steps, \method achieves up to $\gain{200}$ more single predator-prey collisions. In the \texttt{Unseen} setting, the advantage is more substantial, with \method~\redline achieving up to $\gain{500}$ more single predator-prey collisions than LIAM~\orangeline during the final training steps.

\begin{figure*}[t]
    \centering

\vspace{-1mm}
    
    \begin{subfigure}[t]{\textwidth}
        \centering
        \includegraphics[width=0.99\linewidth]{img/liam-slicr.pdf}
    \end{subfigure}

    \begin{subfigure}[t]{0.32\textwidth}
        \centering
        \includegraphics[width=\linewidth]{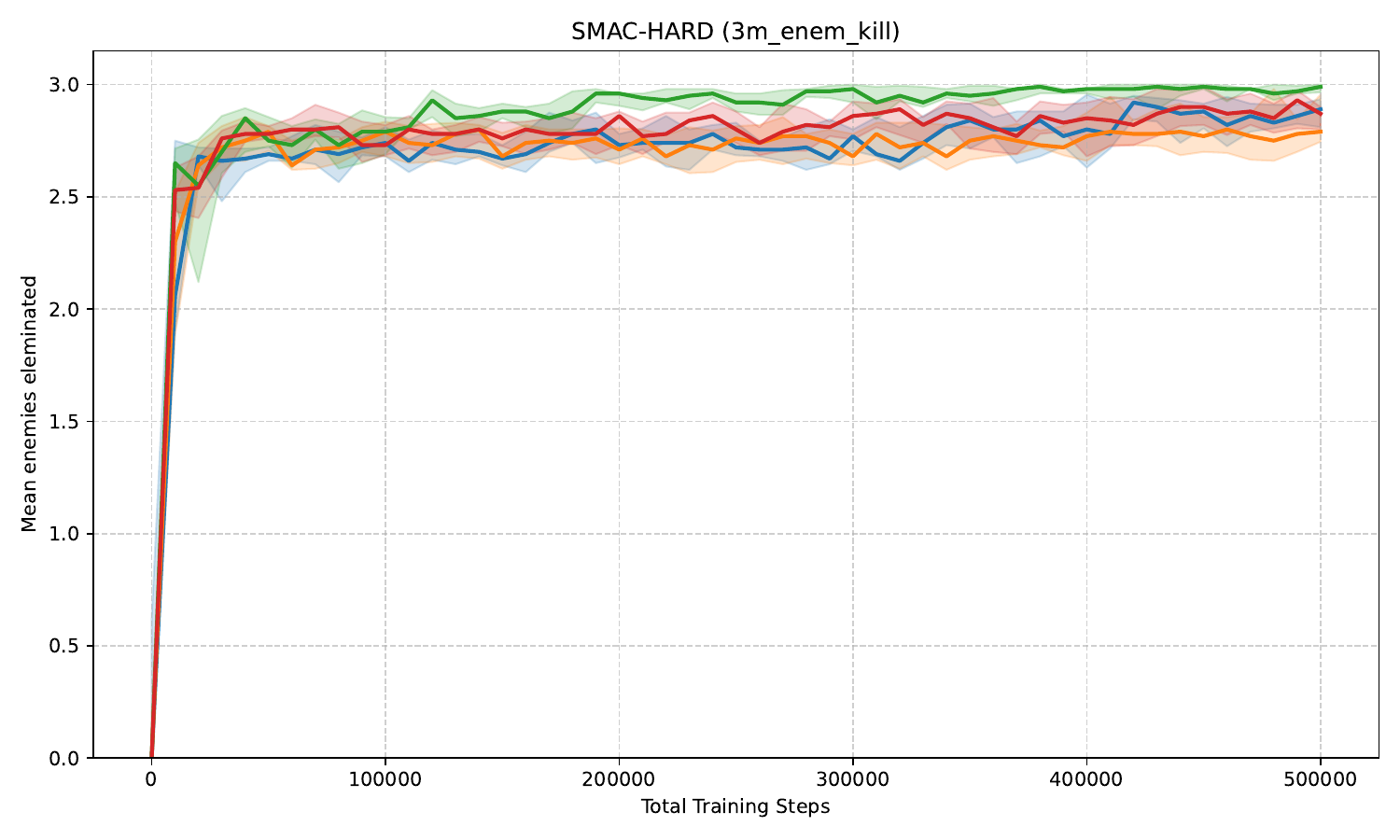}
        \caption{SMAC-HARD (3m)}
        \label{fig:3m-enem}
    \end{subfigure}
    \hfill
    \begin{subfigure}[t]{0.32\textwidth}
        \centering
        \includegraphics[width=\linewidth]{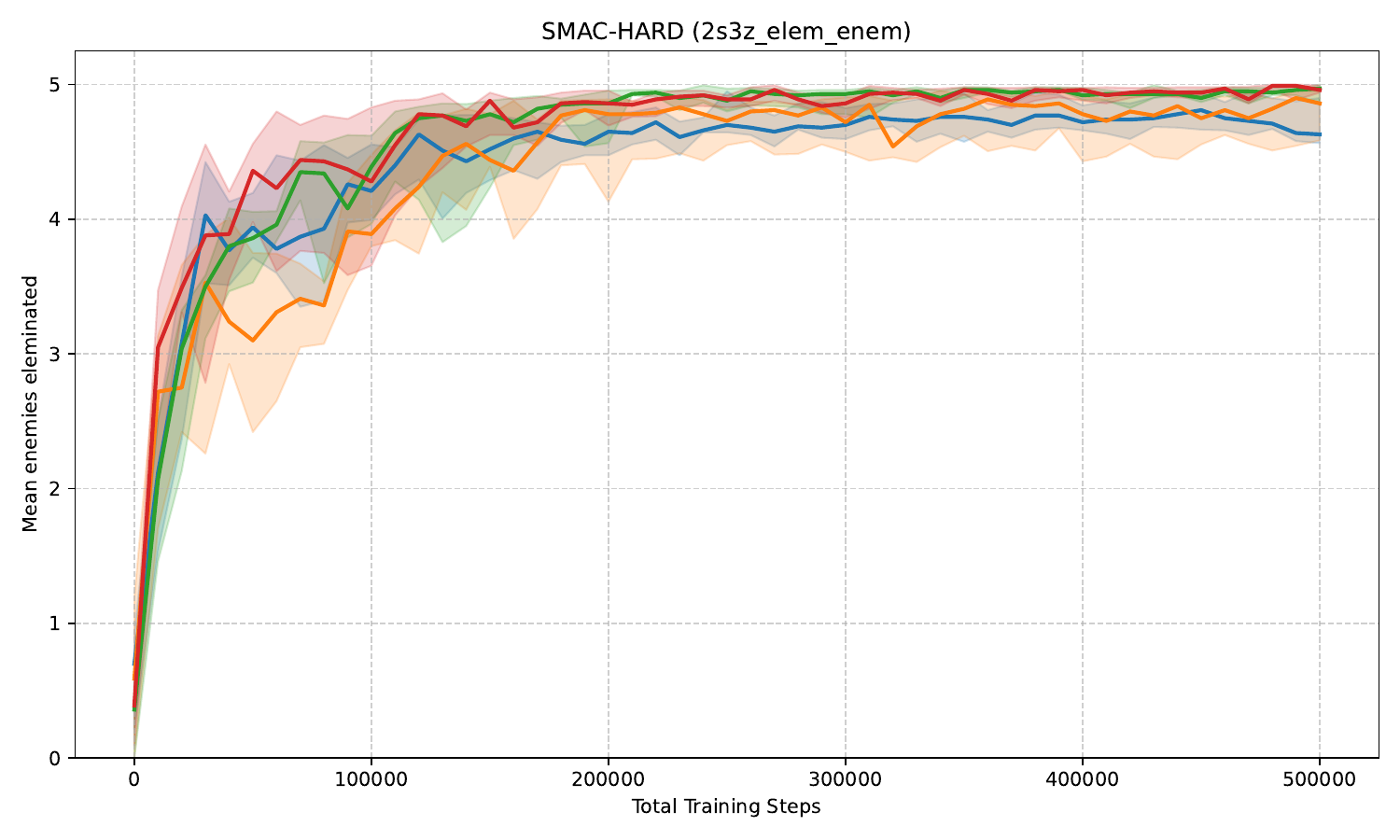}
        \caption{SMAC-HARD (2s3z)}
        \label{fig:2s3z-enem}
    \end{subfigure}
    \hfill
    \begin{subfigure}[t]{0.32\textwidth}
        \centering
        \includegraphics[width=\linewidth]{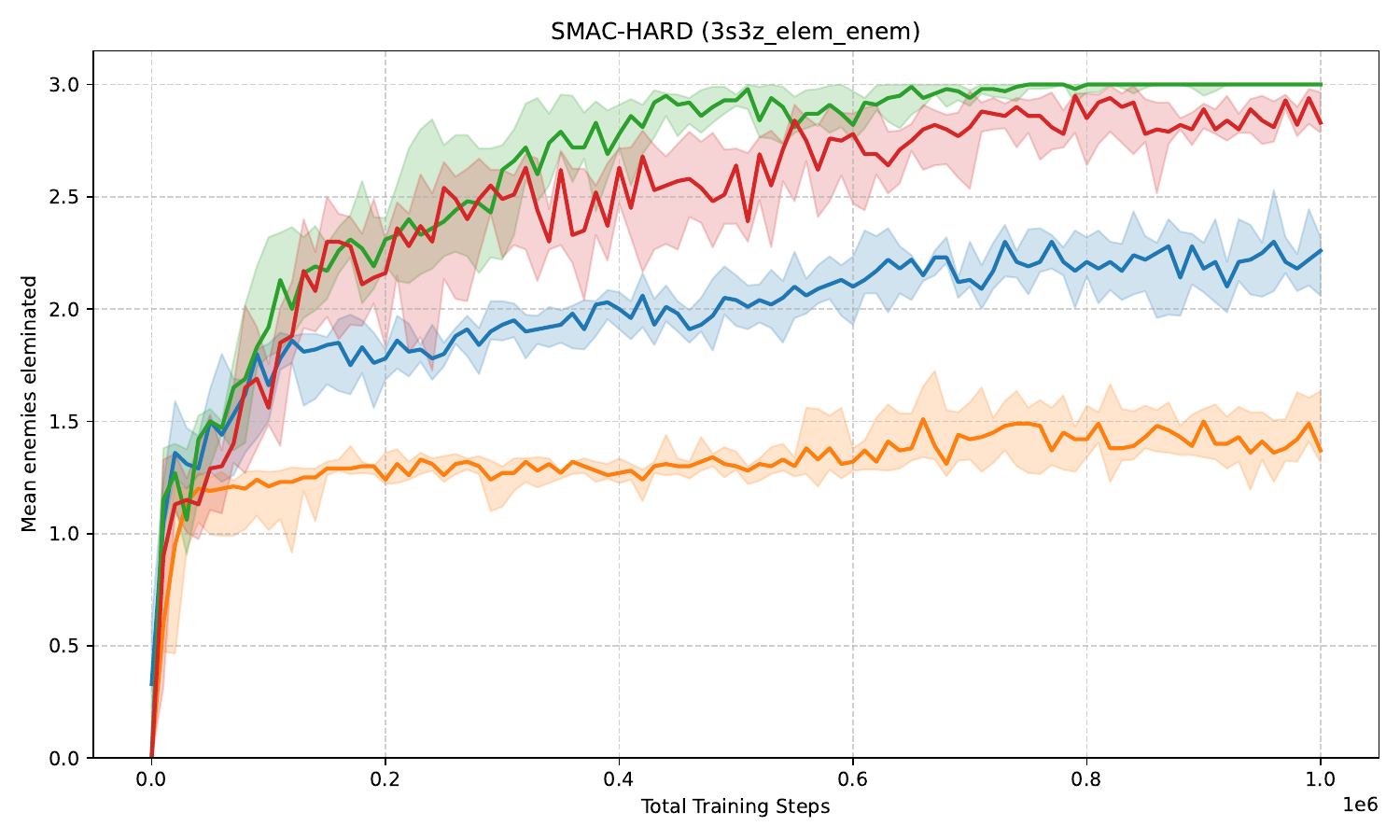}
        \caption{SMAC-HARD (3s\_vs\_3z)}
        \label{fig:3svs3z-enem}
    \end{subfigure}

    \begin{subfigure}[t]{0.32\textwidth}
        \centering
        \includegraphics[width=\linewidth]{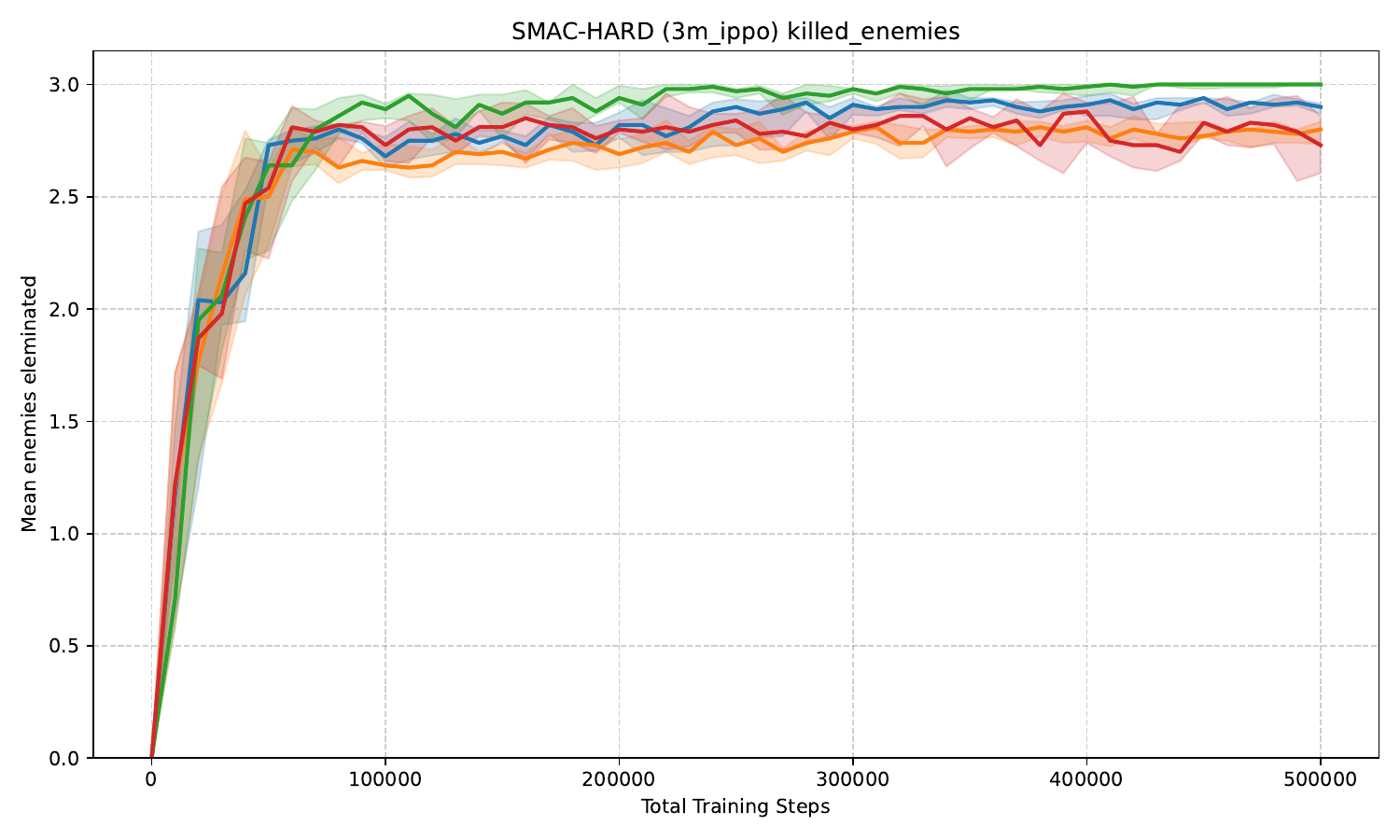}
        \caption{SMAC-HARD (3m--IPPO)}
        \label{fig:3m-enem-ippo}
    \end{subfigure}
    \hfill
    \begin{subfigure}[t]{0.32\textwidth}
        \centering
        \includegraphics[width=\linewidth]{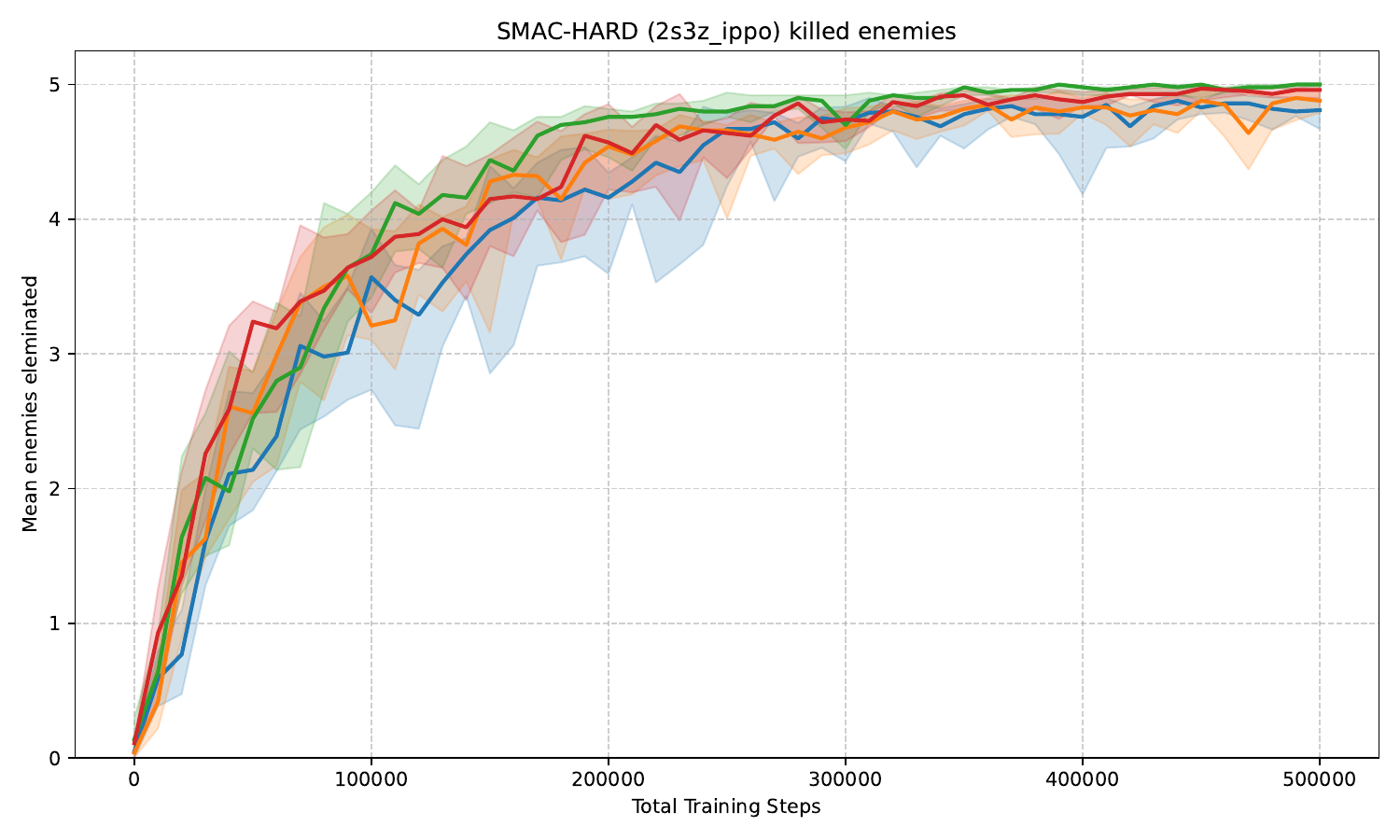}
        \caption{SMAC-HARD (2s3z--IPPO)}
        \label{fig:2s3z-enem-ippo}
    \end{subfigure}
    \hfill
    \begin{subfigure}[t]{0.32\textwidth}
        \centering
        \includegraphics[width=\linewidth]{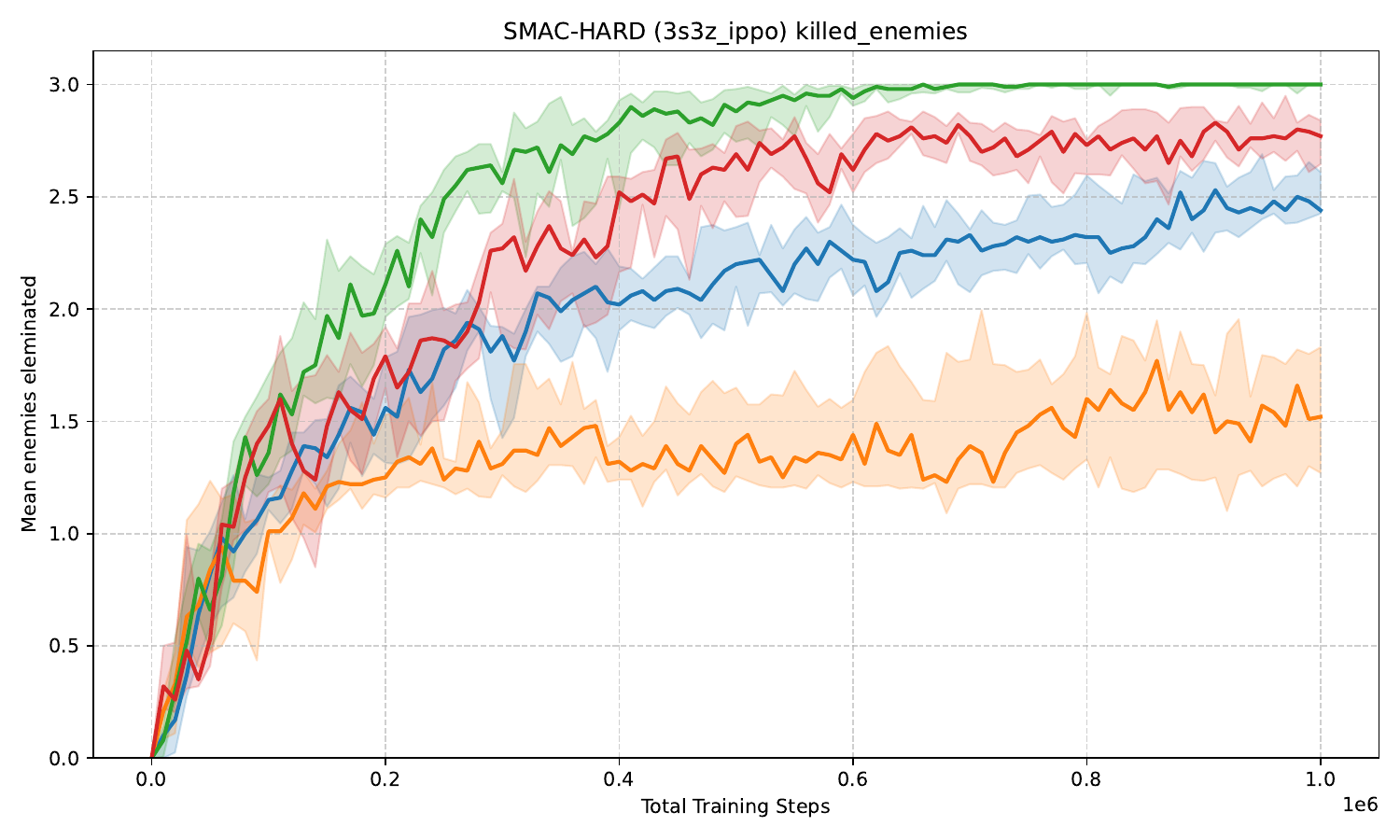}
        \caption{SMAC-HARD (3s\_vs\_3z--IPPO)}
        \label{fig:3svs3z-enem-ippo}
    \end{subfigure}

    \caption{Learning curves for \method and LIAM on SMAC-HARD, showing the median number of eliminated opponents (universally quantified) and the 25th–75th percentiles.}
    \label{fig:main-base-slicr-enem}
\end{figure*}

\begin{figure*}[t]
    \centering

\vspace{-1mm}
    
    \begin{subfigure}[t]{\textwidth}
        \centering
        \includegraphics[width=0.99\linewidth]{img/liam-slicr.pdf}
    \end{subfigure}

    \begin{subfigure}[t]{0.32\textwidth}
        \centering
        \includegraphics[width=\linewidth]{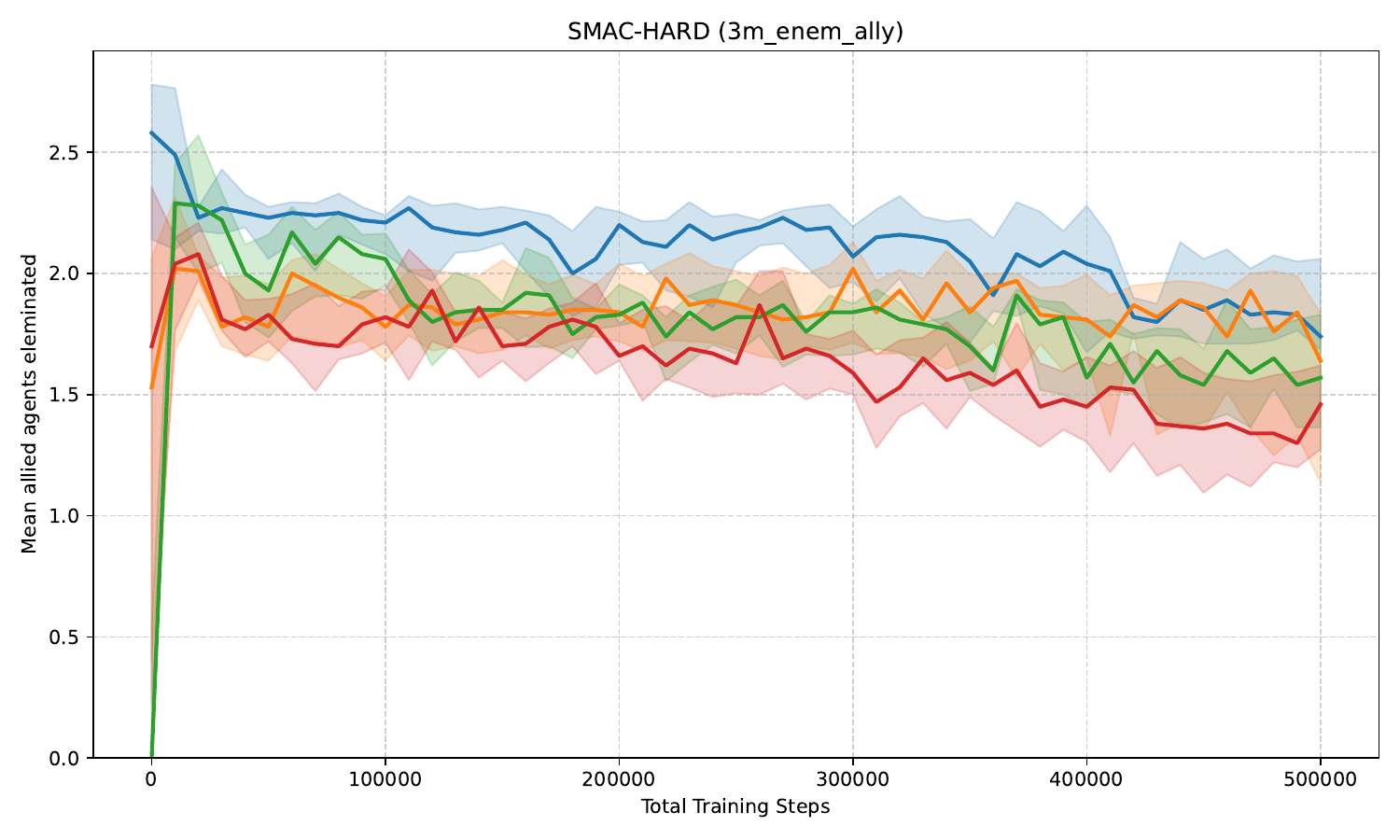}
        \caption{SMAC-HARD (3m)}
        \label{fig:3m-ally}
    \end{subfigure}
    \hfill
    \begin{subfigure}[t]{0.32\textwidth}
        \centering
        \includegraphics[width=\linewidth]{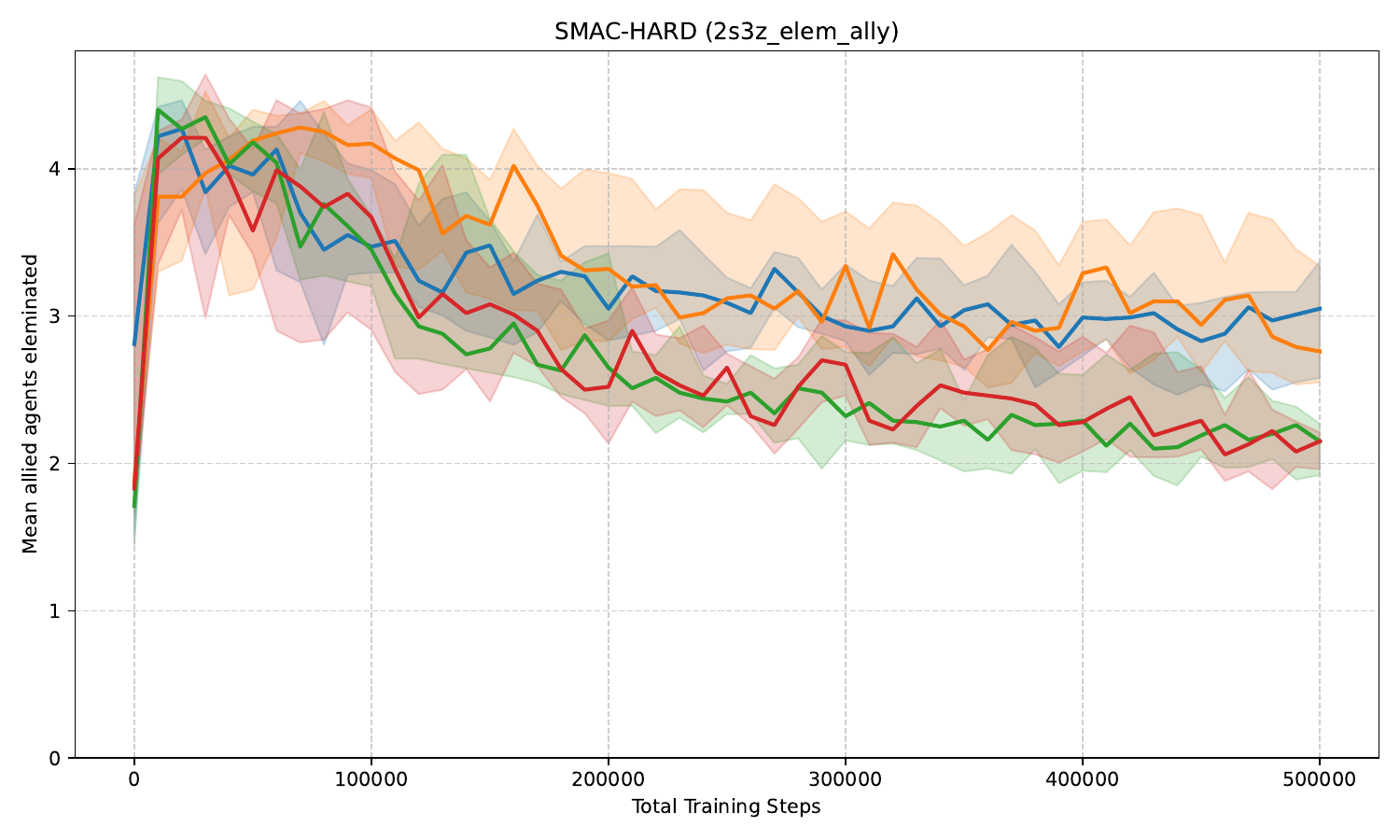}
        \caption{SMAC-HARD (2s3z)}
        \label{fig:2s3z-ally}
    \end{subfigure}
    \hfill
    \begin{subfigure}[t]{0.32\textwidth}
        \centering
        \includegraphics[width=\linewidth]{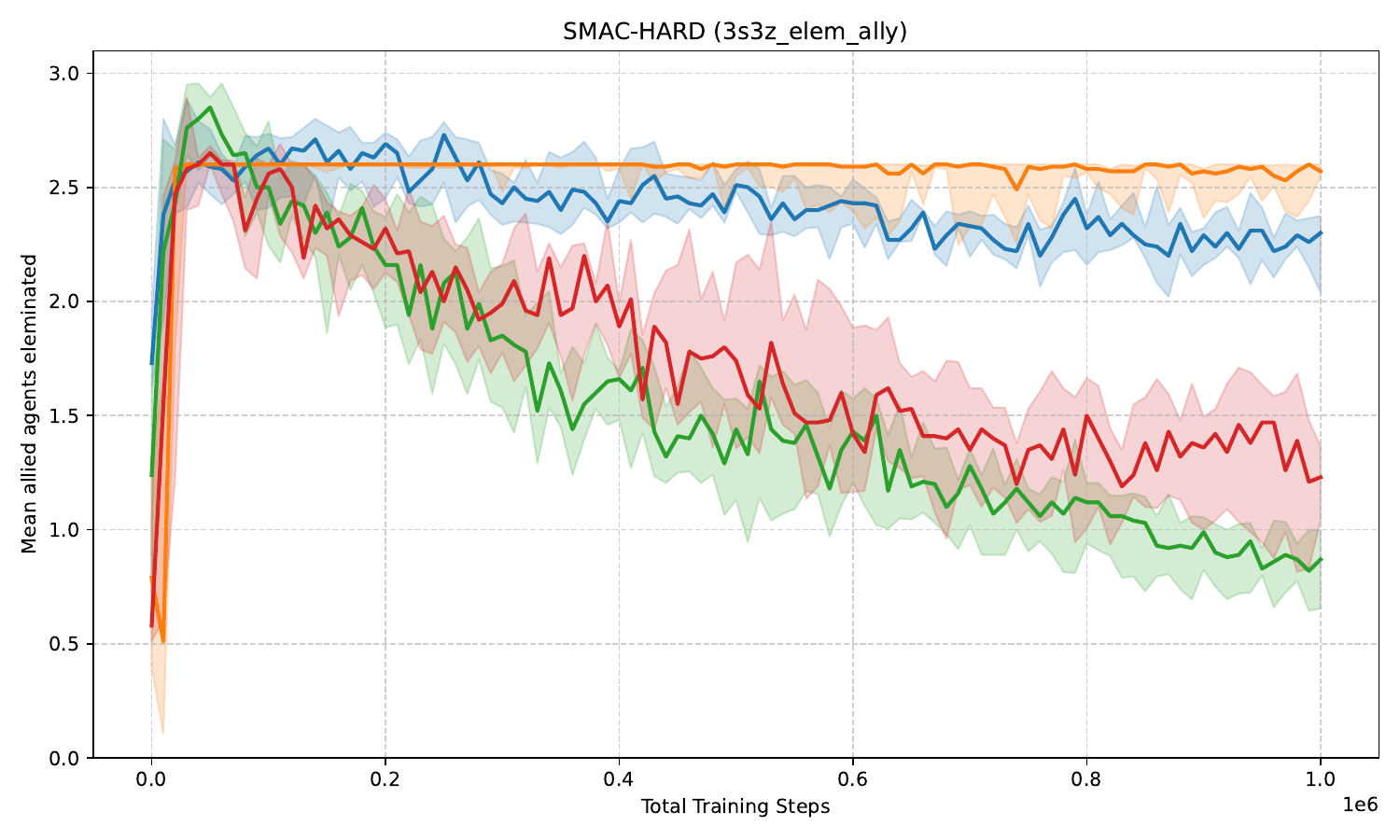}
        \caption{SMAC-HARD (3s\_vs\_3z)}
        \label{fig:3svs3z-ally}
    \end{subfigure}

        \begin{subfigure}[t]{0.32\textwidth}
        \centering
        \includegraphics[width=\linewidth]{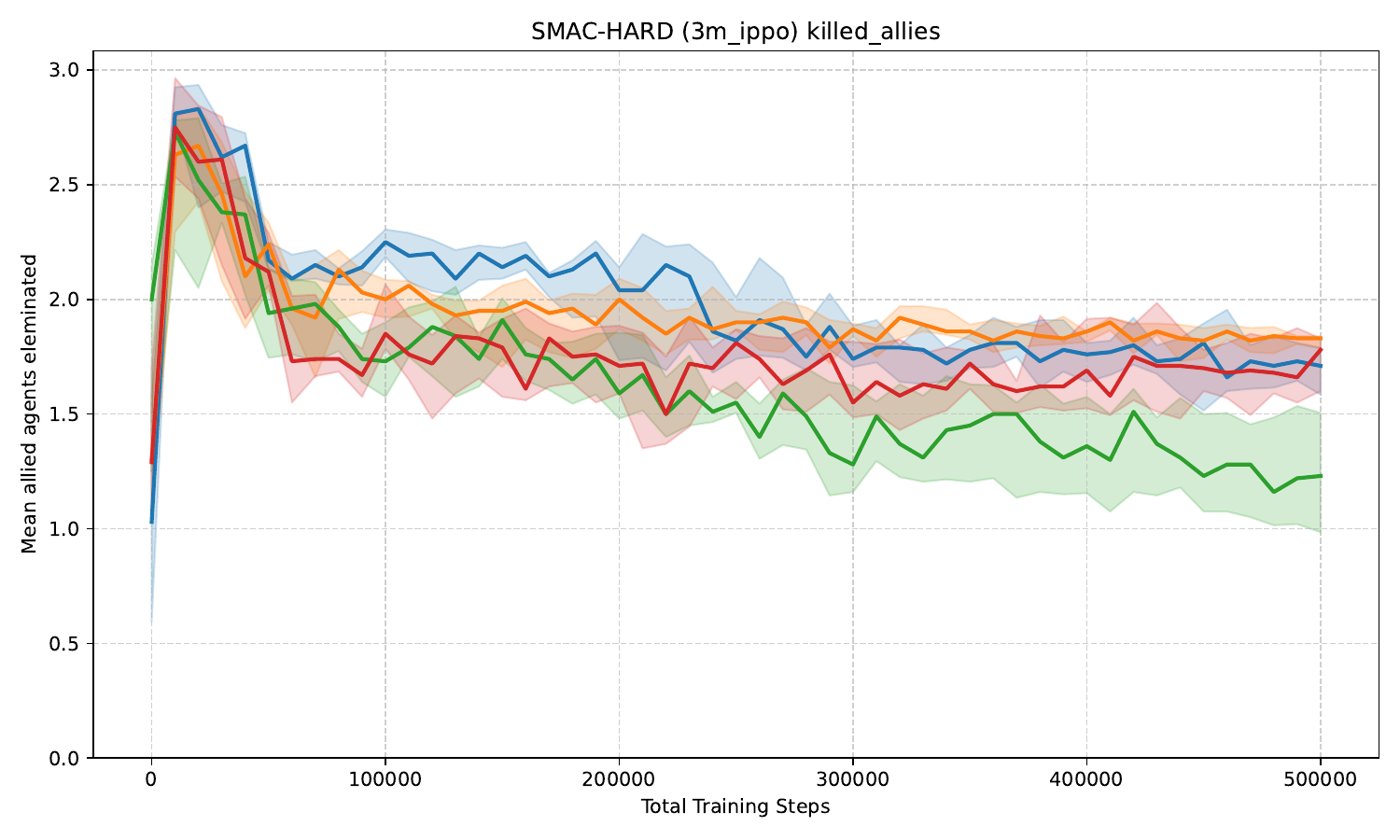}
        \caption{SMAC-HARD (3m--IPPO)}
        \label{fig:3m-ally-ippo}
    \end{subfigure}
    \hfill
    \begin{subfigure}[t]{0.32\textwidth}
        \centering
        \includegraphics[width=\linewidth]{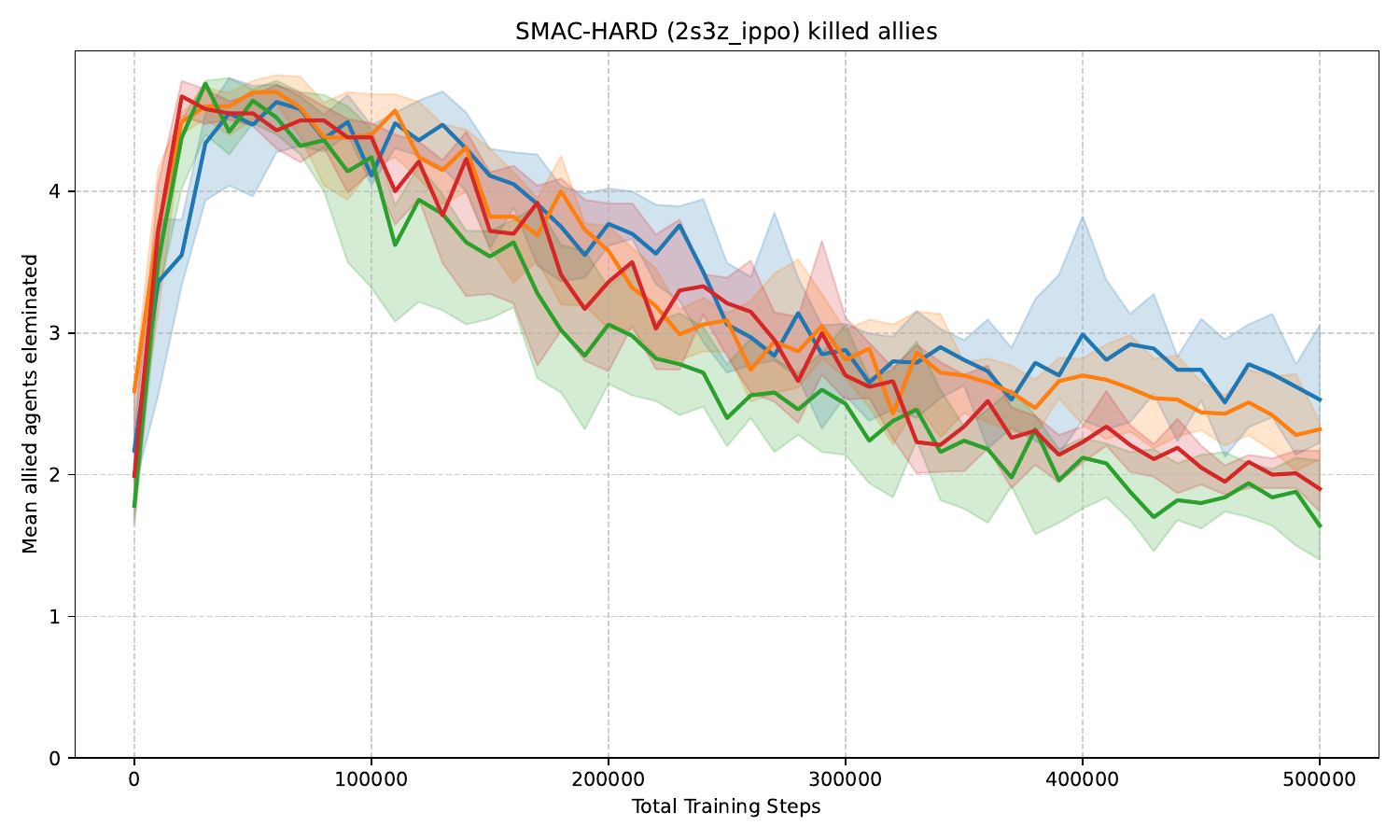}
        \caption{SMAC-HARD (2s3z--IPPO)}
        \label{fig:2s3z-ally-ippo}
    \end{subfigure}
    \hfill
    \begin{subfigure}[t]{0.32\textwidth}
        \centering
        \includegraphics[width=\linewidth]{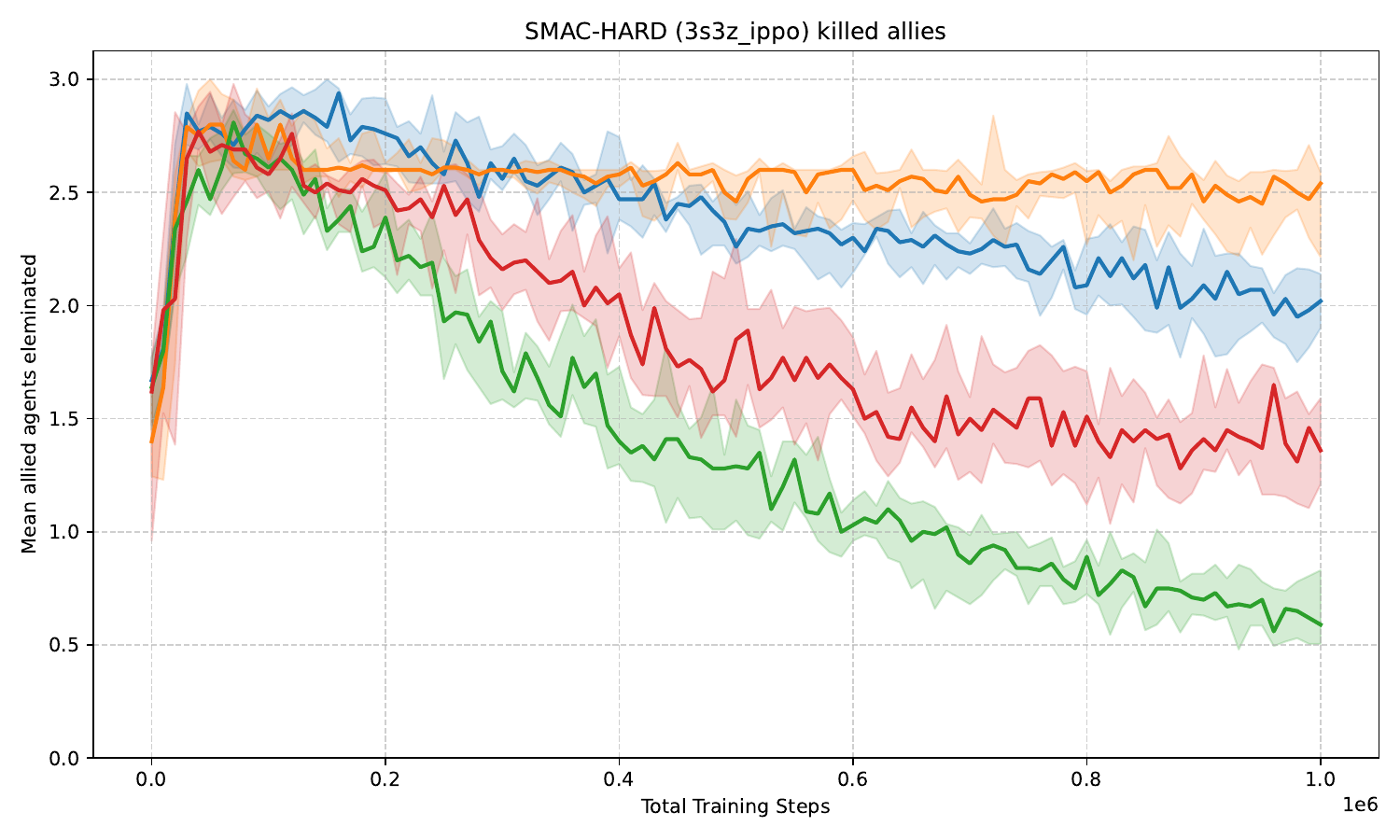}
        \caption{SMAC-HARD (3s\_vs\_3z--IPPO)}
        \label{fig:3svs3z-ally-ippo}
    \end{subfigure}

    \caption{Learning curves for \method and LIAM on SMAC-HARD, showing the median number of eliminated Allied Agents (existentially quantified) and the 25th–75th percentiles.}
    \label{fig:main-base-slicr-ally}
\end{figure*}

We also report the number of eliminated agents bound to universally quantified variables in~\cref{fig:main-base-slicr-enem} and existentially quantified variables in~\cref{fig:main-base-slicr-ally}.
As shown in~\cref{fig:3svs3z-enem} in \texttt{3s\_vs\_3z} map, \method~\greenline eliminates all agents bound to universally quantified variables in the \texttt{Seen} setting. In the \texttt{Unseen} setting, \method~\redline eliminates nearly all such agents, whereas LIAM~\orangeline eliminates only approximately half of them.
Similarly, on the \texttt{3s\_vs\_3z} map (see~\cref{fig:3svs3z-ally}), \method~\greenline preserves two agents bound to existentially quantified variables in the \texttt{Seen} setting. In the \texttt{Unseen} setting, \method~\redline preserves approximately two agents, whereas LIAM~\orangeline loses nearly all of them.

\begin{figure*}[t]
    \centering

\vspace{-1mm}
    
    \begin{subfigure}[t]{\textwidth}
        \centering
        \includegraphics[width=0.99\linewidth]{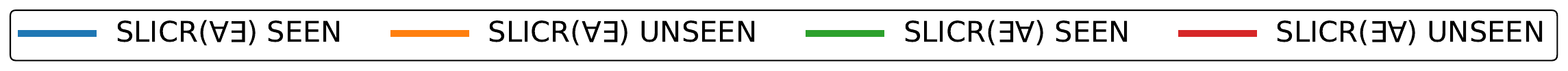}
    \end{subfigure}
    \begin{subfigure}[t]{0.32\textwidth}
        \centering
        \includegraphics[width=\linewidth]{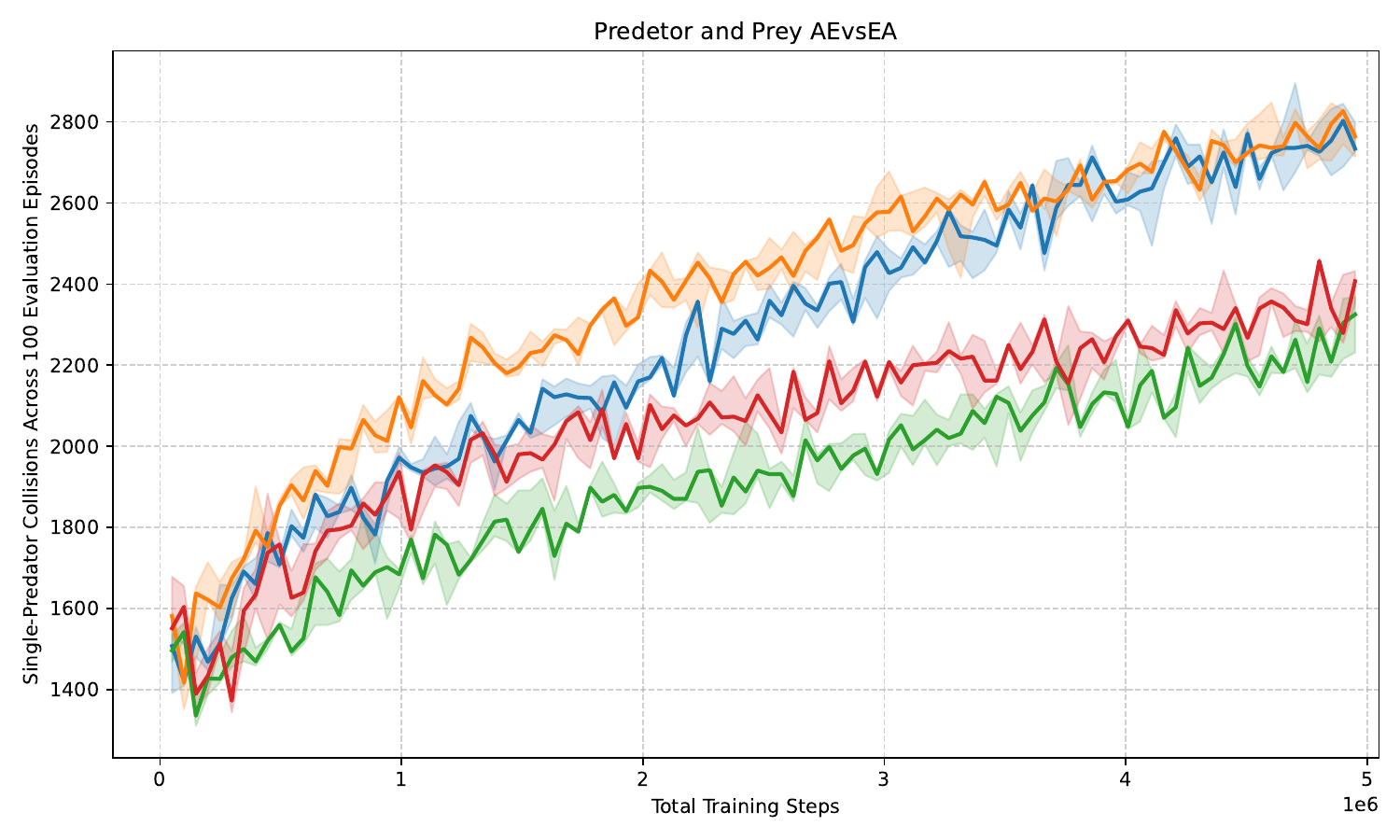}
        \caption{PP}
        \label{fig:PP-ae}
    \end{subfigure}\hfill
    \begin{subfigure}[t]{0.32\textwidth}
        \centering
        \includegraphics[width=\linewidth]{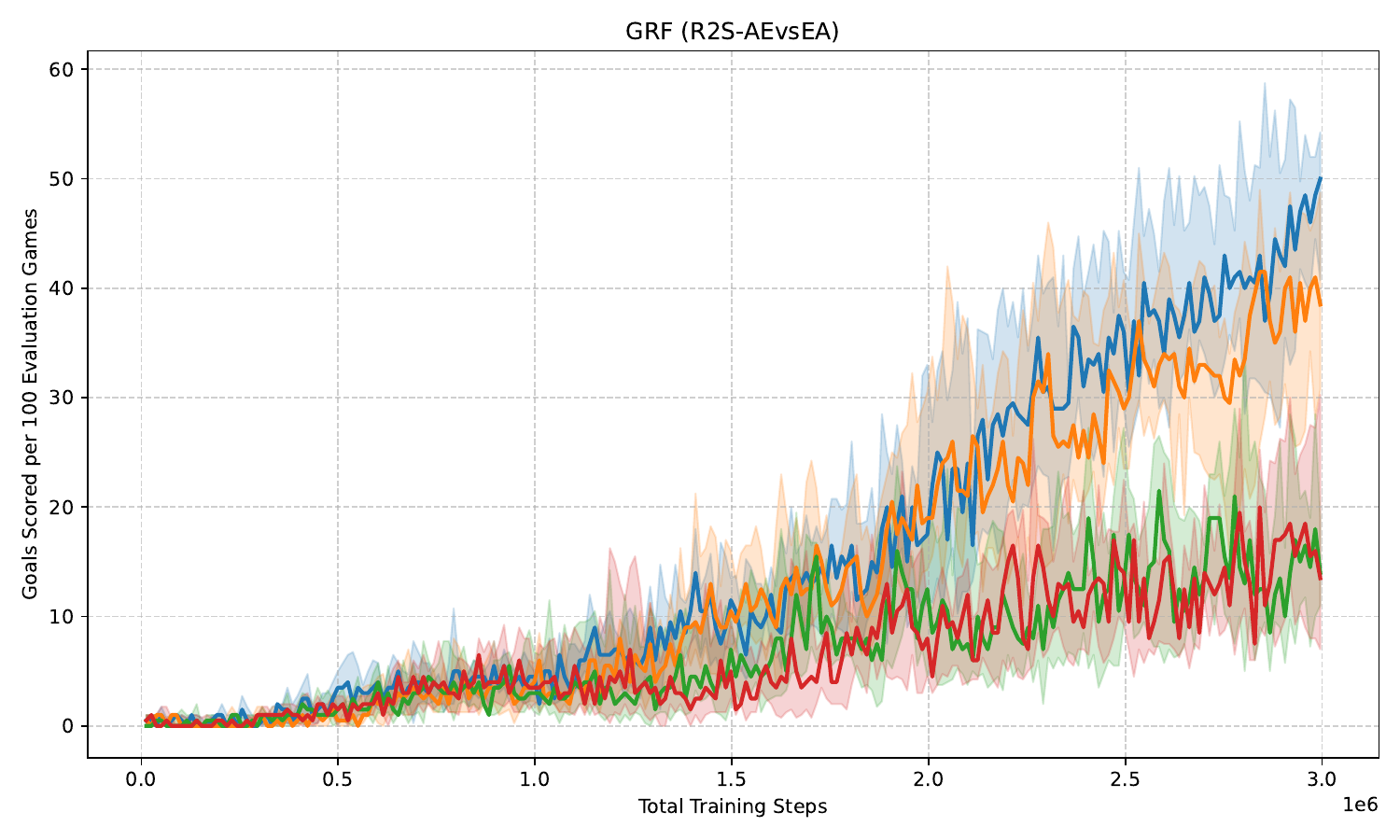}
        \caption{GRF (R2S)}
        \label{fig:GRF(R2S)-ae} 
    \end{subfigure}\hfill
    \begin{subfigure}[t]{0.32\textwidth}
        \centering
        \includegraphics[width=\linewidth]{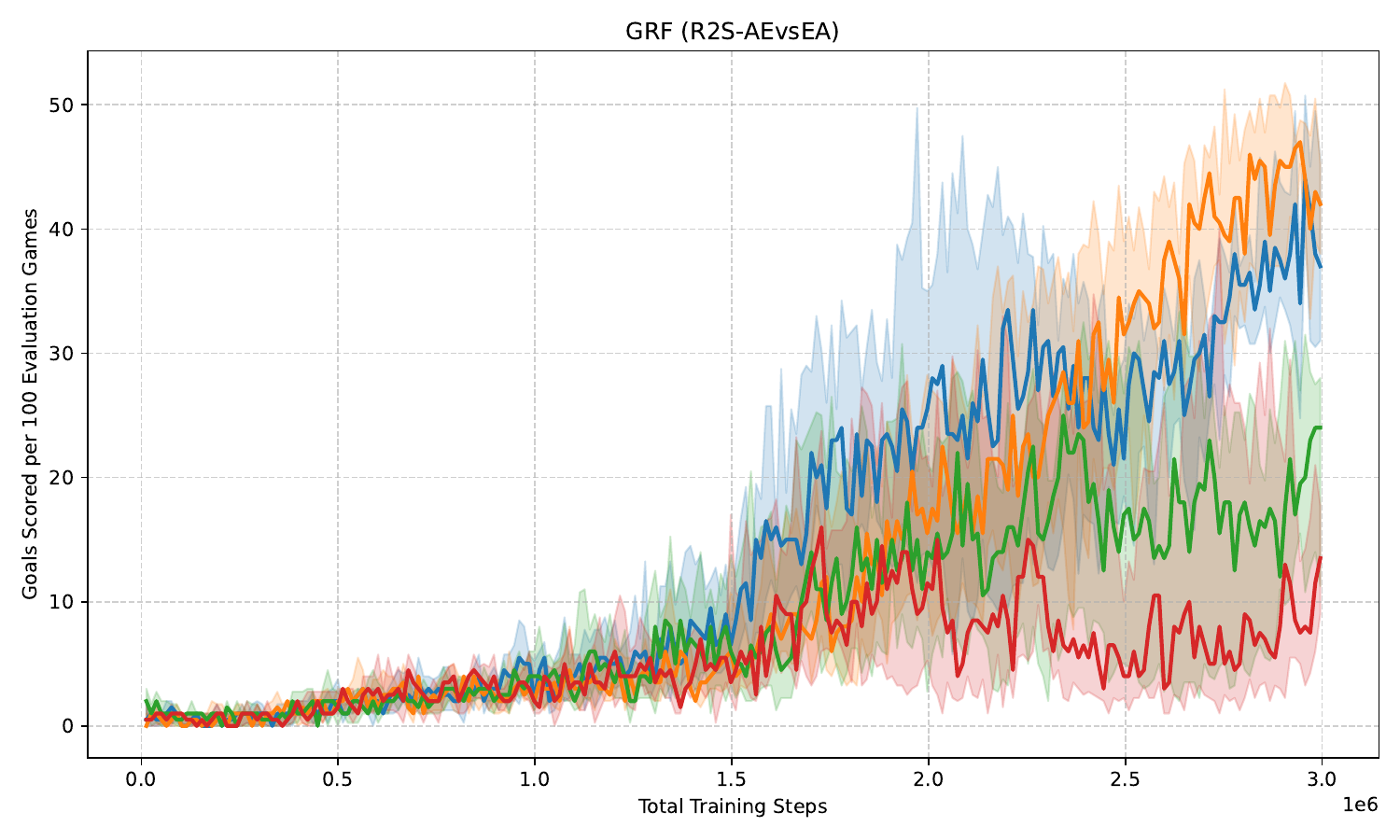}
        \caption{GRF (R2S-DEF)}
        \label{fig:GRF(R2S-HARD-ae)}
    \end{subfigure}
        \hfill
    \begin{subfigure}[t]{0.32\textwidth}
        \centering
        \includegraphics[width=\linewidth]{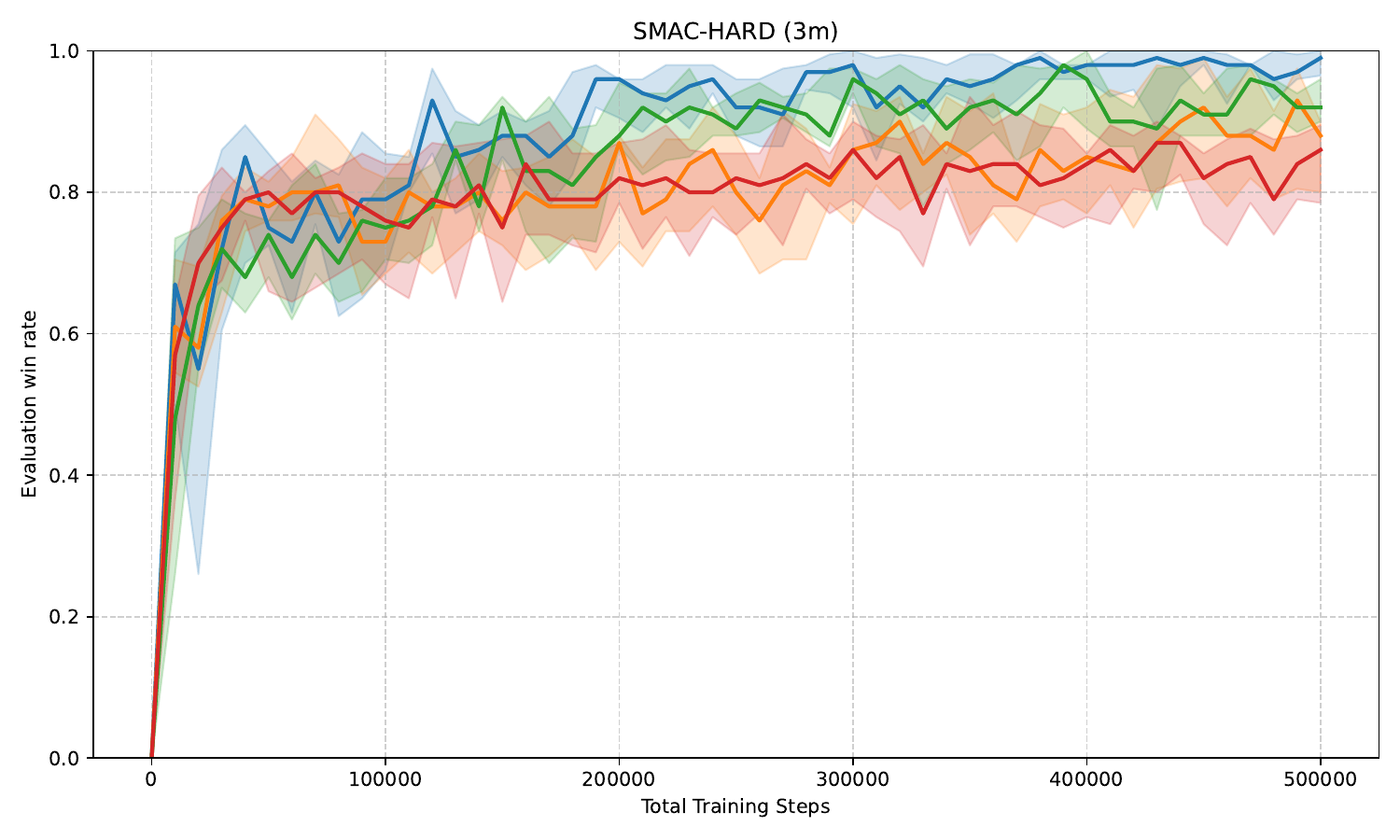}
        \caption{SMAC-HARD (3m)}
        \label{fig:3m-ae}
    \end{subfigure}
    \hfill
    \begin{subfigure}[t]{0.32\textwidth}
        \centering
        \includegraphics[width=\linewidth]{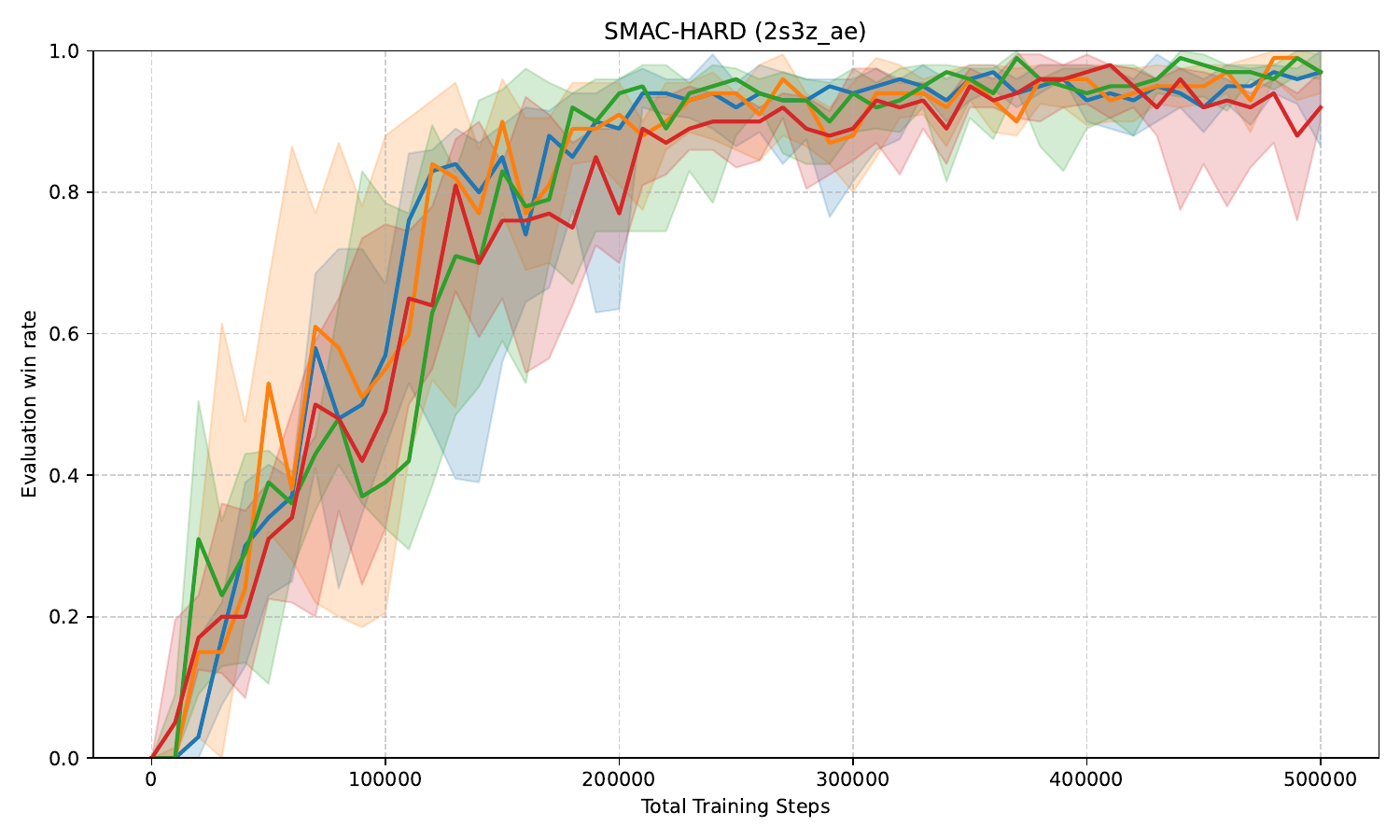}
        \caption{SMAC-HARD (2s3z)}
        \label{fig:2s3z-ae}
    \end{subfigure}
    \hfill
    \begin{subfigure}[t]{0.32\textwidth}
        \centering
        \includegraphics[width=\linewidth]{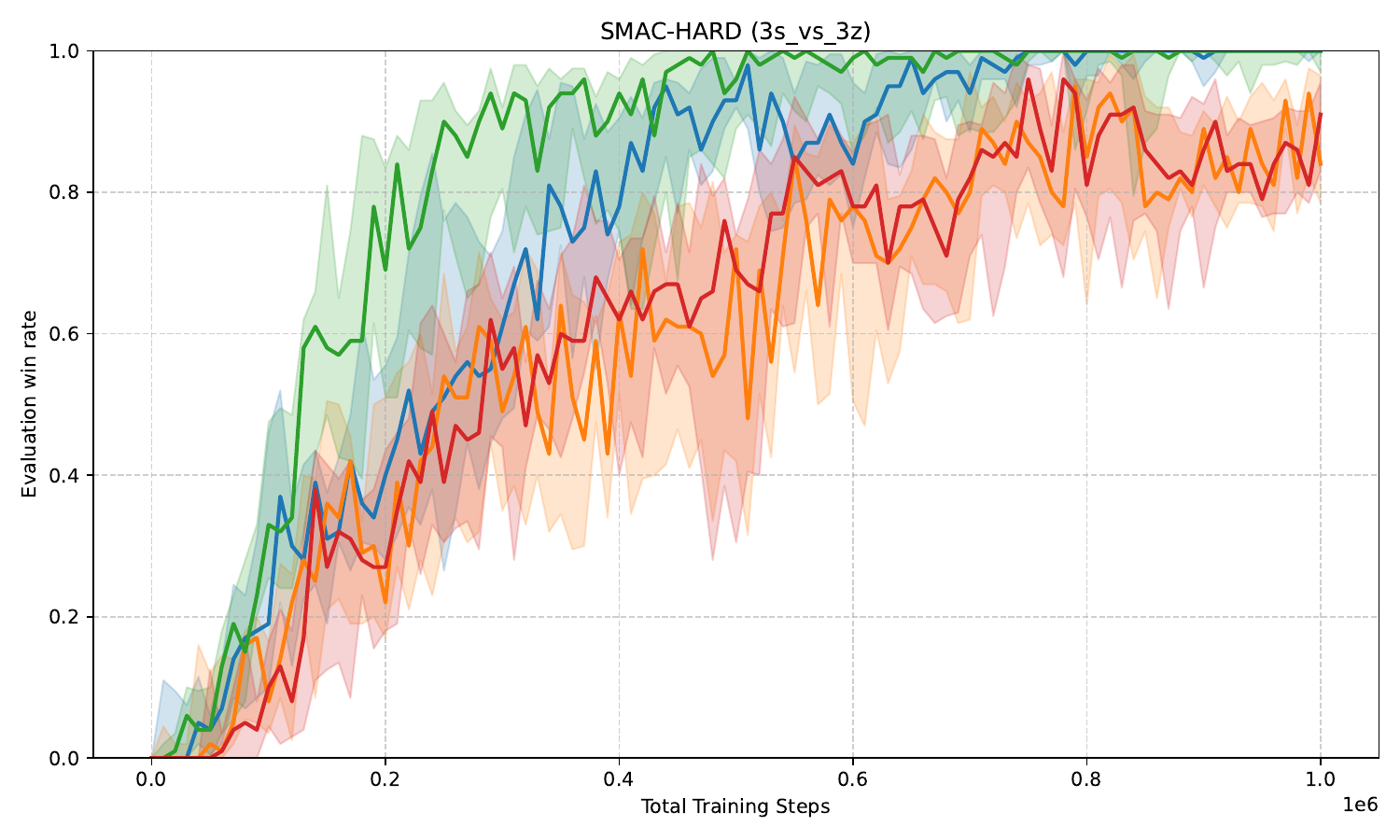}
        \caption{SMAC-HARD (3s\_vs\_3z)}
        \label{fig:3svs3z-ae}
    \end{subfigure}

    \caption{Learning curves for \method $\forall^*\exists^*$ vs. $\exists^*\forall^*$, showing the median and 25th–75th percentiles.}
    \label{fig:slicr-ae}
\end{figure*}

\subsection{\method $\forall^*\exists^*$ vs. $\exists^*\forall^*$}
\label{app:aevsea}
As shown in~\cref{tab:param}, on PP, GRF R2S, and GRF R2S-DEF, \method($\forall^*\exists^*$)~\blueline, \orangeline performs significantly better than \method($\exists^*\forall^*$)~\greenline, \redline in both the \texttt{Seen} and \texttt{Unseen} settings.
Although~\cref{tab:param} shows that \method($\forall^*\exists^*$) and \method($\exists^*\forall^*$) achieve similar overall performance on the SMAC-HARD scenarios, their learning curves reveal important differences. As shown in~\cref{fig:3m-ae,fig:2s3z-ae,fig:3svs3z-ae}, \method($\forall^*\exists^*$)~\blueline, \orangeline converges faster and achieves better results than \method($\exists^*\forall^*$)~\greenline, \redline on the \texttt{3m} map in both the \texttt{Seen} and \texttt{Unseen} settings.

\subsection{Ablation Study of the Shield}
\label{app:shield-ab}
\begin{figure}[h!]
    \centering
    \includegraphics[width=0.6\linewidth]{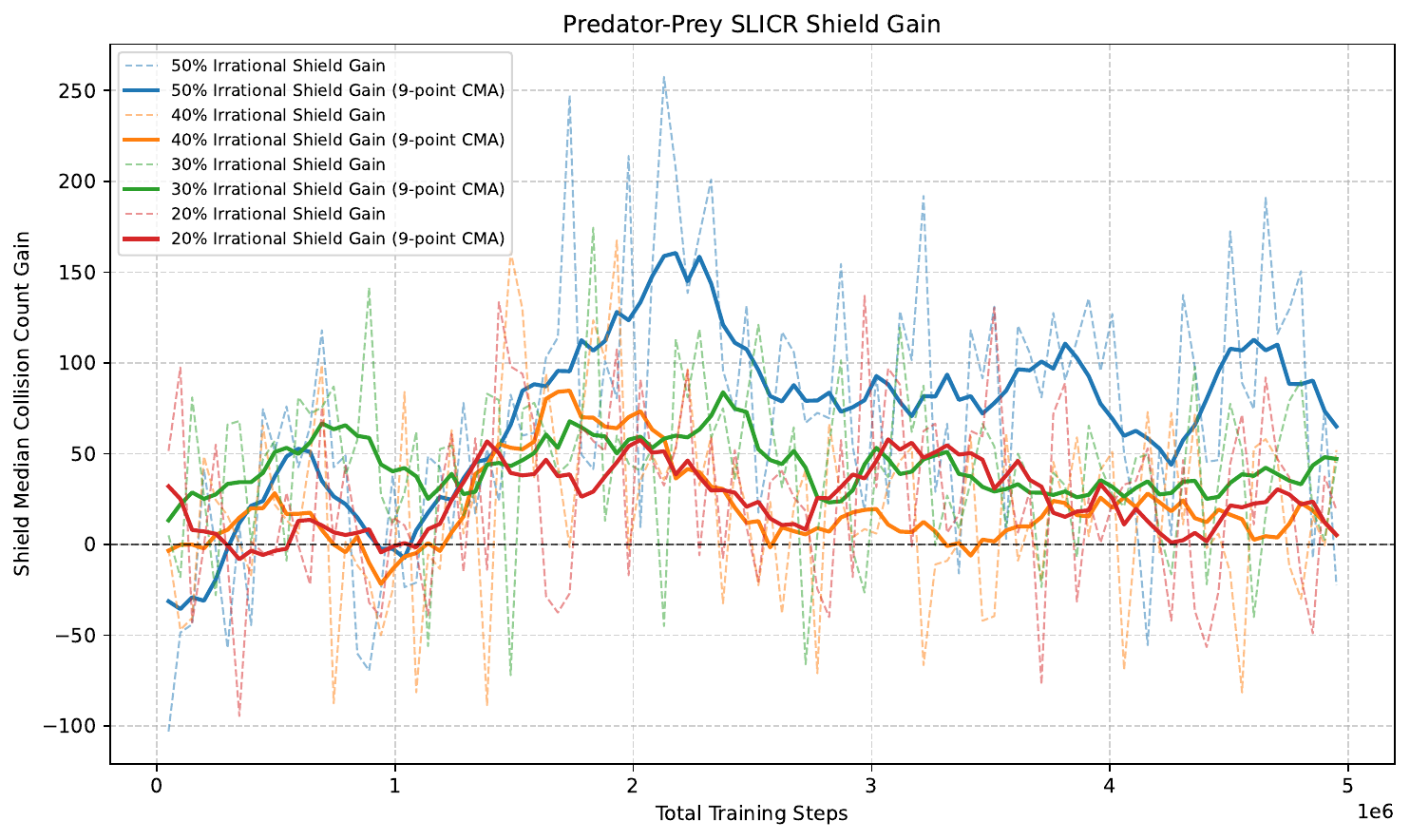}
    \captionof{figure}{Performance gains from the shield under different proportions of irrational strategies.}
    \label{fig:app-ab-shield}
\end{figure}

In addition to the experiments comparing \method with LIAM in the presence of irrational strategies, we conduct an ablation study comparing \method with \method without the shield $\shield$, as shown in~\Cref{fig:app-ab-shield}. The results show that the shield improves the performance of \method across all considered proportions of irrational strategies. In particular, when $50\%$ of the strategies are irrational~\blueline, the shield enables \method to achieve up to $\gain{250}$ more single-predator collisions, with an average gain of approximately $\gain{100}$. When the strategy set contains $40\%$~\orangeline or $30\%$~\greenline irrational strategies, the performance gains are smaller but still reach up to $\gain{150}$ additional single-predator collisions. Finally, when $20\%$ of the strategies are irrational~\redline, the shield continues to provide a positive average performance gain over \method without the shield.

\subsection{Inner-\LTL vs. Shaped Reward.}
\label{app:ltl-ab}
\begin{figure}[h!]
    \centering
    \includegraphics[width=0.6\linewidth]{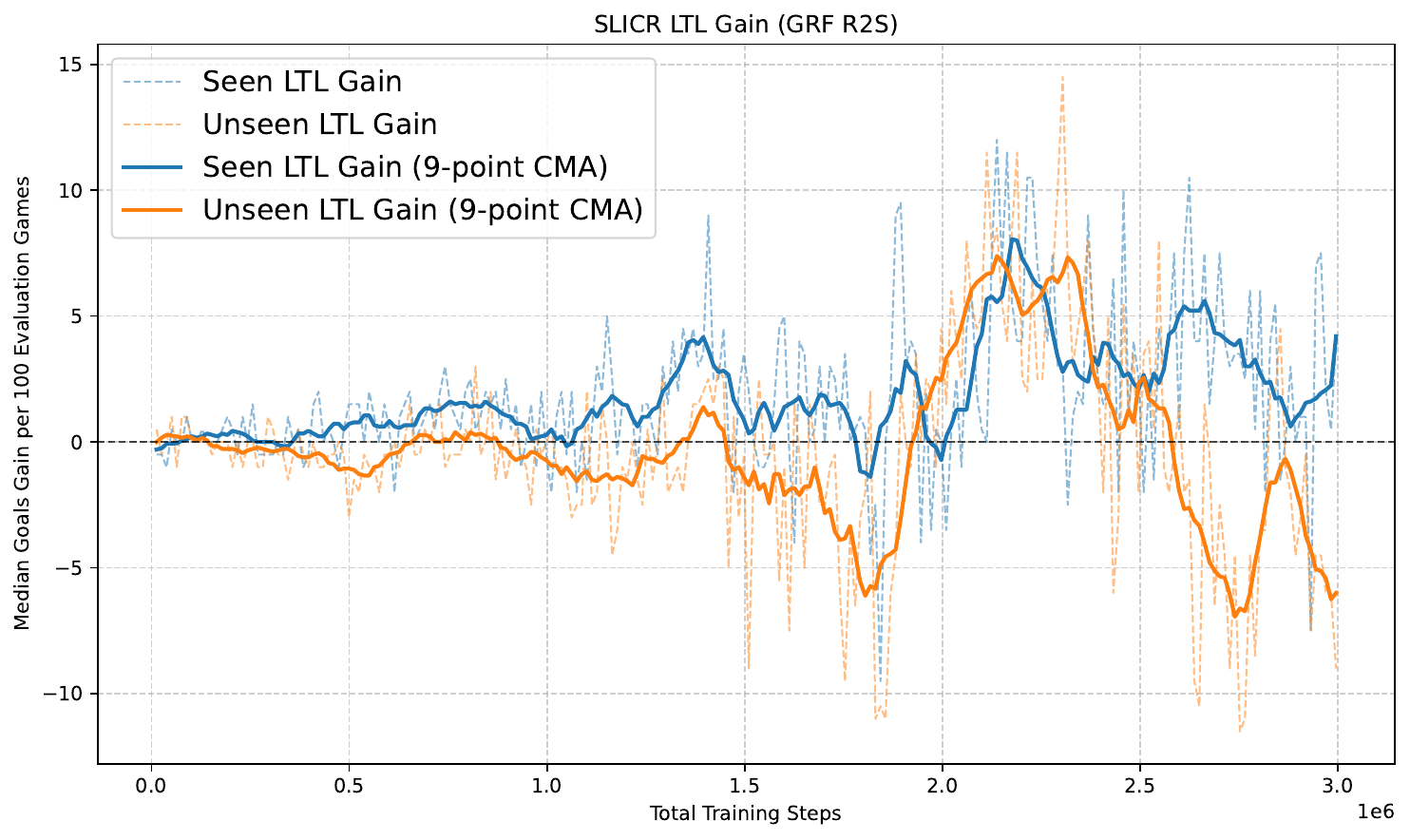}
    \captionof{figure}{Gain from inner-\LTL guidance over shaped rewards on GRF R2S.}
    \label{fig:ltl-app}
\end{figure}

\begin{figure}[h!]
    \centering
    \includegraphics[width=0.6\linewidth]{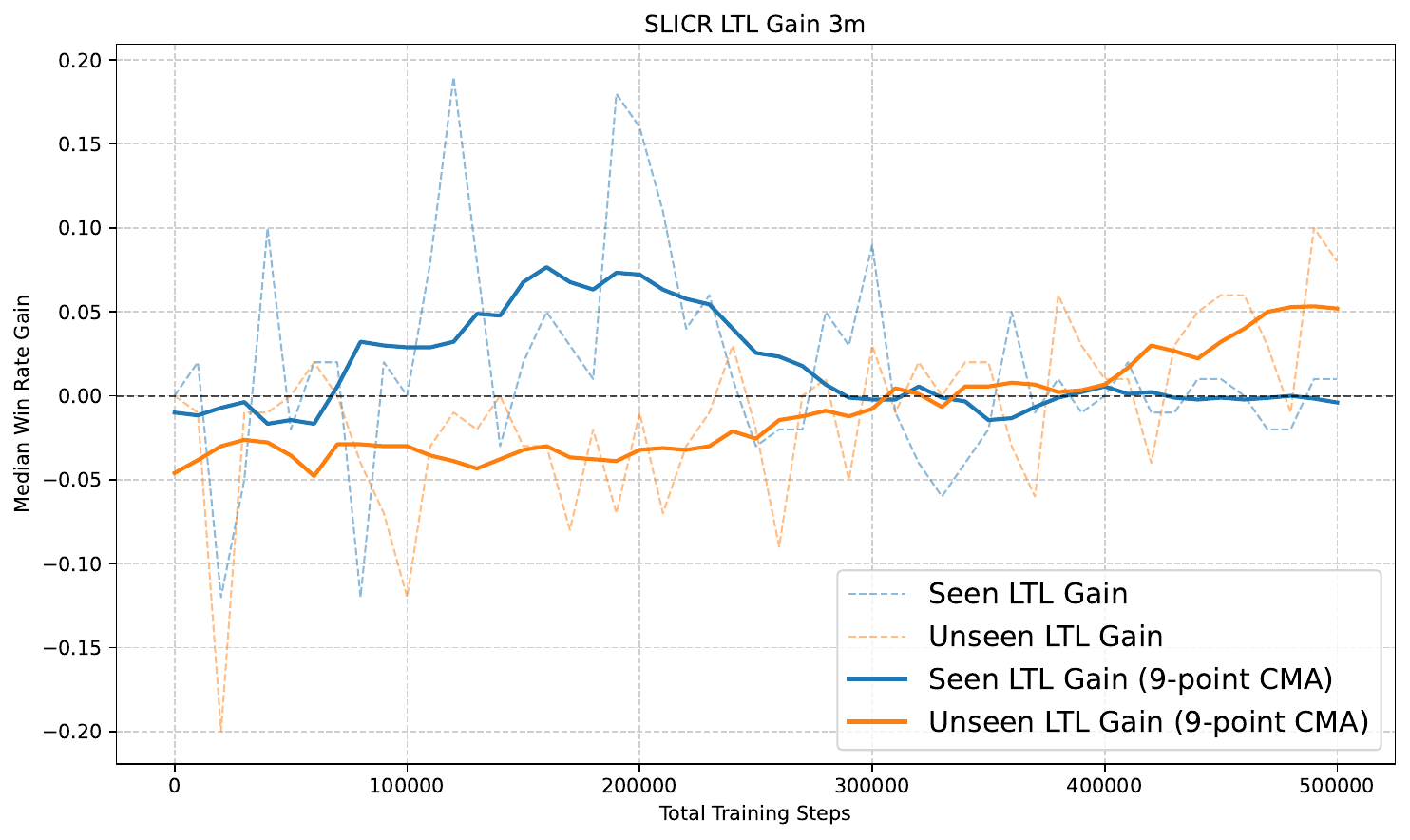}
    \captionof{figure}{Gain from inner-\LTL guidance over shaped rewards on \texttt{3m}.}
    \label{fig:ltl-app-3m}
\end{figure}

\begin{figure}[h!]
    \centering
    \includegraphics[width=0.6\linewidth]{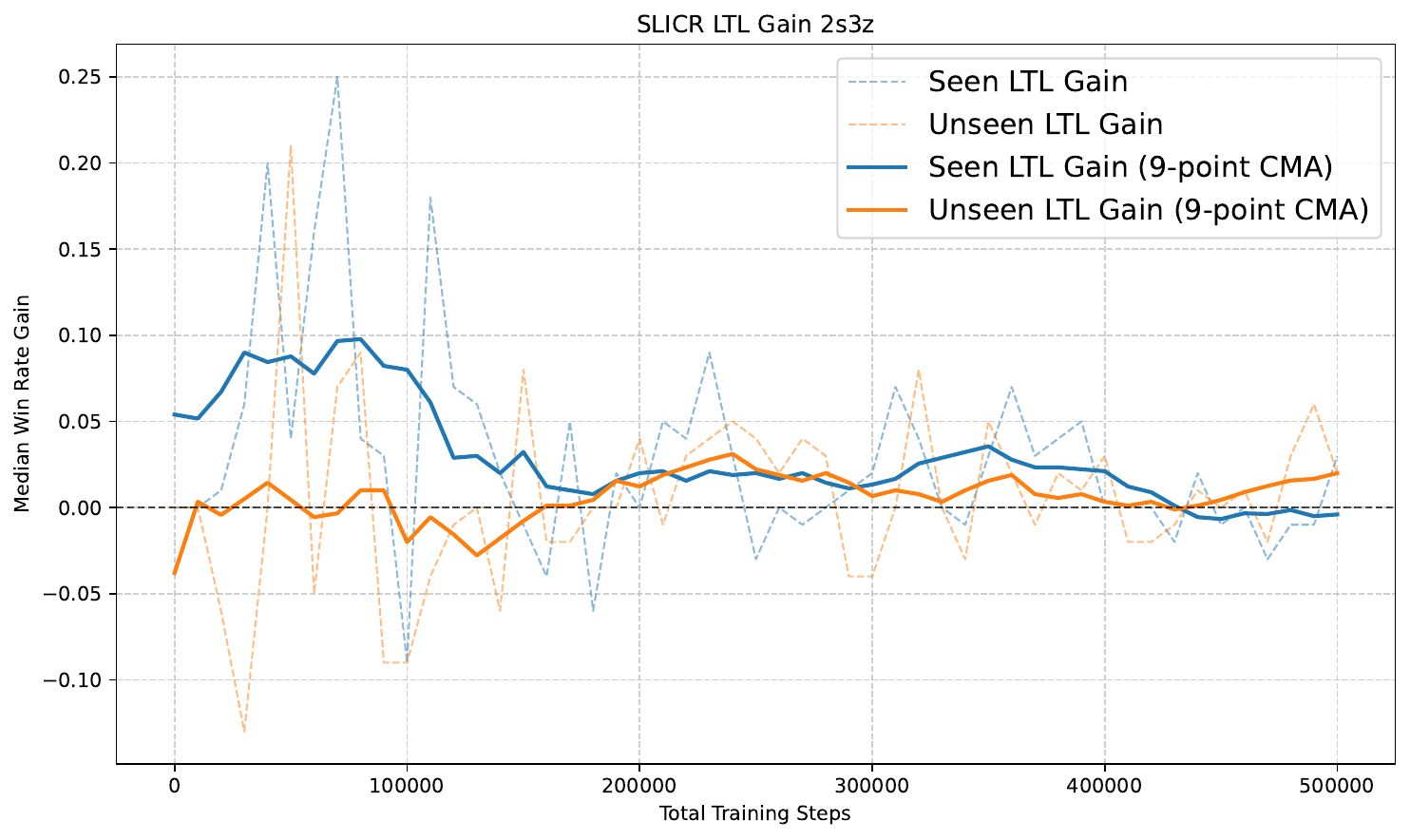}
    \captionof{figure}{Gain from inner-\LTL guidance over shaped rewards on SMAC-HARD \texttt{2s3z}.}
    \label{fig:ltl-app-2s3z}
\end{figure}

In addition to GRF R2S-DEF, where we investigate the effect of inner-\LTL guidance. In~\cref{fig:ltl-app}, we conduct the same analysis on the GRF R2S scenario. In the \texttt{Seen} setting~\blueline, inner-\LTL of formula $\formulars$ guidance achieves up to $\gain{10}$ more goals than the shaped reward. However, in the \texttt{Unseen} setting~\orangeline, the shaped reward performs better during most of training, with inner-\LTL guidance outperforming it only between 2M and 2.5M training steps. We conduct similar experiments on the SMAC-HARD benchmarks. On the \texttt{3m} map (see~\Cref{fig:ltl-app-3m}), inner-\LTL guidance achieves up to a $\gain{18\%}$ higher win rate early in training in the \texttt{Seen} setting~\blueline. However, the gain becomes negligible once both \method and \method without inner-\LTL guidance reach nearly $100\%$ win rates. Similarly, on the \texttt{2s3z} map (see~\Cref{fig:ltl-app-2s3z}), inner-\LTL guidance achieves gains of up to $\gain{25\%}$ and $\gain{20\%}$ early in training in the \texttt{Seen}~\blueline and \texttt{Unseen}~\orangeline settings, respectively. After approximately 200K training steps, both variants reach nearly $100\%$ win rates, and their performance becomes similar.

\section{Computational Resources}
\label{app:compute}
The experimental platform consisted of two AMD EPYC 7742 processors with 128 physical cores and 256 threads, approximately 1 TiB of RAM, and three NVIDIA RTX PRO 6000 Blackwell Max-Q Workstation Edition GPUs with 96 GB of memory each.

%% file: img/robustnessltl.tex
\renewcommand{\arraystretch}{1}
\[
	\begin{array}{llll}
	\centering
	&
	\rbvalueof{\traceof{\trajfromto{\traj}{\ell}{k}}, \tru} 			
	&=& \rbvalue_{\mathit{max}}
	\\
	&\rbvalueof{\traceof{\trajfromto{\traj}{\ell}{k}}, 
	\predicatefuncof{ { \labelsof{\state_\ell}}} } 			
	&=&  \predicatefuncof{\labelsof{\state_\ell}}\\ 
	&
	\rbvalueof{\traceof{\trajfromto{\traj}{\ell}{k}}, \neg\psi} 	 
	&=& - \rbvalueof{\traceof{\trajfromto{\traj}{\ell}{k}}, \psi}\\
	&
	\rbvalueof{\traceof{\trajfromto{\traj}{\ell}{k}}, \X\psi} 			
	&=& \rbvalueof{\traceof{\trajfromto{\traj}{\ell+1}{k}}, \psi} ~~\text{if} ~~ (k > \ell).\\
	&\rbvalueof{\traceof{\trajfromto{\traj}{\ell}{k}}, \G\psi} 			
	&=& \minamong{i\in[\ell,k) }\rbvalueof{\traceof{\trajfromto{\traj}{i}{k}}, \psi}\\  
	&
	\rbvalueof{\traceof{\trajfromto{\traj}{\ell}{k}}, \F\psi} 			
	&=& \maxamong{i \in[\ell,k) }\rbvalueof{\traceof{\trajfromto{\traj}{i }{k}},\psi}  \\
	&\rbvalueof{\traceof{\trajfromto{\traj}{\ell}{k}}, \psi_1 \land \psi_2} 
	&=& ~\minimum\big(\rbvalueof{\traceof{\trajfromto{\traj}{\ell}{k}},  \psi_1},
	\rbvalueof{\traceof{\trajfromto{\traj}{\ell}{k}}, \psi_2}\big) \\
	&\rbvalueof{\traceof{\trajfromto{\traj}{\ell}{k}}, \psi_1 \lor \psi_2}
	&=& ~\maximum\big(\rbvalueof{\traceof{\trajfromto{\traj}{\ell}{k}}, \psi_1} ,
	\rbvalueof{\traceof{\trajfromto{\traj}{\ell}{k}}, \psi_2}\big) \\
	&\rbvalueof{\traceof{\trajfromto{\traj}{\ell}{k}}, \psi_1 \until \psi_2} 
	&=& 
	\maxamong{ i \in[\ell,k) }
	\Big(
	\minimum
	\Big(
	\rbvalueof{\traceof{\trajfromto{\traj}{i}{k}}, \psi_2}, 
	\minamong{ j \in[\ell, i ) }\rbvalueof{\traceof{\trajfromto{\traj}{j}{i}}, \psi_1}
	\Big)
	\Big) \\
    &\rbvalueof{\traceof{\trajfromto{\traj}{\ell}{k}}, \psi_1 \then \psi_2} 
	&=& 
	\maxamong{ i \in[\ell,k) }
	\Big(
	\minimum
	\Big(
	\rbvalueof{\traceof{\trajfromto{\traj}{i}{k}}, \psi_2}, 
	\maxamong{ j \in[\ell, i ) }\rbvalueof{\traceof{\trajfromto{\traj}{j}{i}}, \psi_1}
	\Big)
	\Big) 
\end{array}
\]